\documentclass[12pt]{article}

\RequirePackage{fix-cm}
\usepackage{graphicx}
\usepackage{fullpage}

\usepackage{mathptmx}
\usepackage{amsmath}
\usepackage{amssymb}
\usepackage{epsfig}

\usepackage[scaled=.90]{helvet}
\usepackage{courier}
\usepackage{graphicx}
\usepackage{pgf}
\usepackage{stmaryrd}
\usepackage{amsfonts}
\usepackage{eucal}

\usepackage{amsthm}

\usepackage{mathtools}
\usepackage{multirow}
\usepackage{tabularx}
\usepackage{fancyhdr}

\usepackage[latin1]{inputenc}

\usepackage{titlesec}

\usepackage{caption}

\newtheorem{theorem}{Theorem}
\newtheorem{definition}{Definition}

\newtheorem{observation}{Observation}
\newtheorem{example}{Example}
\newtheorem{proposition}{Proposition}
\newtheorem{lemma}{Lemma}
\newtheorem{corollary}{Corollary}
\newtheorem*{acknowledgements}{\bf Acknowledgements}

\newcommand{\etc}{\textit{etc.}}

\newcommand{\ie}{\textit{i.e. }}

\newcommand{\calA}{\mathcal{A}}
\newcommand{\calB}{\mathcal{B}}

\newcommand{\calF}{\mathcal{F}}
\newcommand{\calI}{\mathcal{I}}

\newcommand{\calL}{\mathcal{L}}
\newcommand{\calM}{\mathcal{M}}

\newcommand{\calP}{\mathcal{P}}
\newcommand{\calR}{\mathcal{R}}
\newcommand{\calS}{\mathcal{S}}
\newcommand{\calW}{\mathcal{W}}
\newcommand{\calX}{\mathcal{X}}
\newcommand{\calY}{\mathcal{Y}}
\newcommand{\calZ}{\mathcal{Z}}
\newcommand{\calV}{\mathcal{V}}

\newcommand{\AF}{\mathcal{AF}}

\newcommand{\Ag}{\mathit{Ag}}
\newcommand{\comps}{\mathtt{Comps}}
\newcommand{\adms}{\mathtt{Adms}}
\newcommand{\labs}{\mathtt{Labs}}
\newcommand{\calLAP}{\calL\calA\calP}

\newcommand{\myin}{\mathtt{in}}
\newcommand{\myout}{\mathtt{out}}
\newcommand{\myundec}{\mathtt{undec}}
\newcommand{\mydec}{\mathtt{dec}}
\newcommand{\comment}[1]{}

\usepackage{authblk}

\begin{document}

\title{Pareto Optimality and Strategy Proofness in Group Argument Evaluation (Extended Version)
}
\author[1,2]{Edmond Awad}
\author[3]{Martin Caminada}
\author[4]{Gabriella Pigozzi}
\author[5]{Miko\l{}aj Podlaszewski}
\author[1,2,$\dagger$]{Iyad Rahwan}

\affil[1]{The Media Lab, Massachusetts Institute of Technology, USA}
\affil[2]{Masdar Institute, UAE}
\affil[3]{School of Computer Science $\&$ Informatics, Cardiff University, UK}
\affil[4]{Universit\'e Paris-Dauphine, PSL Research University, CNRS, UMR [7243], LAMSADE, 75016 Paris, France}
\affil[5]{University of Luxembourg, Luxembourg}

\affil[$\dagger$]{Correspondence should be addressed to
\texttt{irahwan@mit.edu}}
\date{}






\maketitle

\begin{abstract}
An inconsistent knowledge base can be abstracted as a set of arguments and a defeat relation among them. There can be more than one consistent way to evaluate such an argumentation graph. Collective argument evaluation is the problem of aggregating the opinions of multiple agents on how a given set of arguments should be evaluated. It is crucial not only to ensure that the outcome is logically consistent, but also satisfies measures of social optimality and immunity to strategic manipulation. This is because agents have their individual preferences about what the outcome ought to be. In the current paper, we analyze three previously introduced argument-based aggregation operators with respect to Pareto optimality and strategy proofness under different general classes of agent preferences. We highlight fundamental trade-offs between strategic manipulability and social optimality on one hand, and classical logical criteria on the other. Our results motivate further investigation into the relationship between social choice and argumentation theory. The results are also relevant for choosing an appropriate aggregation operator given the criteria that are considered more important, as well as the nature of agents' preferences.

\end{abstract}
\clearpage

\section{Introduction}
Argumentation has recently become one of the main approaches for non-monotonic reasoning and multi-agent interaction in artificial intelligence and computer science \cite{benchcapon:dunne:2007,besnard:hunter:2008,rahwan:simari:2009}. The most prominent approach in argumentation models is probably the abstract argumentation framework (AAF) by Dung  \cite{dung:1995}. In AAF, the contents of the arguments are abstracted from and the framework can be represented as a directed graph in which nodes represent arguments (a set $\calA$), and arcs between these nodes represent binary \emph{defeat} relations (denoted as $\rightharpoonup$) over them.

An important question is which arguments to accept. In his seminal paper, Dung has defined extension-based semantics which correspond to different criteria of acceptability of arguments. For example, if we have two arguments that defeat each other, we cannot accept both. We may accept only one of them. Another equivalent labeling-based semantics is proposed by Caminada \cite{caminada:2006,caminada:gabbay:2009}. Using this approach, an argument is labeled $\myin$ (i.e. accepted), $\myout$ (i.e. rejected), or $\myundec$ (i.e. undecided). 

One of the essential {properties}, that is common, is the condition of \emph{admissibility}: that accepted arguments must not attack one another, and must defend themselves against counter-arguments, by attacking them back. {A stronger} notion is called \emph{completeness}, and is captured, in terms of labelings, in the following two conditions:
\begin{enumerate}
	\item An argument is labeled \emph{accepted} (or \emph{in}) if and only if all its defeaters are rejected (or \emph{out}).
	\item An argument is labeled \emph{rejected} (or \emph{out}) if and only if at least one of its defeaters is accepted (or \emph{in}).
\end{enumerate}

In all other cases, an argument should be labeled \emph{undecided} (or \emph{undec}). Thus, evaluating a set of arguments amounts to labeling each argument using a labeling function $\calL ~:~ \calA \rightarrow \{\myin, \myout, \myundec\}$ to capture these three possible labels. Any labeling that satisfies the above conditions is a \emph{legal labeling}, and corresponds to a complete labeling (to be discussed in more detail below). Every complete ($\ie$ legal) labeling represents a consistent self-defending point of view. { We will use \emph{legal labeling} and \emph{complete labeling} interchangably.}

\begin{figure}[htbp]
    \centering
       \includegraphics[scale=0.6]{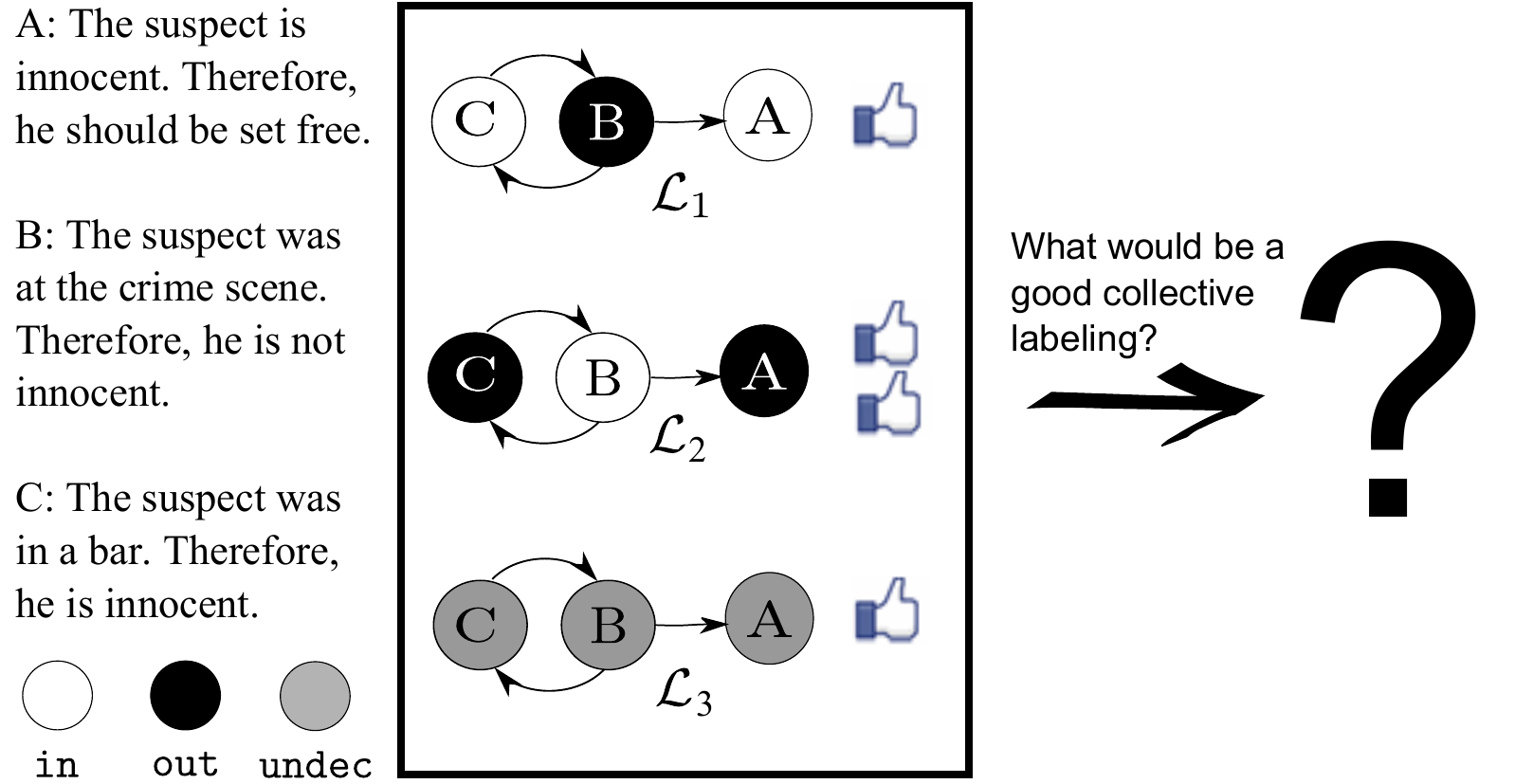}
    \caption{Three different labelings with the votes on each labeling.}
    \label{fig:aggprb}
\end{figure}

Since there can be different reasonable positions regarding the evaluation of an argumentation graph, choosing one legal labeling above another is not a trivial task. Therefore, in a multi-agent setting, different agents can subscribe to different positions. Hence, a group of agents with an argumentation graph would need to find a collective labeling that best reflects the opinion of the group. Consider the following example which is depicted in Figure \ref{fig:aggprb}.

\begin{example}[A Murder Case]
{A murder case is under investigation. There is an argument that the suspect is innocent, which suggests that he should be set free ($A$). However, there is some evidence that the suspect was at the crime scene during the crime time, which suggests that the suspect is not innocent ($B$). Weirdly enough, a witness confirmed that she saw someone who looks like the suspect in a bar during the crime time, which suggests that the suspect is innocent ($C$). 

Clearly, $B$ and $C$ defeat each other since they support negating conclusions. Also, $B$ defeats $A$ since it provides enough evidence to nullify it.

A team of four jurors has been assigned to decide on this case. They have been provided with the previous information. Figure \ref{fig:aggprb} shows the three possible legal labelings. Each juror's judgment can correspond to only one of these labelings. Suppose they voted as shown in Figure \ref{fig:aggprb} (the four thumbs-ups), what would be a labeling that best reflects the opinion of the team?
}
\end{example}

Despite the apparent simplicity of the problem, the aggregation of individual evaluations can result into an inconsistent group outcome i.e. even when each individual submits a legal labeling, the aggregation outcome might not be a legal labeling. This problem of aggregating labelings can be compared to preference aggregation (PA) \cite{arrow:1951,arrow:etal:2002,gartner:2006,vincke1982aggregation}, judgment aggregation (JA) \cite{list:puppe:2009,list2010theory,list2010introduction,grossi2014judgment}, and {non-binary} judgment aggregation \cite{dokow2010aggregation,dokow2010}. These areas have so far blossomed around impossibility results. There exist many differences between labelings and preference relations stemming from their corresponding order-theoretic characterizations. Labeling aggregation differ from JA in that arguments (which are the counterparts of propositions) can have three values instead of two traditionally considered in JA. { Considering the general framework in \cite{dokow2010}, our settings can be considered as focusing on special classes of feasible evaluations, which are the conditions imposed by the legal labelling (or other semantics). Additionally, the possible evaluations of each issue (argument, in our case) are to accept (labels as $\myin$), reject (labels as $\myout$), or be undecided (labels as $\myundec$). However, translation of results between labeling aggregation and non-binary JA amounts to encoding argument semantics in propositional logic, which is not a trivial task \cite{besnard2004checking,besnard2014encoding}.}

Recently, the problem of aggregating valid labelings has been the topic of some studies \cite{rahwan:tohme:2010,awad2015judgment,caminada:pigozzi:2010,booth2014interval,booth2014distances}. In \cite{rahwan:tohme:2010,awad2015judgment}, the argument-wise plurality rule (AWPR) which chooses the collective label of each argument by plurality, independently from other arguments, was defined and analyzed. On the other hand, Caminada and Pigozzi \cite{caminada:pigozzi:2010} showed how judgment aggregation concepts can be applied to formal argumentation in a different way. They proposed three possible operators for aggregating labelings, namely the skeptical operator, the credulous operator, and the super credulous operator. These operators guarantee not only a well-formed outcome but also a compatible one, that is, it does not go against the judgment of any individual. 

In order to assess the three operators, we assume that individuals have preferences over the outcomes. Although the outcomes of the three aggregation operators proposed in \cite{caminada:pigozzi:2010} are compatible with every individual's labeling, this does not mean that they are the most desirable given individuals' preferences. It is possible that other compatible labelings are more desirable. Moreover, it is possible that some agents submit {an} insincere opinion in order to get more desirable outcomes. Given that, it is interesting to study the following two questions:

\begin{enumerate}
	\item Are the social outcomes of the aggregation operators in \cite{caminada:pigozzi:2010} Pareto optimal if preferences between different outcomes are also taken into consideration?
	\item How robust are these operators against strategic manipulation? And what are the effects of strategic manipulation from the perspective of social welfare?
\end{enumerate}

The first question studies the Pareto optimality of the outcomes of these operators. A Pareto optimal outcome (given individuals preferences) cannot be replaced with another outcome that is more preferred by all individuals and is strictly more preferred by at least one individual. Pareto optimality is a fundamental concept in any social choice setting and a clearly desirable property for any aggregation operator.

The second question studies the strategy proofness of the operators. Strategy proofness is fundamental in any realistic multi-agent setting. A strategy-proof operator is one that produces outcomes where individuals have no incentive to misrepresent their votes (i.e. to lie). Unfortunately, as we will see later, most strategy proofness results for the three operators are negative. However, we show later that lies do not always have bad effects on other agents. 

One can realize that individuals' preferences (over all the labelings) play a vital role in answering the previous two questions. However, aggregation operators usually do not give the chance for individuals to disclose these preferences. The labeling an agent submits is the only information available about agent's preferences. It seems a natural choice to assume that the submitted labeling is the most preferred one according to agents' individual preference. Moreover we assume that the rest of agent's preferences can be modeled using distance from the most preferred one. For example, if the top preferred outcome for agent $i$ is the outcome $O_1$ ($\ie$ $\forall O_j$, $O_1 \succeq_i O_j$), then $O_2 \succ_i O_3$ iff $dist(O_1,O_2) < dist(O_1,O_3)$ where $dist(O_1,O_2)$ is the distance between the two outcomes $O_1$ and $O_2$. 

In this work, we investigate different classes of preferences based on different distance measures, and use them to analyze the three aggregation operators proposed in \cite{caminada:pigozzi:2010} with respect to the aforementioned two questions.

This paper makes three distinct contributions. First, it introduces the first thorough study of Pareto optimality and strategy proofness for aggregation operators in the context of argumentation. In doing so, the paper highlights that considering argumentation in multi-agent conflict resolution calls for criteria other than logical consistency such as social optimality and strategic manipulation.

Second, the paper introduces different families of agents' preferences. Building on the new concept of \emph{issue}, proposed by Booth et al. \cite{booth2012quantifying}, we define a new class of agents preferences. We also define a new class of preferences which consider the label $\myundec$ as a middle label between $\myin$ and $\myout$. These new families of preferences capture the intuitions, are more natural, and broaden the scope of analysis of preferences.

The third contribution of this paper is {establishing relations between the different classes of preferences. Some of these relations hold for any aggregation operator and others for some special aggregation operators. Additionally, we provide a full comparison for three previously introduced labeling aggregation operators with respect to the proposed classes of preferences. Moreover, we also consider cases where agents do not share the same classes of preference.} Our results are based on two fundamental criteria, namely Pareto optimality and strategy proofness. For most classes of preferences we establish the superiority of the skeptical operator. However, we also characterize situations where the credulous and super credulous operators are as good as the skeptical operator. This highlights a trade-off between the two criteria on one hand, and seeking more committed outcomes on the other hand.

Our results bridge a gap in our understanding of the social optimality and strategic manipulation of labeling aggregation operators. As for the Pareto optimality, we show the persistence of the superiority of the skeptical operator. However, there are situations where the credulous and super credulous operators are as good as the skeptical operator. This has an implication on the choice of the appropriate aggregation operator given the criteria that is considered more important, as well as, the nature of agents preferences. 

As for the strategy proofness, we establish the fragility of the three operators against strategic manipulation. This negative result is consistent even for a wide range of individual agent preference criteria (except for two cases). This highlights a major limitation of these otherwise attractive approaches to collective argument evaluation.

Despite the negative results, our results show that lies with the skeptical operator are always benevolent i.e. every strategic lie by an agent does not hurt others, but rather improves their welfare. Furthermore, we show that this effect is surprisingly consistent for a wide range of individual agent preference criteria. This shows an important advantage for such an approach to labeling aggregation.\footnote{Part of the results of this paper have been presented in \cite{caminada2011manipulation}.}



\section{Preliminaries}
\subsection{Abstract Argumentation Framework (AAF).\protect\footnote{Readers familiar with AAF can skip this part.}}
\begin{figure}[htbp]
    \centering
       \includegraphics[scale=1]{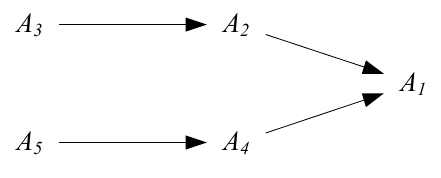}
    \caption{A simple argumentation graph}
    \label{fig:argument_graph1}
\end{figure}

{The seminal paper by Dung \cite{dung:1995} introduced the fundamental notion of abstract argumentation framework} that can be represented as a directed graph where the vertices represent arguments (ignoring details about their contents) and the directed arcs represent the defeat relations between these arguments.\footnote{We will use ``argumentation graph'' and ``argumentation framework'' interchangeably.} For example, in Figure \ref{fig:argument_graph1}, argument $A_1$ is defeated by arguments $A_2$ and $A_4$ which are, in turn, defeated by arguments $A_3$ and $A_5$.

\begin{definition}[Argumentation framework \cite{dung:1995}]
    An \emph{argumentation framework} is a pair $\AF = \langle
    \calA, \rightharpoonup \rangle$ where $\calA$ is a finite set of
    arguments and $\rightharpoonup \subseteq \calA \times \calA$
    is a defeat relation. We say that an argument $A$
    \emph{defeats} an argument $B$ if $(A, B) \in
    \rightharpoonup$ (sometimes written $A ~\rightharpoonup~
    B$).
\end{definition}

There are two approaches to define semantics that assess the acceptability of arguments. One of them is extension-based semantics by Dung \cite{dung:1995}, which produces a set of arguments that are accepted together. Another equivalent labeling-based  semantics is proposed by Caminada \cite{caminada:2006,caminada:gabbay:2009}, which gives a labeling for each argument. With argument labelings, we can accept arguments (by labeling them as $\myin$), reject arguments (by labeling them as $\myout$), and abstain from deciding whether to accept or reject (by labeling them as $\myundec$). As \cite{caminada:pigozzi:2010} employed the labeling approach, so we continue to use it here.

\begin{definition}[Argument labeling \cite{caminada:2006,caminada:gabbay:2009}]\label{definition:labeling}
Let $\AF = \langle \calA, \rightharpoonup \rangle$ be an argumentation
framework. An \emph{argument labeling} is a total function $\calL :
\calA \rightarrow \{\myin, \myout, \myundec \}$.
\end{definition}

\begin{figure}[htbp]
    \centering
       \includegraphics[scale=0.5]{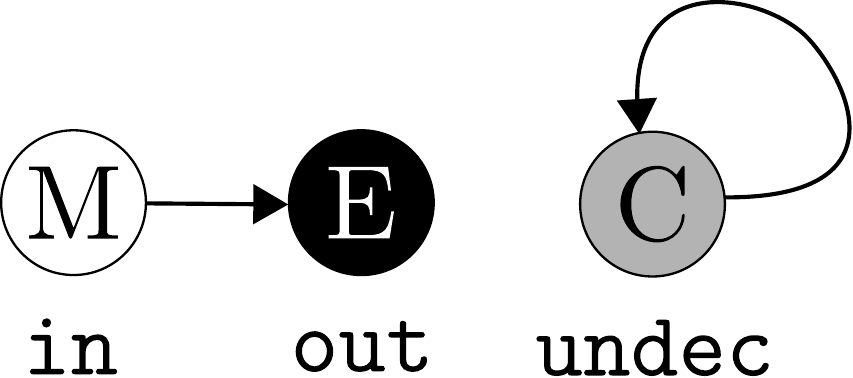}
    \caption{A labeling of an argumentation graph.}
    \label{fig:lab_argument_graph}
\end{figure}

For the purposes of this paper, we use the following marking convention, as shown in Figure \ref{fig:lab_argument_graph}, arguments labeled $\myin$ are shown in white, $\myout$ in black, and $\myundec$ in gray.

We write $\myin(\calL)$ for the set of arguments that are labeled $\myin$ by $\calL$, $\myout(\calL)$ for the set of arguments that are labeled $\myout$ by $\calL$, and $\myundec(\calL)$ for the set of arguments that are labeled $\myundec$ by $\calL$. A labeling $\calL$ can be represented as $\calL=(\myin(\calL)$,$\myout(\calL)$,$\myundec(\calL))$. Equivalently, we also denote a labeling $\calL$ as: $\calL=\{(A,l)|$ $\calL(A)=l \text{ for all } A \in \calA, l \in \{\myin,\myout,\myundec\}\}$.

However, labelings should follow some given conditions. If an argument is labeled $\myin$ then all of its defeaters are labeled $\myout$. If an argument is labeled $\myout$ then at least one of its defeaters is labeled $\myin$. We call a labeling that follows the previous two conditions an \emph{admissible labeling}.

\begin{definition}[Admissible labeling \cite{caminada:2006,caminada:gabbay:2009}]
Let $\AF = \langle \calA, \rightharpoonup \rangle$ be an argumentation framework. An admissible labeling is a mapping $\calL: \calA \rightarrow \{\myin,\myout,\myundec\}$ such that for each $A \in \calA$ it holds that:

if $\calL(A) = \myin$ then $\forall B{\in \calA}:$ $(B \rightharpoonup A \Rightarrow \calL(B) = \myout)$,  and

if $\calL(A) = \myout$ then $\exists B{\in \calA}:$ $(B \rightharpoonup A$ $\wedge$ $\calL(B) = \myin)$.
\end{definition}

Some examples for \emph{admissible} labelings, in Figure \ref{fig:argument_graph1}, can include the following: 
$(\{A_1,A_3,A_5\},$ $\{A_2,A_4\},$ $\emptyset)$, $(\{A_3,A_5\},$ $\{A_2,A_4\},$ $\{A_1\})$, $(\{A_3\},$ $\{A_2\},$ $\{A_1,A_4,A_5\})$, $(\{A_5\},$ $\emptyset,$ $\{A_1,A_2,A_3,A_4\})$, and $(\emptyset,$ $\emptyset,$ $\{A_1,A_2,A_3,A_4,A_5\})$. 

One can realize that in {an} admissible labeling, unlike $\myin$-labeled and $\myout$-labeled arguments, $\myundec$-labels do not need to be justified i.e. an argument can be labeled $\myundec$ under an admissible labeling without any condition.
  
The \emph{complete} semantics, however, force $\myundec$-labels to be also justified. A \emph{complete} labeling is an admissible labeling with the following extra condition: If an argument is labeled $\myundec$ then there is no defeating argument that is labeled $\myin$ (that is, there is insufficient ground to label the argument $\myout$) and not all defeating arguments are labeled $\myout$ (that is, there is insufficient ground to label the argument $\myin$). We call a labeling which follows these rules a \emph{complete labeling}.

\begin{definition}[Complete labeling \cite{caminada:2006,caminada:gabbay:2009}]
Let $\AF = \langle \calA, \rightharpoonup \rangle$ be an argumentation framework. A complete labeling is a mapping $\calL: \calA \rightarrow \{\myin,\myout,\myundec\}$ such that for each $A \in \calA$ it holds that:

if $\calL(A) = \myin$ then $\forall B{\in \calA}:$ $(B \rightharpoonup A \Rightarrow \calL(B) = \myout)$,

if $\calL(A) = \myout$ then $\exists B{\in \calA}:$ $(B \rightharpoonup A$ $\wedge$ $\calL(B) = \myin)$,  and

if $\calL(A) = \myundec$ then: 

$\neg [\forall B{\in \calA}:$ $(B \rightharpoonup A \Rightarrow \calL(B) = \myout)] \wedge \neg [\exists B{\in \calA}:$ $(B \rightharpoonup A$ $\wedge$ $\calL(B) = \myin)]$
\end{definition}

As an example for a \emph{complete} labeling, in Figure \ref{fig:argument_graph1}, we have only one complete labeling, namely $(\{A_1,A_3,A_5\},$ $\{A_2,A_4\},$ $\emptyset)$. 

\subsection{Aggregation Operators}

{
Perhaps, the most common aggregation rules are the majority rules, in which an alternative is chosen if and only if it receives a number of votes that exceeds some presepecified threshold $k > 0.5 \times N$, where $N$ is the number of voters. However, these rules are not always appropriate. One example is in juries, when the legal or the moral responsibility of the outcome is shared by all individuals. Indeed Ronnegard \cite{ronnegard2015fallacy} argued that the attribution of moral responsibility to all members of a committee is legitimate when the decision is taken through unanimous voting, while it is not necessarily the case otherwise. Another example is when the outcome of the decision can potentially harm some individuals. It was shown in \cite{bonnefon2010behavioral} that people show a preference for more conservative aggregation procedures when the outcome of the decision may involve the infliction of personal harm. Aiming to address such specific scenarios, Caminada and Pigozzi \cite{caminada:pigozzi:2010} proposed three aggregation rules that ensure the compatibility of the outcome with all individuals votes. 
}

{
Before introducing the aggregation operators that were defined in \cite{caminada:pigozzi:2010}, we first define the problem of aggregation. The problem of labeling aggregation can be formulated as a set of individuals that collectively decide how an argumentation framework $\AF = \langle \calA, \rightharpoonup \rangle$ must be labelled. 

\begin{definition}[Labeling aggregation problem \cite{awad2015judgment}]
Let $\Ag=\{1,\ldots,n\}$ be a finite non-empty set of agents, and $\AF = \langle \calA, \rightharpoonup \rangle$ be an argumentation framework. A labeling aggregation problem is a pair $\calLAP= \langle \Ag, \AF \rangle$.
\end{definition}

Each individual $i \in \Ag$ has a labeling $\calL_i$ which expresses the evaluation of $\AF$ by this individual.} A labeling profile $P$ is a set of the labelings submitted by agents in $\Ag$: $P=\{\calL_1,\ldots,\calL_n\}$.\footnote{We follow \cite{caminada:pigozzi:2010} in assuming that the profile is a set of labelings instead of a list of labelings. Although this is not common in judgment aggregation literature where the number of votes matter in many operators, it is not the case for the three operators considered in this study, since they focus on compatibility instead of cardinality. As such, although we list $n$ labelings in the profile, it is possible that a profile has less than $n$ elements, since agents can submit similar labelings.}

A labeling aggregation operator is a function that maps a set of $n$ labelings, chosen from {the set of all labelings,} $\labs$, into a collective labeling.\footnote{Although it would be more precise to use $\labs_{\AF}^S$ to denote the set of all labelings for $\AF = \langle \calA, \rightharpoonup \rangle$ according to semantics $S$, we will often drop $\AF$ and $S$, and use $\labs$ instead when there is no {ambiguity} about the argumentation framework. The same goes for all other notations (e.g. $O_\AF$) that were defined for an $\AF$, when there is no {ambiguity} about the argumentation framework.}

\begin{definition}[Labeling aggregation operator $O_{\AF}$ \cite{caminada:pigozzi:2010}]
{Let $\calLAP= \langle \Ag, \AF \rangle$ be a labeling aggregation problem.} A labeling aggregation operator {for $\calLAP$} is a function $O_{\AF}: 2^{\labs}\setminus\{\emptyset\} \rightarrow \labs$ such that $O_{\AF}(\{\calL_1,\ldots,\calL_n\})=\calL_{Coll}$, where $\calL_{Coll}$ is the collective labeling.
\end{definition}

A labeling $\calL_1$ is said to be \emph{less or equally committed} than another labeling $\calL_2$ if and only if every argument that is labeled $\myin$ by $\calL_1$ is also labeled $\myin$ by $\calL_2$ and every argument that is labeled $\myout$ by $\calL_1$ is also labeled $\myout$ by $\calL_2$.

\begin{definition}[Less or equally committed $\sqsubseteq$ \cite{caminada:pigozzi:2010}]
Let $\calL_1$ and $\calL_2$ be two  labelings of argumentation framework $\AF=\langle \calA, \rightharpoonup \rangle$. We say that $\calL_1$ is less or equally committed as $\calL_2$ ($\calL_1 \sqsubseteq \calL_2$) iff $(\myin(\calL_1) \subseteq \myin(\calL_2)) \wedge (\myout(\calL_1) \subseteq \myout(\calL_2))$.
\end{definition}

Two labelings $\calL_1$ and $\calL_2$ are said to be \emph{compatible} with each other if and only if for every argument, there is no $\myin-\myout$ conflict between the two. In other words, every argument that is labeled $\myin$ by $\calL_1$ is not labeled $\myout$ by $\calL_2$ and every argument that is labeled $\myout$ by $\calL_1$ is not labeled $\myin$ by $\calL_2$.

\begin{definition}[Compatible labelings $\approx$ \cite{caminada:pigozzi:2010}]
Let $\calL_1$ and $\calL_2$ be two labelings of argumentation framework $\AF=\langle \calA, \rightharpoonup \rangle$. We say that $\calL_1$ is compatible with $\calL_2$ ($\calL_1 \approx \calL_2$) iff $(\myin(\calL_1) \cap \myout(\calL_2) = \emptyset) \wedge (\myout(\calL_1) \cap \myin(\calL_2)=\emptyset)$
\end{definition}

{
We now define a compatible operator as the following:
\begin{definition}[Compatible operator]
Let $\calLAP= \langle \Ag, \AF \rangle$ be a labeling aggregation problem, and let $O_{\AF}$ be a labeling aggregation operator for $\calLAP$. We say $O_{\AF}$ is a compatible operator if given any labeling profile $P = \{\calL_1,\ldots,\calL_n\}$, $O_{\AF}(P) \approx \calL_i,\forall i\in\Ag$ i.e. the outcome of $O_{\AF}$ is compatible with each individual's labeling.  
\end{definition}
}

In \cite{caminada:pigozzi:2010}, Caminada and Pigozzi proposed three different aggregation operators, namely the skeptical operator, the credulous operator and the super credulous operator. Each of these operators maps a set of labelings, that are submitted by individuals, into a collective labeling.
{
The following two definitions are used in the definition of these operators:

\begin{definition}[Initial operators $\large\sqcap$, $\large\sqcup$ \cite{caminada:pigozzi:2010}]
{Let $\calLAP= \langle \Ag, \AF \rangle$ be a labeling aggregation problem.} The skeptical initial $\large\sqcap$ and credulous initial $\large\sqcup$ operators are labeling aggregation operators {for $\calLAP$} defined as the following:
\begin{itemize}
\item $\large\sqcap(\{\calL_1,\ldots,\calL_n\})=$ $\{(A,\myin)|\forall i \in \Ag: \calL_i(A)=\myin\}$ $\cup$ $\{(A,\myout)|\forall i \in \Ag: \calL_i(A)=\myout\}$ $\cup$ $\{(A,\myundec)| \exists i \in \Ag:\calL_i(A)\neq \myin \wedge \exists j \in \Ag: \calL_j(A)\neq \myout\}$

\item $\large\sqcup(\{\calL_1,\ldots,\calL_n\})=$ $\{(A,\myin)|\exists i \in \Ag: \calL_i(A)=\myin \wedge \neg \exists j \in \Ag:\calL_j(A)=\myout\}$ $\cup$ $\{(A,\myout)|\exists i \in \Ag: \calL_i(A)=\myout \wedge \neg \exists j \in \Ag:\calL_j(A)=\myin\}$ $\cup$ $\{(A,\myundec)| \forall i \in \Ag: \calL_i(A)=\myundec \vee (\exists j \in \Ag:\calL_j(A)=\myin \wedge \exists k \in \Ag:\calL_k(A)=\myout)\}$ \footnote{We will often use $sio_{\AF}$ and $cio_{\AF}$ to refer to the skeptical initial and credulous initial operators, respectively.}
\end{itemize}
\end{definition}

\begin{definition}[Down-admissible $\downarrow$ and up-complete $\uparrow$ labelings \cite{caminada:pigozzi:2010}]
Let $\calL$ be a labeling of argumentation framework $\AF=\langle \calA, \rightharpoonup \rangle$. The down-admissible labeling of $\calL$, denoted as $\calL$$\downarrow$, is the biggest element of the set of all admissible labelings that are less or equally committed than $\calL$:
\[\forall \calL'\in \adms~:~ (\calL' \sqsubseteq \calL \Rightarrow \calL' \sqsubseteq (\calL\downarrow) \sqsubseteq \calL)\]
where $\adms$ is the set of all admissible labelings for $\AF$. The up-complete labeling of $\calL$, denoted as $\calL$$\uparrow$, is the smallest element of the set of all complete labelings that are bigger or equally committed than $\calL$.
\[\forall \calL'\in \comps~:~ (\calL \sqsubseteq \calL' \Rightarrow \calL \sqsubseteq (\calL\uparrow) \sqsubseteq \calL')\]
\end{definition}

Now, we provide the definitions of the three operators:
\begin{definition}[Skeptical $so_{\AF}$, Credulous $co_{\AF}$ and Super Credulous $sco_{\AF}$ operators \cite{caminada:pigozzi:2010}]
{Let $\calLAP= \langle \Ag, \AF \rangle$ be a labeling aggregation problem.} The skeptical $so_{\AF}$, the credulous $co_{\AF}$ and super credulous $sco_{\AF}$ operators are labeling aggregation operators {for $\calLAP$} defined as the following:
\begin{itemize}
\item $so_{\AF}(\{\calL_1,\ldots,\calL_n\}) = (\large\sqcap(\{\calL_1,\ldots,\calL_n\}))\downarrow$.
\item $co_{\AF}(\{\calL_1,\ldots,\calL_n\}) = (\large\sqcup(\{\calL_1,\ldots,\calL_n\}))\downarrow$.
\item $sco_{\AF}(\{\calL_1,\ldots,\calL_n\}) = ((\large\sqcup(\{\calL_1,\ldots,\calL_n\}))\downarrow)\uparrow$.
\end{itemize}
\end{definition}

Given the set of all admissible labelings $\adms$ for some argumentation framework, it is shown that the outcome of the skeptical aggregation operator is the biggest element in $\adms$ that is less or equally committed to every individual's labeling.

\begin{theorem}[\cite{caminada:pigozzi:2010}]\label{thm.bigadm}
Let $\calL_1, \ldots, \calL_n$ ($n \geq 1$) be labelings of argumentation framework $\AF=\langle \calA, \rightharpoonup \rangle$. Let $\calL_{SO}$ be $so_{\AF}(\{\calL_1, \ldots, \calL_n\})$. It holds that $\calL_{SO}$ is the biggest admissible labeling such that for every $i \in \Ag: \calL_{SO} \sqsubseteq \calL_i$.
\end{theorem}
}
\subsection{Distance Measures}
In this part, we define the family of distance measures that we use to define preferences. {Each of the distance measures we consider is characterized by three choices:
\begin{itemize}
\item Individual arguments vs. Issues (set of arguments). 
\item Set inclusion vs. Quantitative distance.
\item Uniform vs. $\myundec$ in the middle.
\end{itemize}
The combination of all of these choices produces eight different distance measures. We start from the third choice. The uniform vs. $\myundec$ in the middle choice captures the intuition that the distance between accepting an argument ($\myin$) and rejecting it ($\myout$) may be set as equal or superior to the distance of accepting (or rejecting) an argument ($\myin$ or $\myout$) and abstaining on the same argument ($\myundec$). In other words, an $\myin$/$\myout$ disagreement may be as serious or more serious (depending on the contexts) than a $\myin$/$\myundec$ (or a $\myout$/$\myundec$) disagreement.} 

Thus, we consider the following two cases. First, $\myin$, $\myout$, and $\myundec$ are equally distant from each other. {In other words, $dist(\myin,\myout)=dist(\mydec,\myundec)$, where $dist(.)$ is the difference between two labels for one argument, and $\mydec$ is either $\myin$ or $\myout$.} In the other case, we assume that $\myundec$ is in the middle between $\myin$ and $\myout$. Thus, we differentiate between two types of disagreement. One between $\myin$ and $\myout$, and the other between $\mydec$ and $\myundec$. {When considering distance, we assume $dist(\myin,\myout)>dist(\mydec,\myundec)$.}

\subsubsection{Case 1: $\myin$, $\myout$, and $\myundec$ are Equally Distant from Each Other}
\paragraph{Hamming Set and Hamming Distance}~\\
The Hamming set between two labelings $\calL_1$ and $\calL_2$ is the set of arguments that these two labelings disagree upon.

\begin{definition}[Hamming Set $\varominus$]
Let $\calL_1$, $\calL_2$ be two labelings of $\AF=\langle \calA, \rightharpoonup \rangle$. We define the Hamming set between these two labelings as:
\begin{equation}
 \calL_1 \varominus \calL_2 = \{A \in \calA| \calL_1(A)\neq \calL_2(A)\}
\end{equation}
\end{definition}

The Hamming distance between two labelings $\calL_1$ and $\calL_2$ is the number of arguments that these two labelings disagree upon.

\begin{definition}[Hamming Distance $\left |\varominus \right|$]
Let $\calL_1$ and $\calL_2$ be two labelings of $\AF=\langle \calA, \rightharpoonup \rangle$. We define the Hamming distance between these two labelings as:
\begin{equation}
 \calL_1 \left |\varominus\right | \calL_2 = |\calL_1 \varominus \calL_2|
\end{equation}
\end{definition}

\begin{figure}[htbp]
    \centering
       \includegraphics[scale=0.6]{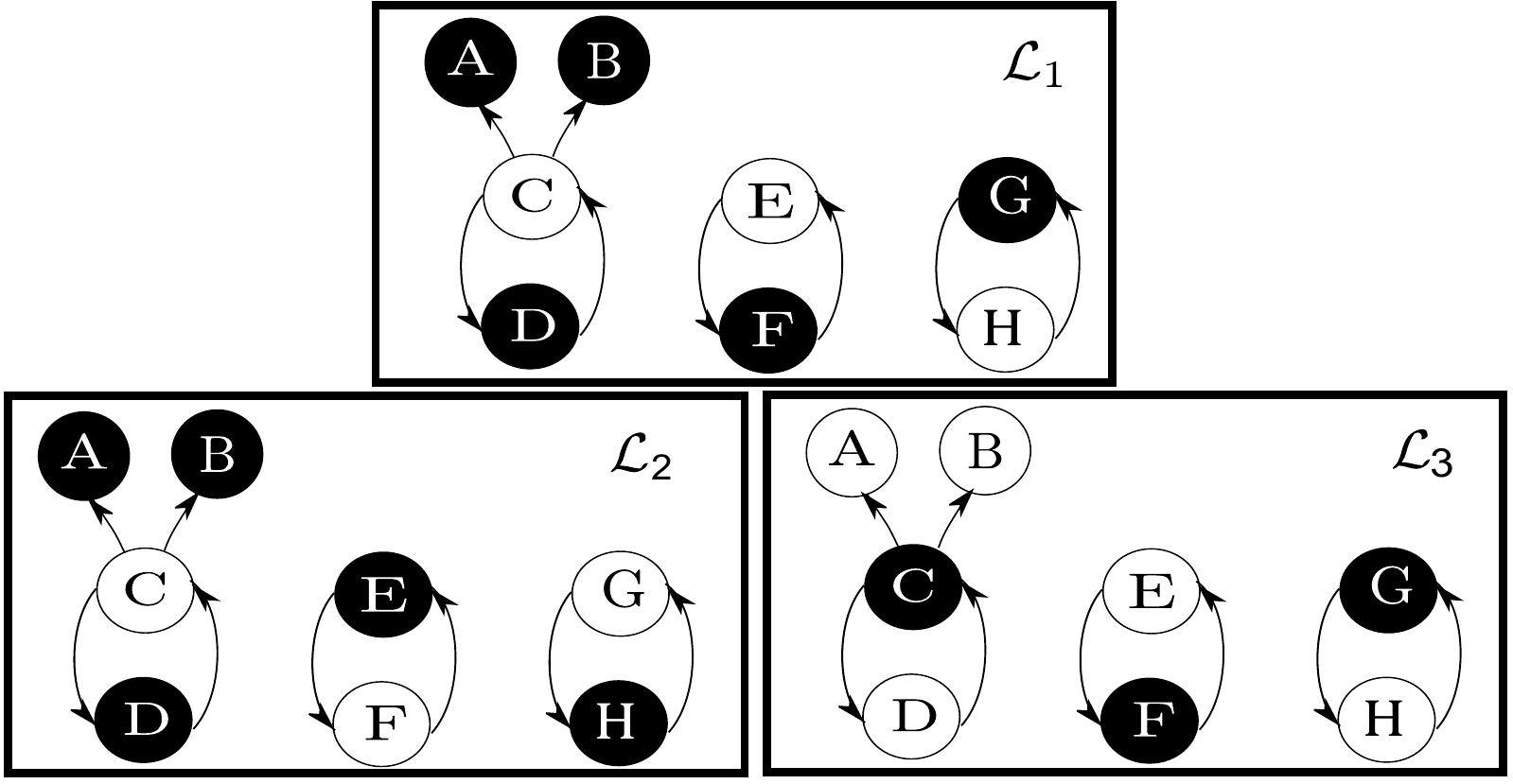}
    \caption{An example showing how Hamming distance is not a suitable distance measure for argument labeling \cite{booth2012quantifying}.}
    \label{fig:issues}
\end{figure}

\paragraph{Issue-wise Set and Issue-wise Distance.}~\\
The label of an argument depends on the labels of the defeating arguments. Therefore, measuring the distance by treating arguments independently might not give an accurate sense of how far two labelings are from each other. Consider the example in Figure \ref{fig:issues}. Using Hamming distance, we have $\calL_1 \left |\varominus\right | \calL_2=\calL_1 \left |\varominus\right | \calL_3=4$.

However, one can argue that $\calL_3$ is closer (than $\calL_2$) to $\calL_1$. Intuitively speaking, if $\calL_1$ and $\calL_3$ further agreed on the labeling of $C$ (or $D$), then they would have been equivalent. On the other hand, $\calL_1$ and $\calL_2$ should further agree on $E$ (or $F$) and $G$ (or $H$) in order to become equivalent. In other words, the number of arguments whose labelings need to be switched in order to make the two labelings be equivalent is less between $\calL_1$ and $\calL_3$ than between $\calL_1$ and $\calL_2$.  

{
Motivated by this example, Booth et al. \cite{booth2012quantifying} proposed a new distance method, using the notion of ``issue'', which they defined. This distance method captures the idea in the previous example, {while satisfying a set of axiomatic properties which they listed as essential for any distance measure.}
}

Crucial to the definition of the ``issue'' is the concept of ``in-sync''. We say that two arguments $A$ and $B$ are \emph{in-sync} if for any pair of labelings $\calL,\calL' \in \labs$, $\calL(A)$ cannot be changed to $\calL'(A)$ without causing a change of equal magnitude when moving from $\calL(B)$ to $\calL'(B)$, and vice versa.

\begin{definition}[in-Sync $\equiv$ {for semantics $S$} \cite{booth2012quantifying}]
Let $\labs^{S}$ be the set of all labelings { according to semantics $S$} for argumentation framework $\AF=\langle \calA, \rightharpoonup \rangle$. We say that two arguments $A, B \in \calA$ are in-sync {for semantics $S$} ($A \equiv^{S} B$):
\begin{equation}
A \equiv^{S} B \text{ iff } (A \equiv_1^{S} B \; \vee \; A \equiv_2^{S} B)
\end{equation}
where:
\begin{itemize}
\item $A \equiv_1^{S} B$ iff $\forall \calL \in \labs^{S}:$ $\calL(A)=\calL(B)$.
\item $A \equiv_2^{S} B$ iff $\forall \calL \in \labs^{S}:$ $(\calL(A)=\myin$ $\Leftrightarrow$ $\calL(B)=\myout)$ $\wedge$ $(\calL(A)=\myout$ $\Leftrightarrow$ $\calL(B)=\myin)$
\end{itemize}
\end{definition}

In-sync is an equivalence relation. We can partition the set of arguments in any argumentation framework $\AF$ into the \emph{in-sync} equivalence classes, which form what is called \emph{issues}.\footnote{The definition of \emph{issue}, along with all the definitions depending on it, can be defined for semantics $S$ (as the case for ``in-sync''). However, from now on, we will restrict all of these definitions to the \emph{complete} semantics, and drop the letter $S$. Thus, ``issues'' in what follows refers to the equivalnce classes of \emph{in-sync} for the \emph{complete} semantics.}


\begin{definition}[Issue \cite{booth2012quantifying}]
Given the argumentation framework $\AF=\langle \calA, \rightharpoonup \rangle$, a set of arguments $\calB \subseteq \calA$ is called an issue iff it forms an equivalence class of the relation \emph{in-Sync ($\equiv$)}. 
\end{definition}

The Issue-wise set between two labelings $\calL_1$ and $\calL_2$ is the set of issues that these two labelings disagree upon.

\begin{definition}[Issue-wise Set $\varominus_\calW$]
Let $\calL_1$, $\calL_2$ be two labelings of $\AF=\langle \calA, \rightharpoonup \rangle$ and let $\calI$ be the set of all issues in $\AF$. We define the Issue-wise set between these two labelings as:
\begin{equation}
 \calL_1 \varominus_\calW \calL_2 = \{\calB \in \calI | \calL_1(A)\neq \calL_2(A) \text{ for some (equiv. all) } A \in \calB\}
\end{equation}
\end{definition}

Note that the sentence ``for some (equiv. all)'' follows from the definition of issues. One can realize that:

\begin{equation}
\forall \calL_1,\calL_2 \in \labs, \forall \calB \in \calI: (\exists A \in \calB \;s.t.\; \calL_1(A) \neq \calL_2(A) \Leftrightarrow \forall A \in \calB:\;  \calL_1(A) \neq \calL_2(A))
\end{equation}

The Issue-wise distance between two labelings $\calL_1$ and $\calL_2$ is the number of issues that these two labelings disagree upon.

\begin{definition}[Issue-wise Distance $\left |\varominus_\calW\right |$]
Let $\calL_1$, $\calL_2$ be two labelings of $\AF=\langle \calA, \rightharpoonup \rangle$. We define the Issue-wise distance between these two labelings as:
\begin{equation}
 \calL_1 \left |\varominus_\calW\right | \calL_2 = |\calL_1 \varominus_\calW \calL_2|
\end{equation}
\end{definition}

{ For example, in Figure \ref{fig:issues}, the Issue-wise sets between $\calL_1$ and the other two labellings are:
\[\calL_1 \varominus_\calW \calL_2=\{\{E,F\},\{G,H\}\}\]
\[\calL_1 \varominus_\calW \calL_3=\{\{A,B,C,D\}\}\]
 
While the corresponding Issue-wise distances are:
\[\calL_1 \left |\varominus_\calW\right | \calL_2=|\{\{E,F\},\{G,H\}\}|=2\]
\[\calL_1 \left |\varominus_\calW\right | \calL_3=|\{\{A,B,C,D\}\}|=1\] 
}

\subsubsection{Case 2: $\myundec$ is in the Middle between $\myin$ and $\myout$}~\\
In this section, we consider the case where $\myundec$ is in the middle between $\myin$ and $\myout$. Thus, we differentiate between two types of disagreement: 1) $\myin/\myout$ disagreement, and 2) $\mydec/\myundec$ disagreement. When considering distance, we assume $dist(\myin,\myout)= 2\times dist({\mydec},\myundec) = 2$. \footnote{ The use of $2$ here is chosen carefully to satisfy the triangle inequality. However, the use of any $\alpha$ s.t. $1<\alpha\leq 2$ would not affect the results of this paper. We just use $2$ here for simplicity.}

\begin{figure}[htbp]
    \centering
       \includegraphics[scale=0.4]{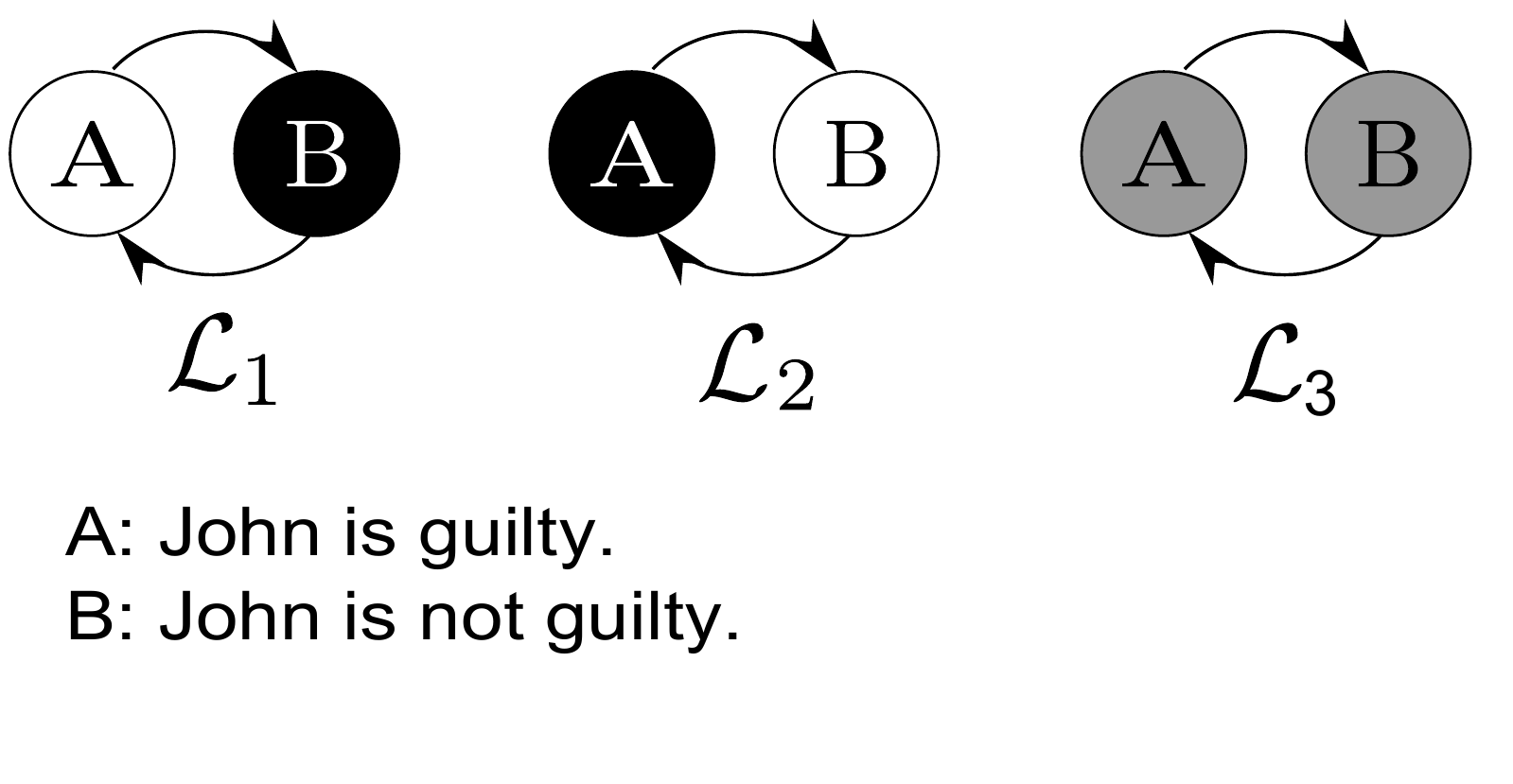}
    \caption{An example showing the need for considering $\myundec$ as a middle labeling between $\myin$ and $\myout$.}\label{fig:iuo}
\end{figure}

To illustrate the difference from the previous case, consider the example shown in Figure \ref{fig:iuo}. In this example, one can realize that the labelings $\calL_2$ and $\calL_3$ are equally distant from labeling $\calL_1$ when considering Hamming set/distance or Issue-wise set/distance. 

However, one can argue that $\calL_3$ is closer than $\calL_2$ to $\calL_1$. Consider the arguments in Figure \ref{fig:iuo}. Labelings $\calL_1$ and $\calL_2$ seem to be on completely different sides regarding their evaluations for $A$ and $B$. On the other hand, the difference between $\calL_1$ and $\calL_3$ is less drastic, because $\calL_3$ abstains from taking any position about $A$ and $B$.

We use IUO (short for In-Undec-Out i.e. Undec is in the middle) to denote this class of preferences. 

\paragraph{IUO Hamming Set{s} and IUO Hamming Distance}~\\
The $\myin-\myout$ Hamming set ($\varominus^{io}$) between two labelings $\calL_1$ and $\calL_2$ is the set of arguments that both labelings label as decided (i.e. $\myin$ or $\myout$), but on which {they} disagree upon. The $\mydec-\myundec$ Hamming set ($\varominus^{du}$) between two labelings $\calL_1$ and $\calL_2$ is the set of arguments that one of the two labelings labels as decided (whether $\myin$ or $\myout$) and the other labels as undecided.

\begin{definition}[IUO Hamming set{s} $\varominus^\calM$]
Let $\calL_1$, $\calL_2$ be two labelings of $\AF=\langle \calA, \rightharpoonup \rangle$. We define the IUO Hamming set{s} as a pair $\varominus^\calM = (\varominus^{io},\varominus^{du})$, where $\varominus^{io}$ is $\myin-\myout$ Hamming set and $\varominus^{du}$ is $\mydec-\myundec$ Hamming set:

\begin{equation}
\begin{aligned}[b]
\calL_1 \varominus^{io} \calL_2 = \{A \in \calA | (\calL_1(A)=\myin \wedge \calL_2(A)=\myout) \vee \\(\calL_1(A)=\myout \wedge \calL_2(A)=\myin)\}
\end{aligned}
\end{equation}

\begin{equation}
\begin{aligned}[b]
 \calL_1 \varominus^{du} \calL_2 = \{A \in \calA | (A \in \mydec(\calL_1) \wedge \calL_2(A)=\myundec) \vee \\(\calL_1(A)=\myundec \wedge A \in \mydec(\calL_2))\}
\end{aligned}
\end{equation}
where $\mydec(\calL_1)$ is the set of decided ($\myin$ or $\myout$) arguments according to the labeling $\calL_1$.
\end{definition}

The IUO Hamming distance between two labelings $\calL_1$ and $\calL_2$ is the number of arguments in $\calL_1 \varominus^{du} \calL_2$ added to twice the number of the arguments in $\calL_1 \varominus^{io} \calL_2$.
\begin{definition}[IUO Hamming Distance $\left |\varominus^\calM \right |$]
Let $\calL_1$, $\calL_2$ be two labelings of $\AF=\langle \calA, \rightharpoonup \rangle$. We define the IUO Hamming distance between these two labelings as:
\begin{equation}
 \calL_1 \left |\varominus^\calM \right | \calL_2 = 2 \times |\calL_1 \varominus^{io} \calL_2| + |\calL_1 \varominus^{du} \calL_2|
\end{equation}
\end{definition}

\paragraph{IUO Issue-wise Set{s} and IUO Issue-wise Distance}~\\
The $\myin-\myout$ Issue-wise set ($\varominus^{io}_\calW$) between two labelings $\calL_1$ and $\calL_2$ is the set of issues that both of the two labelings label as decided, but on which {they} disagree upon. The $\mydec-\myundec$ Issue-wise set ($\varominus^{du}_\calW$) between two labelings $\calL_1$ and $\calL_2$ is the set of issues that one of the two labelings labels as decided and the other labels as undecided.

\begin{definition}[IUO Issue-wise set{s} $\varominus^\calM_\calW$]
Let $\calL_1$, $\calL_2$ be two labelings of $\AF=\langle \calA, \rightharpoonup \rangle$ and let $\calI$ be the set of all issues in $\AF$. We define the IUO Issue-wise set{s} as $\varominus^\calM_\calW = (\varominus^{io}_\calW,\varominus^{du}_\calW)$, where $\varominus^{io}_\calW$ is the $\myin-\myout$ Issue-wise set and $\varominus^{du}_\calW$ is the $\mydec-\myundec$ Issue-wise set:

\begin{equation}
\begin{aligned}[b]
\calL_1 \varominus^{io}_\calW \calL_2 = \{\calB \in \calI | (\calL_1(A)=\myin \wedge \calL_2(A)=\myout) \vee \\(\calL_1(A)=\myout \wedge \calL_2(A)=\myin) \text{ for some (equiv. all) } A \in \calB\}
\end{aligned}
\end{equation}

\begin{equation}
\begin{aligned}[b]
 \calL_1 \varominus^{du}_\calW \calL_2 = \{\calB \in \calI | (A \in \mydec(\calL_1) \wedge \calL_2(A)=\myundec) \vee \\(\calL_1(A)=\myundec \wedge A \in \mydec(\calL_2)) \text{ for some (equiv. all) } A \in \calB\}
\end{aligned}
\end{equation}

\end{definition}

Note that given the definition of issues, for every labeling $\calL$, an issue is either decided (all arguments in it are labeled $\myin$ or $\myout$ by $\calL$) or undecided (all arguments in it are labeled undecided by $\calL$):

\begin{equation}
\forall \calL \in \labs, \forall \calB \in \calI: (\exists A \in \calB\; s.t.\; A \in \mydec(\calL) \Leftrightarrow \forall A \in \calB:\;   A \in \mydec(\calL))
\end{equation}

The IUO Issue-wise distance between two labelings $\calL_1$ and $\calL_2$ is the number of issues in $\calL_1 \varominus^{du}_\calW \calL_2$ added to twice the number of the issues in $\calL_1 \varominus^{io}_\calW \calL_2$.

\begin{definition}[IUO Issue-wise Distance $\left |\varominus^\calM_\calW\right |$]
Let $\calL_1$, $\calL_2$ be two labelings of $\AF=\langle \calA, \rightharpoonup \rangle$. We define the IUO Issue-wise distance between these two labelings as:
\begin{equation}
 \calL_1 \left |\varominus^\calM_\calW\right | \calL_2 = 2 \times |\calL_1 \varominus^{io}_\calW \calL_2| + |\calL_1 \varominus^{du}_\calW \calL_2|
\end{equation}
\end{definition}

{ For example, in Figure \ref{fig:iuo}, the IUO Issue-wise sets between $\calL_1$ and the other two labellings are:
\[\calL_1 \varominus^{io}_\calW \calL_2=\{\{A,B\}\},\calL_1 \varominus^{du}_\calW \calL_2=\{\}\]
\[\calL_1 \varominus^{io}_\calW \calL_3=\{\}, \calL_1 \varominus^{du}_\calW \calL_3=\{\{A,B\}\}\]
 
While the corresponding IUO Issue-wise distances are:
\[\calL_1 \left |\varominus^\calM_\calW\right | \calL_2=2\times|\{\{A,B\}\}|+0=2\]
\[\calL_1 \left |\varominus^\calM_\calW\right | \calL_3=2\times0+|\{\{A,B\}\}|=1\] 
}

Table \ref{tab:Prefs} summarizes the distance measures we consider.

\begin{table}[htbp]
  \centering
  
   \begin{tabular}{|c|c|c|c|}
    \hline
               \multicolumn{2}{|c|}{}                      & \textbf{Uniform}     & \textbf{IUO}  \\
      
    \hline
    \multirow{2}{*}{\textbf{Hamming}}    & \textbf{Set}    & Hamming Set {$\varominus$}         & IUO Hamming Set{s} {$\varominus^\calM$}\\
  \cline{2-4}
                                         &\textbf{Distance}& Hamming Distance  {$\left |\varominus\right |$}   & IUO Hamming Distance {$\left |\varominus^\calM\right |$}\\
    \hline
    \multirow{2}{*}{\textbf{Issue-wise}} & \textbf{Set}    & Issue-wise Set  {$\varominus_\calW$}     & IUO Issue-wise Set{s} {$\varominus^\calM_\calW$}\\
  \cline{2-4}
                                         &\textbf{Distance}& Issue-wise Distance {$\left |\varominus_\calW\right |$} & IUO Issue-wise Distance {$\left |\varominus^\calM_\calW\right |$} \\

    \hline

    \end{tabular}%
  \caption{Full family of distance measures.}\label{tab:Prefs}
\end{table}%

\subsection{Preferences}

Given the distance measures defined earlier, we define agents' preferences. { We say an agent's preferences are $x$-based, if her preferences are calculated using the distance measure $x$ (e.g. Hamming distance based preferences).} We use $\succeq_{i,x}$ to denote a \emph{weak preference} relation by agent $i$ whose preferences are $x$-based i.e. for any pair $\calL_1, \calL_2 \in \labs$, $\calL_1 \succeq_{i,x} \calL_2$ denotes that $\calL_1$ is more or equally preferred than $\calL_2$ by agent $i$ with $x$-based preferences. Further, we use $\succ_{i,x}$ to denote a \emph{strict preference} relation ($\calL_1 \succ_{i,x} \calL_2$ iff $(\calL_1 \succeq_{i,x} \calL_2) \wedge \neg(\calL_2 \succeq_{i,x} \calL_1)$), $\sim$ to denote an \emph{incomparability} relation ($\calL_1 \sim_{i,x} \calL_2$ iff $\neg(\calL_1 \succeq_{i,x} \calL_2) \wedge \neg(\calL_2 \succeq_{i,x} \calL_1)$), and $\cong$ to denote an \emph{indifference} relation ($\calL_1 \cong_{i,x} \calL_2$ iff $(\calL_1 \succeq_{i,x} \calL_2) \wedge (\calL_2 \succeq_{i,x} \calL_1)$).

We define the subset relation over pairs of sets as the following.
\begin{definition}[Subset Over Pairs $\subseteq$]
Let $A_1,A_2,B_1,B_2$ be four sets, and Let $S_1=(A_1,B_1)$, $S_2=(A_2,B_2)$ be two pairs of sets. We use $S_1 \subseteq S_2$ to denote the subset relation over pairs of subsets:
\begin{equation}
S_1 \subseteq S_2 \text{ iff } A_1 \subseteq A_2 \wedge B_1 \subseteq B_2
\end{equation} 
\end{definition}

{
Given a set measure $\otimes \in \{\varominus,\varominus^\calM,\varominus_\calW, \varominus^\calM_\calW\}$, an agent $i$, who has $\otimes$-set based preferences (and whose top preference is $\calL_i$), would prefer a labelling $\calL$ over another labeling $\calL'$ if and only if the set of arguments in $\calL_i \otimes \calL$ is a subset of $\calL_i \otimes \calL'$ (where ``subset'' here refers to the standard definition of subset as well as the definition of ``subset over pairs'' defined above).} Note that the set based preference yields a partial order over the labelings.\footnote{Although formally, the set-based criteria are not measures but mappings to sets, we will slightly abuse terminology and refer to all criteria (set based and distance based) as \emph{set and distance measures} for easy reference.} 

\begin{definition}[Set Based Preference $\succeq_{i,\otimes}$]
We say that agent $i$'s preferences are $\otimes$-set based { w.r.t $\calL_i$} iff:
\begin{equation}
\forall \calL,\calL' \in \labs: \calL \succeq_{i,\otimes} \calL' \Leftrightarrow \calL \otimes \calL_i \subseteq \calL' \otimes \calL_i  
\end{equation}
where $\calL_i$ is agent $i$'s most preferred labeling and $\otimes \in \{\varominus,\varominus^\calM,\varominus_\calW, \varominus^\calM_\calW\}$. Note that $\otimes$-set based preferences is read Hamming set based preferences when $\otimes = \varominus$, Issue-wise set based preferences when $\otimes = \varominus_\calW$, $\ldots\etc$
\end{definition}

{
Given a distance measure $\left|\otimes\right| \in \{\left|\varominus\right|,\left|\varominus^\calM\right|,\left|\varominus_\calW\right|, \left|\varominus^\calM_\calW\right|\}$, an agent $i$, who has $\left|\otimes\right|$-distance based preferences (and whose top preference is $\calL_i$), would prefer a labelling $\calL$ over another labeling $\calL'$ if and only if $\calL_i \left|\otimes\right| \calL$ is less than $\calL_i \left|\otimes\right| \calL'$.} Note that the distance based preference yields a total pre-order over the labelings.

{ We now define the classes of preferences which are based on different distance measures, that we defined earlier.}
 
\begin{definition}[Distance Based Preference $\succeq_{i,\left|\otimes\right|}$]
We say that agent $i$'s preferences are $\left|\otimes\right|$-distance based { w.r.t $\calL_i$} iff:
\begin{equation}
\forall \calL,\calL' \in \labs: \calL \succeq_{i,\left |\otimes\right |} \calL' \Leftrightarrow \calL \left |\otimes\right | \calL_i \leq \calL' \left |\otimes\right | \calL_i \end{equation}
where $\calL_i$ is agent $i$'s most preferred labeling and $\left|\otimes\right| \in \{\left|\varominus\right|,\left|\varominus^\calM\right|,\left|\varominus_\calW\right|, \left|\varominus^\calM_\calW\right|\}$. Note that $\left|\otimes\right|$-distance based preferences is read Hamming distance based preferences when $\left|\otimes\right| = \left|\varominus\right|$, Issue-wise distance based preferences when $\left|\otimes\right| = \left|\varominus_\calW\right|$, $\ldots\etc$

\end{definition}

\begin{figure}[htbp]
    \centering
       \includegraphics[scale=0.6]{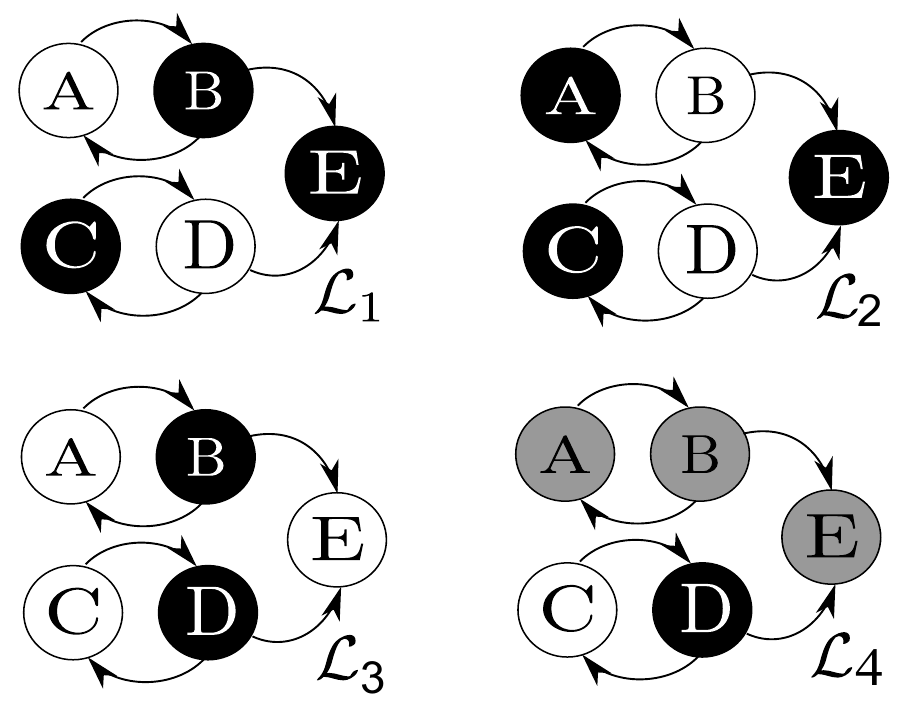}
    \caption{An argumentation graph with four possible complete labelings.}
    \label{fig:hamm}
\end{figure}

To illustrate the set and distance based preferences, we use Hamming set and Hamming distance based preferences for their simplicity. Consider the example in Figure \ref{fig:hamm} with four possible complete labelings. The Hamming sets between $\calL_1$ and the other three labelings are:
\[\calL_1 \varominus \calL_2 = \{A,B\}\]
\[\calL_1 \varominus \calL_3 = \{C,D,E\}\]
\[\calL_1 \varominus \calL_4 = \{A,B,C,D,E\}\]

Consequently, the Hamming distance values between $\calL_1$ and the other three labelings are the cardinality values of the Hamming sets between $\calL_1$ and the other three labelings.

\[\calL_1 \left |\varominus\right | \calL_2 = |\{A,B\}| = 2\]
\[\calL_1 \left |\varominus\right | \calL_3 = |\{C,D,E\}| = 3\]
\[\calL_1 \left |\varominus\right | \calL_4 = |\{A,B,C,D,E\}| = 5\]

Assume we have agents with Hamming set based preferences. Hence, an arbitrary agent $i$ who prefers $\calL_1$ the most, would have the following preferences: {$\calL_1 \succ$ $\calL_2 \succ$ $\calL_4$ and $\calL_1 \succ$ $\calL_3 \succ$ $\calL_4$} (neither $\calL_1 \varominus\ \calL_2$ nor $\calL_1 \varominus\ \calL_3$ is a subset of the other). However, if agents have Hamming distance based preferences, an agent who prefers $\calL_1$ the most, would have the following preferences: $\calL_1 \succ$ $\calL_2 \succ$ $\calL_3 \succ$ $\calL_4$.

We can now {examine} the examples in Figure \ref{fig:issues} and \ref{fig:iuo} in the light of preferences. The example in Figure \ref{fig:issues} shows how an agent $i$ whose top preference is $\calL_i=\calL_1$ would have different opinions about other labelings given the different distance measures used. If agent $i$ has Hamming distance based preferences, {then:
\[\calL_1 \left |\varominus\right | \calL_2=|\{E,F,G,H\}|=4\]
\[\calL_1 \left |\varominus\right | \calL_3=|\{A,B,C,D\}|=4\] 
}
then, her preferences would be $\calL_1 \succ_{i,\left |\varominus\right |} \calL_2 \cong_{i,\left |\varominus\right |} \calL_3$, while if she has Issue-wise distance based preferences, {then:
\[\calL_1 \left |\varominus_\calW\right | \calL_2=|\{\{E,F\},\{G,H\}\}|=2\]
\[\calL_1 \left |\varominus_\calW\right | \calL_3=|\{\{A,B,C,D\}\}|=1\] 
}
then, her preferences would be $\calL_1 \succ_{i,\left |\varominus_\calW\right |} \calL_3 \succ_{i,\left |\varominus_\calW\right |} \calL_2$. Hence, it is interesting to introduce the ``Issue-wise'' concept to define a new class of preferences.

The example in Figure \ref{fig:iuo} shows how an agent $i$ (whose top preference is $\calL_1$) would have different preferences depending on whether $\myin$, $\myout$, and $\myundec$ are equally distant, or $\myundec$ is in the middle between $\myin$ and $\myout$. {In the former case: 
\[\calL_1 \left |\varominus\right | \calL_2=|\{A,B\}|=2\]
\[\calL_1 \left |\varominus\right | \calL_3=|\{A,B\}|=2\] 
and 
\[\calL_1 \left |\varominus_\calW\right | \calL_2=|\{\{A,B\}\}|=1\]
\[\calL_1 \left |\varominus_\calW\right | \calL_3=|\{\{A,B\}\}|=1\]

hence, $\calL_1 \succ_{i,\left |\varominus\right |} \calL_2 \cong_{i,\left |\varominus\right |} \calL_3$ and $\calL_1 \succ_{i,\left |\varominus_\calW\right |} \calL_2 \cong_{i,\left |\varominus_\calW\right |} \calL_3$. In the latter case:
\[\calL_1 \left |\varominus^\calM\right | \calL_2=2\times|\{A,B\}|+0=4\]
\[\calL_1 \left |\varominus^\calM\right | \calL_3=2\times0+|\{A,B\}|=2\] 
and
\[\calL_1 \left |\varominus^\calM_\calW\right | \calL_2=2\times|\{\{A,B\}\}|+0=2\]
\[\calL_1 \left |\varominus^\calM_\calW\right | \calL_3=2\times0+|\{\{A,B\}\}|=1\]

hence, $\calL_1 \succ_{i,\left |\varominus^\calM\right |} \calL_3 \succ_{i,\left |\varominus^\calM\right |} \calL_2$ and $\calL_1 \succ_{i,\left |\varominus^\calM_\calW\right |} \calL_3 \succ_{i,\left |\varominus^\calM_\calW\right |} \calL_2$.} Thus, it is interesting to define Hamming and Issue based preferences with $\myundec$ as a middle labeling.

\section{Pareto Optimality}\label{sec.Pa}
In this section, we study the Pareto optimality of the outcomes of the three operators given different variations of the preferences. Pareto optimality is one of the fundamental concepts that ensures that, given a profile, the social outcome selected by the aggregation procedure cannot be improved. 

A labeling $\calL_1$ Pareto dominates $\calL_2$ if and only if for any agent $i$, $i$ would prefer $\calL_1$ at least as much as she prefers $\calL_2$, and for at least one agent $j$, $j$ would strictly prefer $\calL_1$ over $\calL_2$.

\begin{definition}[Pareto dominance]
Let $\Ag = \{1,\ldots,n\}$ be a set of agents with preferences $\succeq_i,$ $i \in \Ag$. $\calL$ Pareto dominates $\calL'$ iff $\forall i \in \Ag,$ $\calL \succeq_i \calL'$ and $\exists j \in \Ag,$ $\calL \succ_j \calL'$.
\end{definition}

{
A labeling is Pareto optimal in a set, if it is not Pareto dominated by any other labeling from that set.


\begin{definition}[Pareto optimality {of a labeling} in $\calS$]
Let $\calS$ be a set of labelings. A labeling $\calL$ is Pareto optimal in $\calS$ if there is no labeling $\calL' \in \calS$ such that $\calL'$  Pareto dominates $\calL$.
\end{definition}

In our results, the set ${\calS}$ will mainly refer to a set of \emph{admissible} (or \emph{complete}) labelings that are \emph{compatible with} (or \emph{smaller or equal to}) each of the participants' labelings. Moreover, whenever we refer to an operator as Pareto optimal (in a set $\calS$) we mean that it only produces Pareto optimal outcomes (in $\calS$).
}

{
\begin{definition}[Pareto optimality of an operator in $\calS$]
Let $\calS$ be a set of labelings. An operator is Pareto optimal in $\calS$ if it only produces Pareto optimal (in $\calS$) outcomes.
\end{definition}
}

{
\subsection{Connections between Classes of Preferences}

\subsubsection{General Connections}~\\
{

We start our analysis by noticing that some of the defined types are in fact equivalent. Consider the following lemma which establishes the equivalence between some types of preferences:

\begin{lemma}\label{lem.heterosets}
Issue-wise set based preferences coincide with Hamming set based preferences, and IUO Issue-wise set{s} based preferences coincide with IUO Hamming set{s} based preferences.
\end{lemma}

Further, we notice that Pareto optimality carries over from each of the distance-based preferences to its corresponding set-based preferences.

\begin{theorem}\label{thm.HdHs}
Let $\otimes \in \{\ominus,\ominus^\calM,\ominus_\calW,\ominus^\calM_\calW\}$ be a set measure and $|\otimes|$ be its corresponding distance measure (i.e. if $\otimes = \ominus^\calM$ then $|\otimes|=|\ominus^\calM|$). If a labeling\footnote{Note that since an operator is Pareto optimal in a set if and only if all of its outcomes are Pareto optimal in that set, then one can see that in this theorem, and others as well, `labeling' can be substituted with `operator'.} is Pareto optimal in a set $\calS$ given agents with $|\otimes|$-based preferences, then it is Pareto optimal in $\calS$ given agents with $\otimes$-based preferences.  
\end{theorem}
}

Unfortunately, these connections are only one-way. A counterexample for the opposite way is provided in the Appendix.

\subsubsection{Connections in Special Contexts}~\\
{
The implications shown in the previous part hold without restrictions. However, when all labelings in $\calS$ are admissible labelings and are compatible ($\approx$) with each of the individuals' labelings, one can find further connections.
}

\begin{theorem}\label{thm.MHaPaSo}
Let $\calX$ be the set of all admissible labelings that are compatible ($\approx$) with each of the participants' individual labelings. Let $\calS$ be any arbitrary set such that $\calS \subseteq \calX$. A labeling from $\calS$ is Pareto optimal in $\calS$ when individual preferences are Hamming set (resp. distance) based iff it is Pareto optimal in $\calS$ when individual preferences are IUO Hamming set{s} (resp. distance) based.
\end{theorem}

\begin{theorem}\label{thm.MIwPaSo}
Let $\calX$ be the set of all admissible labelings that are compatible ($\approx$) with each of the participants' individual labelings. Let $\calS$ be any arbitrary set such that $\calS \subseteq \calX$. A labeling from $\calS$ is Pareto optimal in $\calS$ when individual preferences are Issue-wise set (resp. distance) based iff it is Pareto optimal in $\calS$ when individual preferences are IUO Issue-wise set{s} (resp. distance) based.
\end{theorem}
{
Note that unlike in the previous part where connections are one-way (from distance based to set based, but not vice versa), the connections in this part hold in both directions.
} 
\subsubsection{Failed Connections}~\\
{
Given the findings so far, one might wonder about the existence of other connections among the eight classes of preferences. Unfortunately, other than the ones found above, there exists no connection, even after considering further restrictions, similar to the ones in the previous part. In the Appendix, we provide counterexamples for the connections that do not hold between the classes of preferences.

Basically, we provide an example for an argumentation framework in which Pareto optimality is satisfied when agents' preferences are Hamming set based, Issue-wise set based, Issue-wise distance based, IUO Hamming set based, IUO Issue-wise set based, and IUO Issue-wise distance based. However, Pareto optimality is not satisfied when agents' preferences are Hamming distance based, or IUO Hamming distance based. This shows that Pareto optimality given Hamming distance based preferences and IUO Hamming distance based preferences cannot be inferred from the other six classes of preferences. 

In a similar way, we provide an example for an argumentation framework which shows that Pareto optimality given Issue-wise distance based preferences and IUO Issue-wise distance based preferences cannot be inferred from the other six classes of preferences. 
}

We summarize all the findings in Table \ref{tab:PaG}. For each cell in the table, a $Y$ means that for any operator, Pareto optimality in any arbitrary set $\calS$ carries over from the preference class in the row to the preference class in the column (in the same set $\calS$), while a $Y^*$ means that it only holds for compatible operators (i.e. that produce compatible outcomes).

\begin{table}[htbp]
  \centering
	\footnotesize
	  
   \begin{tabular}{|c|c|c|c|c|c|c|}
    \hline
          & \multirow{2}{*}{\textbf{HS/IwS}} & \multirow{2}{*}{\textbf{HD}} & \multirow{2}{*}{\textbf{IwD}} & \textbf{IUO} & \textbf{IUO} & \textbf{IUO} \\
          &  &  &  & \textbf{HS/IwS} & \textbf{HD} & \textbf{IwD} \\
    \hline
    \textbf{Hamming set (HS)}   &Y&N&N&Y*&N&N\\
    \textbf{Issue-wise set (IwS)}&&&&&&\\
    \hline
    \textbf{Hamming dist. (HD)} &Y&Y&N&Y*&Y*&N\\
    \hline
    \textbf{Issue-wise dist. (IwD)} &Y&N&Y&Y*&N&Y*\\
    \hline
    \textbf{IUO Hamming set{s} (IUO HS)}   &Y*&N&N&Y&N&N\\
    \textbf{IUO Issue-wise set{s} (IUO IwS)} &&&&&&\\
    \hline
    \textbf{IUO Hamming dist. (IUO HD)} &Y*&Y*&N&Y&Y&N\\
    \hline
    \textbf{IUO Issue-wise dist. (IUO IwD)} &Y*&N&Y*&Y&N&Y\\
    \hline
    \end{tabular}%
  \caption{Pareto optimality relations between the different preference classes. A $Y$ means Pareto optimality carries over from the class in the row to the class in the column, and a $Y^*$ means it only carries over if the operator only produces compatible labelings.}\label{tab:PaG}
\end{table}%

Now we turn to studying the Pareto optimality of the three operators: the skeptical, the credulous and the super credulous, {with respect to the eight classes of preferences.}
}
\subsection{Case 1: $\myin$, $\myout$, and $\myundec$ are Equally Distant from Each Other}\label{sec.par.ED}
\subsubsection{Hamming Set and Hamming Distance}
{
In this part, we establish the first advantage of the skeptical operator over the credulous and super credulous operators. When all individuals' preferences are Hamming set based, or all are Hamming distance based, the skeptical operator is Pareto optimal in the set of admissible labelings that are smaller or equal ($\sqsubseteq$) to each individual's labeling. 
}

\begin{theorem}\label{thm.HaPaSoSet}
If individual preferences are Hamming set based, then the skeptical aggregation operator is Pareto optimal {in the set of} admissible labelings that are smaller or equal ($\sqsubseteq$) to each of the participants' labelings.
\end{theorem}

\begin{theorem}\label{thm.HaPaSoDis}
If individual preferences are Hamming distance based, then the skeptical aggregation operator is Pareto optimal {in the set of} admissible labelings that are smaller or equal ($\sqsubseteq$) to each of the participants' labelings.
\end{theorem}

{ On the other hand, the credulous and super credulous operators are only Pareto optimal when individuals have Hamming set based preferences, and they fail to produce Pareto optimal outcomes when the preferences are Hamming distance based.}

\begin{theorem}\label{thm.HaPaCoSet}
If individual preferences are Hamming set based, then the credulous aggregation operator is Pareto optimal {in the set of} admissible labelings that are compatible ($\approx$) with each of the participants' labelings.
\end{theorem}

\begin{theorem}\label{thm.HaPaScoSet}
If individual preferences are Hamming set based, then the super credulous aggregation operator is Pareto optimal {in the set of} complete labelings that are compatible ($\approx$) with each of the participants' labelings.
\end{theorem}

\begin{observation}\label{obs.HaPaCoDis}
If individual preferences are Hamming distance based, then the credulous (resp. the super credulous) aggregation operator is not Pareto optimal {in the set of} admissible (resp. complete) labelings that are compatible ($\approx$) with each of the participants' labelings. \comment{An example is given in Figure \ref{fig:HSco} where $\calL_{CO}$ represents the outcome of the credulous (or the super credulous) aggregation operator. Both labelings $\calL_{CO}$ and $\calL_X$ are compatible with both $\calL_1$ and $\calL_2$, but $\calL_X$ is closer when applying Hamming distance. $\calL_1 \varominus \calL_{CO} = \calL_2 \varominus \calL_{CO} = \{A, B, E, F, G\}$, so the Hamming distance is 5, whereas $\calL_1 \varominus \calL_X = \calL_2 \varominus \calL_X = \{A, B, C, D\}$, so the Hamming distance is 4.
}
\end{observation}
\comment{
\begin{figure}[ht]
    \centering
  \includegraphics[scale=0.4]{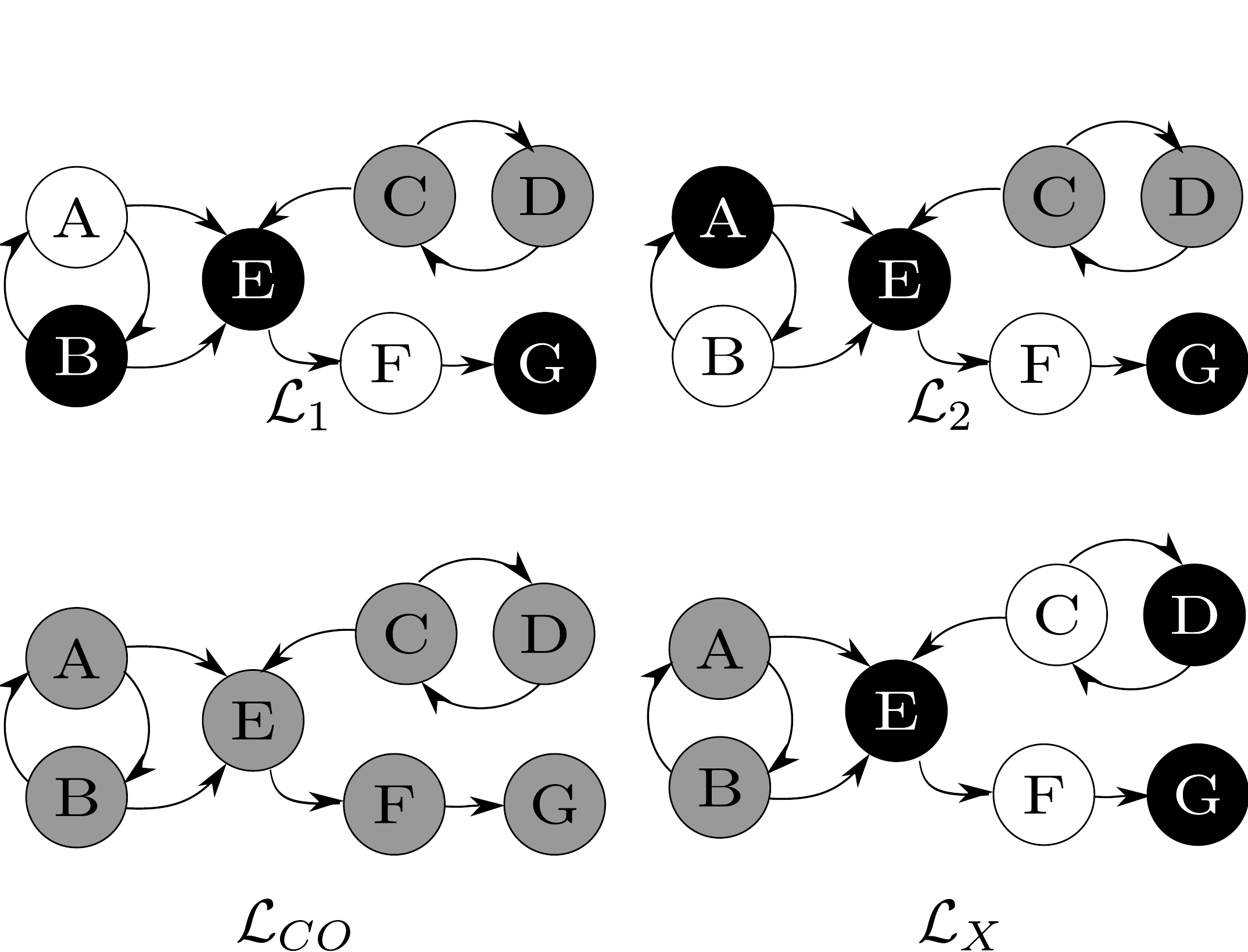}
  \caption{If individuals' preferences are Hamming distance based, the (super) credulous aggregation operator is not Pareto optimal {in the set of admissible (resp. complete) labelings that are compatible ($\approx$) with each of the participants' labelings.}}\label{fig:HSco}
\end{figure}
}

\subsubsection{Issue-wise Set and Issue-wise Distance}~\\
{

Given Lemma \ref{lem.heterosets}, we can substitute ``Hamming set'' with ``Issue-wise set'' in Theorem \ref{thm.HaPaSoSet}, Theorem \ref{thm.HaPaCoSet} and Theorem \ref{thm.HaPaScoSet}.

Unfortunately, Lemma \ref{lem.heterosets} only concerns the implication from Hamming set to Issue-wise set based preferences and vice versa. One might wonder if a similar lemma can be shown for the case with Hamming distance and Issue-wise distance based preferences. However, we show in Examples 1 and 2 in the Appendix, that this is not necessarily the case. As such, one has to show whether Pareto optimality results hold or not for Issue-wise distance based preferences independently from those of the Hamming distance based preferences.

Doing so confirms the superiority of the skeptical operator for the Issue-wise distance-based preferences. As we show next, when agents preferences are Issue-wise distance based, only the skeptical aggregation operator is guaranteed to produce Pareto optimal outcomes. 

}
\begin{theorem}\label{thm.IwPaSoDis}
{If individual preferences are Issue-wise distance based, then the skeptical aggregation operator is Pareto optimal in the set of admissible labelings that are smaller or equal (w.r.t $\sqsubseteq$) to each of the participants' labelings.}
\end{theorem}

\begin{observation}\label{obs.IwPaCoSco}
If individual preferences are Issue-wise distance based, then the credulous (resp. the super credulous) aggregation operator is not Pareto optimal {in the set of} admissible (resp. complete) labelings that are compatible ($\approx$) with each of the participants' labelings. \comment{In Figure \ref{fig:CoSco}, $\calL_{CO}$ represents the outcome of the credulous (or the super credulous) aggregation operator. Note that, both labelings of $\calL_{CO}$ and $\calL_X$ are compatible with both $\calL_1$ and $\calL_2$, but $\calL_X$ is closer when applying Issue-wise distance. $\calL_1 \varominus_\calW \calL_{CO}=$ $\calL_2 \varominus_\calW \calL_{CO}=$ $\{\{C,D\},\{E\},\{F\}\}$, so Issue-wise distance is $3$, whereas $\calL_1 \varominus_\calW \calL_X=$ $\calL_2 \varominus_\calW \calL_{X}=$ $\{\{A,B\},\{C,D\}\}$, so Issue-wise distance is $2$.
}
\end{observation}

\comment{
\begin{figure}[ht]
    \centering
  \includegraphics[scale=0.4]{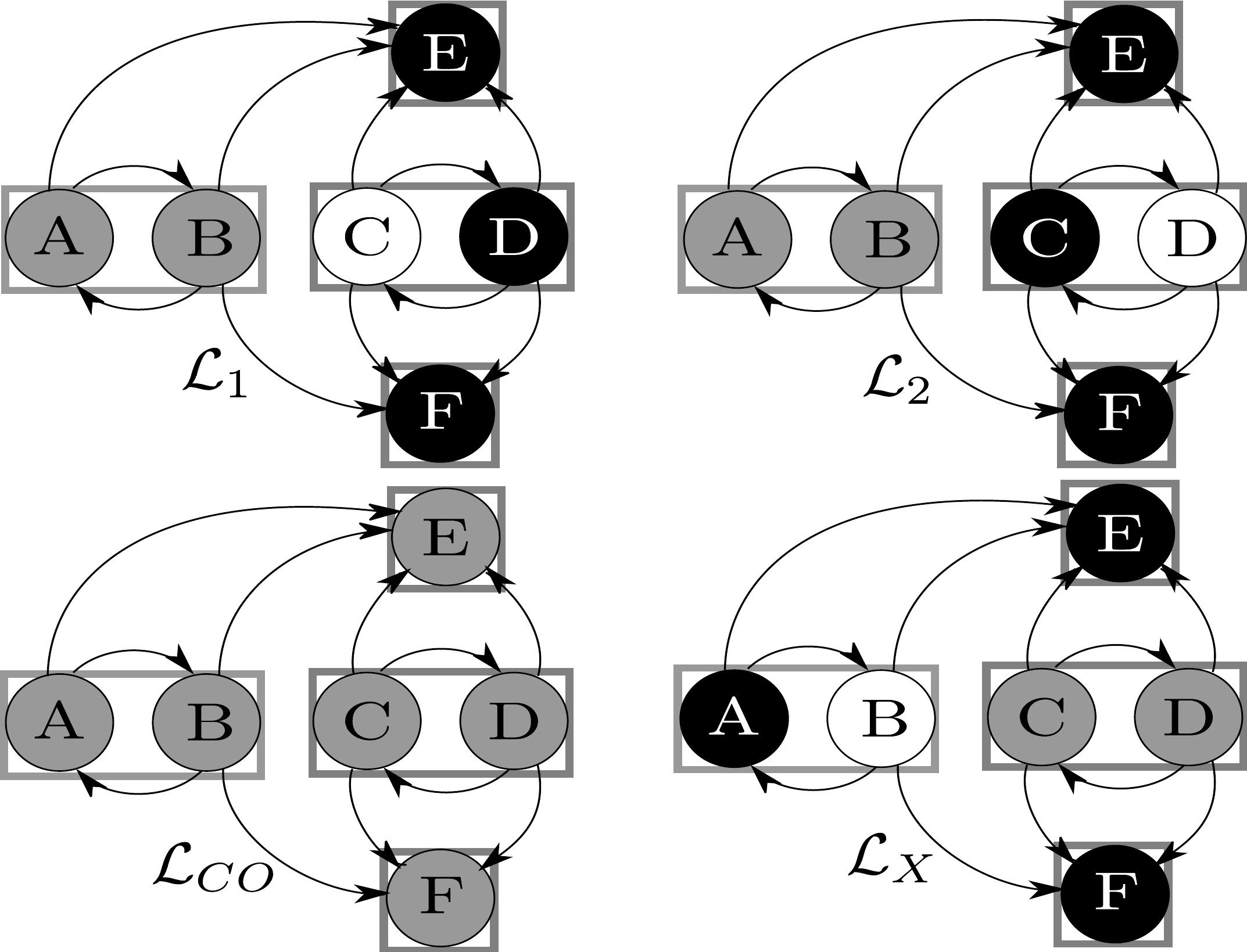}
  \caption{If individuals' preferences are Issue-wise distance based, the (super) credulous aggregation operator is not Pareto optimal {in the set of admissible (resp. complete) labelings that are compatible ($\approx$) with each of the participants' labelings.} The set of arguments located in one box form an issue.}\label{fig:CoSco}
\end{figure}

}

\subsection{Case 2: $\myundec$ is in the Middle between $\myin$ and $\myout$}\label{ssect.paiuo}

We now analyze the Pareto optimality for the three operators given the classes of preferences that assume $\myundec$ to be in the middle between $\myin$ and $\myout$ ($dist(\mydec,\myundec)$ $< dist (\myin,\myout)$). We show that for the three considered operators, it is possible to show an equivalence with the results of Section \ref{sec.par.ED}. 

\subsubsection{IUO Hamming Set{s} and IUO Hamming Distance}~\\
IUO Hamming set{s} (resp. IUO Hamming distance) differs from the Hamming set (resp. Hamming distance) in that the former separates $\myin/\myout$ disagreement from $\mydec/\myundec$ disagreement. {We use Theorem \ref{thm.MHaPaSo} to show the results for this part.}

{
\begin{proposition}\label{prop.MHaPaSo}
If individual preferences are IUO Hamming set{s} (resp. distance) based, then the skeptical aggregation operator is Pareto optimal {in the set of} admissible labelings that are smaller or equal ($\sqsubseteq$) to each of the participants' labelings.
\comment{
\begin{proof}
Let $\calS$ be the set of admissible labelings that are smaller or equal ($\sqsubseteq$) to each of the participants' labelings. Then, $\calS \subseteq \calX$, where $\calX$ is defined in Theorem \ref{thm.MHaPaSo}. From Theorem \ref{thm.HaPaSoSet} and Theorem \ref{thm.MHaPaSo} the skeptical aggregation operator is Pareto optimal in $\calS$ when individual preferences are IUO Hamming set{s} based. From Theorem \ref{thm.HaPaSoDis} and Theorem \ref{thm.MHaPaSo} the skeptical aggregation operator is Pareto optimal in $\calS$ when individual preferences are IUO Hamming distance based.
\end{proof}
}
\end{proposition}

{As for the credulous and super credulous operators, their results given IUO Hamming set and distance based preferences echo their results with the Hamming set and distance based preferences. Both credulous and super credulous produce Pareto optimal outcomes given Hamming set based preferences, but can fail to produce Pareto optimal outcomes given Hamming distance based preferences.}

\begin{proposition}\label{prop.MHaPaSoSet}
If individual preferences are IUO Hamming set{s} based, then the credulous aggregation operator is Pareto optimal {in the set of} admissible labelings that are compatible ($\approx$) with each of the participants' labelings.
\comment{
\begin{proof}
Let $\calS$ be the set of admissible labelings that are compatible ($\approx$) with each of the participants' labelings. Then, $\calS \subseteq \calX$, where $\calX$ is defined in Theorem \ref{thm.MHaPaSo} (actually $\calS = \calX$ here). From Theorem \ref{thm.HaPaCoSet} and Theorem \ref{thm.MHaPaSo} the credulous aggregation operator is Pareto optimal in $\calS$ when individual preferences are IUO Hamming set{s} based.
\end{proof}
}
\end{proposition}

\begin{proposition}\label{prop.MHaPaSoDis}
If individual preferences are IUO Hamming distance based, then the credulous aggregation operator is not Pareto optimal {in the set of} admissible labelings that are compatible ($\approx$) with each of the participants' labelings.
\comment{
\begin{proof}
Similar to the previous proposition, from Observation \ref{obs.HaPaCoDis} and Theorem \ref{thm.MHaPaSo} the credulous aggregation operator is not Pareto optimal in $\calS$ ($\calS$ is defined in the previous proposition) when individual preferences are IUO Hamming distance based.
\end{proof}
}
\end{proposition}

\begin{proposition}\label{prop.MHaPaScoSet}
If individual preferences are IUO Hamming set{s} based, then the super credulous aggregation operator is Pareto optimal {in the set of} complete labelings that are compatible ($\approx$) with each of the participants' labelings.
\comment{
\begin{proof}
Let $\calS$ be the set of complete labelings that are compatible ($\approx$) with each of the participants' labelings. Then, $\calS \subseteq \calX$, where $\calX$ is defined in Theorem \ref{thm.MHaPaSo}. From Theorem \ref{thm.HaPaScoSet} and Theorem \ref{thm.MHaPaSo} the super credulous aggregation operator is Pareto optimal in $\calS$ when individual preferences are IUO Hamming set{s} based.
\end{proof}
}
\end{proposition}

\begin{proposition}\label{prop.MHaPaScoDis}
If individual preferences are IUO Hamming distance based, then the super credulous aggregation operator is not Pareto optimal {in the set of} complete labelings that are compatible ($\approx$) with each of the participants' labelings.
\comment{
\begin{proof}
Similar to the previous proposition, from Observation \ref{obs.HaPaCoDis} and Theorem \ref{thm.MHaPaSo} the super credulous aggregation operator is not Pareto optimal in $\calS$ ($\calS$ is defined in the previous proposition) when individual preferences are IUO Hamming distance based.
\end{proof}
}
\end{proposition}
}
\subsubsection{IUO Issue-wise Set{s} and IUO Issue-wise Distance}~\\

{
Again, given the discussion earlier as a result of Lemma \ref{lem.heterosets} (i.e. IUO Issue-wise set{s} based preferences coincide with IUO Hamming set{s} based preferences), we can substitute ``IUO Hamming set{s}'' with ``IUO Issue-wise set{s}'' in Proposition \ref{prop.MHaPaSo}, Proposition \ref{prop.MHaPaSoSet} and Proposition \ref{prop.MHaPaScoSet}. 

Next, we show the results for the IUO Issue-wise distance based preferences.

\begin{proposition}\label{prop.MIwPaSo}
If individual preferences are IUO Issue-wise distance based, then the skeptical aggregation operator is Pareto optimal {in the set of} admissible labelings that are smaller or equal ($\sqsubseteq$) to each of the participants' labelings.
\comment{
\begin{proof}
Let $\calS$ be the set of admissible labelings that are smaller or equal ($\sqsubseteq$) to each of the participants' labelings. Then, $\calS \subseteq \calX$, where $\calX$ is defined in Theorem \ref{thm.MIwPaSo}. From Theorem \ref{thm.IwPaSoDis} and Theorem \ref{thm.MIwPaSo} the skeptical aggregation operator is Pareto optimal in $\calS$ when individual preferences are IUO Issue-wise distance based.
\end{proof}
}
\end{proposition}

We now turn to the credulous aggregation operator.

\begin{proposition}\label{prop.MIwPaSoDis}
If individual preferences are IUO Issue-wise distance based, then the credulous aggregation operator is not Pareto optimal {in the set of} admissible labelings that are compatible ($\approx$) with each of the participants' labelings.
\comment{
\begin{proof}
Similar to the previous proposition, from Observation \ref{obs.IwPaCoSco} and Theorem \ref{thm.MIwPaSo} the credulous aggregation operator is not Pareto optimal in $\calS$ ($\calS$ is defined in the previous proposition) when individual preferences are IUO Issue-wise distance based.
\end{proof}
}
\end{proposition}

Finally, we turn to the super credulous aggregation operator.

\begin{proposition}\label{prop.MIwPaScoDis}
If individual preferences are IUO Issue-wise distance based, then the super credulous aggregation operator is not Pareto optimal {in the set of} complete labelings that are compatible ($\approx$) with each of the participants' labelings.
\comment{
\begin{proof}
Similar to the previous proposition, from Observation \ref{obs.IwPaCoSco} and Theorem \ref{thm.MIwPaSo} the super credulous aggregation operator is not Pareto optimal in $\calS$ ($\calS$ is defined in the previous proposition) when individual preferences are IUO Issue-wise distance based.
\end{proof}
}
\end{proposition}

}

Table \ref{tab:Pa} summarizes the Pareto optimality results for the three operators given all the eight classes of preferences.

\begin{table}[htbp]
  \centering
  
   \begin{tabular}{|c|c|c|c|}
    \hline
          & \textbf{Skeptical} & \textbf{Credulous} & \textbf{Super Credulous} \\
          & \textbf{Operator} & \textbf{Operator} & \textbf{Operator} \\
    \hline
    \textbf{Hamming set}   & Yes (Thm. \ref{thm.HaPaSoSet})  & Yes (Thm. \ref{thm.HaPaCoSet}) & Yes (Thm. \ref{thm.HaPaScoSet})\\
    \textbf{Issue-wise set} &&&\\
    \hline
    \textbf{Hamming dist.} & Yes (Thm. \ref{thm.HaPaSoDis})  & No (Obs. \ref{obs.HaPaCoDis})  & No (Obs. \ref{obs.HaPaCoDis})\\
    \hline
    \textbf{Issue-wise dist.} & Yes (Thm. \ref{thm.IwPaSoDis})  & No (Obs. \ref{obs.IwPaCoSco})  & No (Obs. \ref{obs.IwPaCoSco})\\
    \hline

    \textbf{IUO Hamming set{s}}   & Yes (Prop. \ref{prop.MHaPaSo})  & Yes (Prop. \ref{prop.MHaPaSoSet}) & Yes (Prop. \ref{prop.MHaPaScoSet})\\
    \textbf{IUO Issue-wise set{s}}&&&\\
    \hline
    \textbf{IUO Hamming dist.} & Yes (Prop. \ref{prop.MHaPaSo})  & No (Prop. \ref{prop.MHaPaSoDis})  & No (Prop. \ref{prop.MHaPaScoDis})\\
    \hline
    \textbf{IUO Issue-wise dist.} & Yes (Prop. \ref{prop.MIwPaSo})  & No (Prop. \ref{prop.MIwPaSoDis})  & No (Prop. \ref{prop.MIwPaScoDis})\\
    \hline

    \end{tabular}%
  \caption{Pareto optimality {in a set ${\calS}$} of the aggregation operators depending on the type of preference. {The set $\calS$ differs for each operator. For the skeptical operator, it is the set of all admissible labelings that are smaller or equal ($\sqsubseteq$) to each of the participants' labelings. For the credulous (resp. super credulous) operator, it is the set of all admissible (resp. complete) labelings that are compatible ($\approx$) with each of the participants' labelings.}}\label{tab:Pa}
\end{table}%

{\color{black}
\subsection{Heterogeneous Preferences}\label{subs.hetero}
The previous subsections have all considered the case where agents have homogeneous preferences i.e. agents share the same class of preferences (e.g. all agents have Hamming set based preferences). However, there can be some scenarios where this assumption does not hold. In this part, we study the effect of removing this assumption.

Let $\calF$ be the set of all classes of preferences, $\calR$ be some arbitrary subset of $\calF$, and $c:\Ag \rightarrow \calF$ be a function defining the class of preferences for each agent. We say that the set of agents $\Ag$ have \emph{homogeneous} preferences from $\calR$ if $\forall i,j \in \Ag: c(i)=c(j) \in \calR$. We say $\Ag$ have \emph{heterogeneous} preferences from $\calR$ if $\forall i \in \Ag : c(i) \in \calR$ and $\exists i,j \in \Ag$ s.t. $c(i) \neq c(j)$.
 
Let $\calR$ be an arbitrary set of classes of preferences. In general, if a labeling $\calL$ is Pareto optimal in a set $\calS$ given that $\Ag$ have \emph{homogeneous} preferences from $\calR$, then $\calL$ might not be Pareto optimal if $\Ag$ have \emph{heterogeneous} preferences from $\calR$ (see Example 3 in the Appendix).

{
However, one can show that some of the classes of preferences that we defined enjoy special relations with each others that make Pareto optimality carry over from homogeneous preference of each of those classes to heterogeneous preferences that combine all of those classes. Consider the following theorem.
}

\begin{theorem}
Let $\calR=\{\varominus,\varominus^\calM\}$ be a set of preference classes, $\Ag$ be a set of agents, $\calS$ be the set of all admissible labelings that are compatible with each individual's labeling, and $\calL$ be a labeling from $\calS$. If $\calL$ is Pareto optimal in $\calS$ given that $\Ag$ have \emph{homogeneous} preferences from $\calR$, then $\calL$ is Pareto optimal in $\calS$ given that $\Ag$ have \emph{heterogeneous} preferences from $\calR$.
\comment{
\begin{proof}
Let $\Ag$ be s.t. agents have heterogeneous preferences from $\calR$. Suppose towards a contradiction, that $\calL$ is not Pareto optimal in $\calS$. Then, $\exists \calL_X \in \calS$ s.t:
\begin{equation}
(\forall i \in \Ag: \calL_X \succeq_{i,c(i)} \calL) \wedge (\exists j \in \Ag \text{ s.t. } \calL_X \succ_{j,c(j)} \calL)
\end{equation}
where $c(i) \in \calR, \forall i \in \Ag$ (i.e. $c(i)=\varominus$ or $c(i)=\varominus^\calM$). Then:
\begin{equation}
(\forall i \in \Ag: \calL_X \;c(i)\; \calL_i \subseteq \calL \;c(i)\; \calL_i) \wedge (\exists j \in \Ag \text{ s.t. } \calL \;c(j)\; \calL_j \not\subseteq \calL_X \;c(j)\; \calL_j)
\end{equation}
However, given the compatibility of $\calL$ and $\calL_X$ with every individuals' labeling (since $\calL,\calL_X\in \calS$) and from Lemma \ref{lem.heterosetscomp} (1), if agents who have Hamming set based preferences switched their classes of preferences to IUO Hamming set{s} based preferences or vice versa, then their preferences would not change. As a result, the previous equation would hold even when $c(k)=c(l), \forall k,l \in \Ag$. This means $\calL$ is not Pareto optimal in $\calS$ when $\Ag$ have homogeneous preferences from $\calR$. Contradiction.
\end{proof}
}
\end{theorem}

Also, given our discussion in Lemma \ref{lem.heterosets}, we can add Issue-wise set $\varominus_\calW$ and IUO Issue-wise set{s} $\varominus_\calW^\calM$ to the set $\calR$ in the previous result. For our three operators, we have the following corollaries.

\begin{corollary}\label{cor.heterosco}
Let $\calR=\{\varominus,\varominus_\calW,\varominus^\calM,\varominus_\calW^\calM\}$. The skeptical operator is Pareto optimal in the set of all admissible labelings that are smaller or equal ($\sqsubseteq$) to each of the participants' labelings given that individuals have heterogeneous preferences from $\calR$.  
\end{corollary}

\begin{corollary}
Let $\calR=\{\varominus,\varominus_\calW,\varominus^\calM,\varominus_\calW^\calM\}$. The credulous operator is Pareto optimal in the set of all admissible labelings that are compatible ($\approx$) with each of the participants' labelings given that individuals have heterogeneous preferences from $\calR$.  
\end{corollary}

\begin{corollary}
Let $\calR=\{\varominus,\varominus_\calW,\varominus^\calM,\varominus_\calW^\calM\}$. The super credulous operator is Pareto optimal in the set of all complete labelings that are compatible ($\approx$) with each of the participants' labelings given that individuals have heterogeneous preferences from $\calR$.  
\end{corollary}

{
We showed earlier that the skeptical operator is always Pareto optimal no matter which class of preferences the individuals have, as long as all agents have the same class i.e. homogenous preferences (as Table \ref{tab:Pa} shows). We show here even a stronger result, that is even when agents preferences are hererogenous, and no matter what the combination of classes of preferences that they have, the skeptical operator sustains Pareto optimality. This establishes the robustness of the skeptical operator when it comes to Pareto optimality.
}

\begin{theorem}
Let $\calR=\{\varominus,\varominus_\calW,\varominus^\calM,\varominus_\calW^\calM, \left|\varominus\right|,\left|\varominus_\calW\right|,\left|\varominus^\calM\right|,\left|\varominus_\calW^\calM\right|\}$. The skeptical operator is Pareto optimal in the set of all admissible labelings that are smaller or equal ($\sqsubseteq$) to each of the participants' labelings given that individuals have heterogeneous preferences from $\calR$.  
\comment{
\begin{proof}
Let $\calS$ be the set of all admissible labelings that are smaller or equal ($\sqsubseteq$) to individuals' labelings. Suppose, towards a contradiction that the skeptical operator is not Pareto optimal in $\calS$ given that the set of individuals $\Ag$ have heterogeneous preferences from $\calR$. Then, $\exists \calL_X \in \calS$ s.t. $\calL_X$ Pareto dominates $\calL$ (given heterogeneous preferences). Then:
\begin{equation}
\exists j \in \Ag \text{ s.t. } \calL_X \succ_{j,c(j)} \calL
\end{equation} 

However, from Theorem \ref{thm.bigadm}, $\calL$ is the biggest admissible labeling that is smaller or equal ($\sqsubseteq$) to each individual's labeling. Then, $\forall \calL' \in \calS: \calL' \sqsubseteq \calL \sqsubseteq \calL_i, \forall i \in \Ag$. Hence, for any agent $i$'s labeling $\calL_i$, and for any argument (and consequently, any issue) on which $\calL$ disagrees with $\calL_i$, then $\calL_X$ would disagree in exactly the same way. Contradiction.  
\end{proof}
}
\end{theorem}
}

\section{Strategy Proofness}\label{sec.SP}

Strategic manipulability is usually an undesirable property in which an agent, upon knowing the preferences of other individuals, has incentive to misrepresent her own true opinion in order to force a collective outcome which is closer to her true opinion. A strategic lie is what an agent can say if and when she has the opportunity to vote strategically. 

\begin{definition}[Strategic lie]
Let $P$ be a profile and $\calL_k \in P$ be the most preferred labeling of an agent with preference $\succeq_k$. Let $Op$ be any aggregation operator. A labeling $\calL'_k$ such that $Op(P_{\calL_k/\calL'_k}) \succ_k Op(P)$ is called a strategic lie. Where $P_{\calL_k/\calL'_k}$ is the profile that results from the profile $P$ after agent $k$ changes her vote from $\calL_k$ to $\calL'_k$.
\end{definition}

A strategy proof operator is one where individuals have no incentive to make strategic lies.

\begin{definition}[Strategy proof operator]
An aggregation operator $Op$ is strategy proof if strategic lies are not possible.
\end{definition}

Despite the fact that, as we shall see, for most classes of preference, the aggregation operators turned out to be vulnerable to strategic manipulation, a novel type of lie emerged: the benevolent lie. Unlike the malicious lie, the benevolent lie has positive effects on some of the other agents and no negative effects on any agent.

\begin{definition}[Malicious lie]
Let $Op$ be some aggregation operator and $P$ be a profile of labelings. We say that a strategic lie $\calL'_k$ is malicious iff, for some agent $j \neq k,$ $Op(P)\succ_j Op(P_{\calL_k/\calL'_k})$.
\end{definition}

\begin{definition}[Benevolent lie]
Let $Op$ be some aggregation operator and $P$ be a profile of labelings. We say that a strategic lie $\calL'_k$ is benevolent iff, for any agent $i$ $Op(P_{\calL_k/\calL'_k})\succeq_i Op(P)$ and there exists an agent $j \neq k,$ $Op(P_{\calL_k/\calL'_k})\succ_j Op(P)$.
\end{definition}

{
\subsection{Connections between Classes of Preferences}

{
Similar to the previous section, we start by showing general connections. First, note that the results of Lemma \ref{lem.heterosets} holds in this case. This means that the benevolence property (that is, all strategic lies are benevolent) carries over between Hamming set and Issue-wise set based preferences. The same goes for IUO Hamming set and IUO Issue-wise set based preferences. Fruther, we show that this benevolence property carries over from Hamming distance to IUO Hamming distance based preferences, and from Issue-wise distance to IUO Issue-wise distance based preferences.
}

\begin{theorem}\label{thm.MHaSpSo}
Consider an operator $Op$ that only produces labelings that are compatible ($\approx$) with each individual's labeling. If all strategic lies are benevolent when agents have Hamming distance based preferences then all strategic lies are benevolent when agents have IUO Hamming distance based preferences.
\comment{
\begin{proof}
Let $Op$ be {a \emph{compatible operator}}. Let $P$ be a profile, and $\calL'_k$ a strategic lie of agent $k$. Denote $\calL_{Op}=Op_{\AF}(P)$ and $\calL'_{Op}=Op_{\AF}(P_{\calL_k/\calL'_k})$. {From Lemma \ref{lem.heterosetscomp} (1), since the operator $Op$ only produces labelings that are compatible with all individuals' labelings, then} for every agent $j$ s.t. $j \neq k$: $(\calL_{Op} \succeq_{j,\left |\varominus\right |} \calL'_{Op}$ iff $\calL_{Op} \succeq_{j,\left |\varominus^\calM \right |} \calL'_{Op})$ i.e. Hamming distance based preferences and IUO Hamming distance based preferences are equivalent for all agents other than agent $k$.

Now given Lemma \ref{lem.carryOverHam}, every {strategic lie} with the operator $Op$ given IUO Hamming distance based preferences is also a {strategic lie} given Hamming distance based preferences. However, all those lies are benevolent for every agent $j \neq k$ whether she has Hamming distance based preferences or IUO Hamming distance based preferences. Hence, every lie given IUO Hamming distance based preferences is benevolent. 
\end{proof}
}
\end{theorem}

\begin{theorem}\label{thm.MIwSpSo}
Consider an operator $Op$ that only produces labelings that are compatible ($\approx$) with each individual's labeling. If all strategic lies are benevolent when agents have Issue-wise distance based preferences then all strategic lies are benevolent when agents have IUO Issue-wise distance based preferences.
\comment{
\begin{proof}
This proof is similar to the one for Theorem \ref{thm.MHaSpSo}, with the use of Lemma \ref{lem.heterosetscomp} (2) and Lemma \ref{lem.carryOverIss}.
\end{proof}
}
\end{theorem}

Now we turn to studying the strategy proofness of the three operators: the skeptical, the credulous and the super credulous.

\subsection{Case 1: $\myin$, $\myout$, and $\myundec$ are Equally Distant from Each Other}
\subsubsection{Hamming Set and Hamming Distance}~\\

{
Following, we show that none of the three operators is strategy proof given Hamming set (resp. Hamming distance) based preferences.
}
\begin{observation}\label{obs.HaSpSo}
The skeptical aggregation operator is not strategy proof for neither Hamming set nor Hamming distance based preferences. \comment{Consider the three labelings in Figure \ref{fig:SPSo}. Labeling $\calL_1$ of agent 1 when aggregated with $\calL_2$ gives labeling $\calL_3$, which disagrees with $\calL_1$ on all three arguments. But, when the agent strategically lies and reports labeling $\calL_2$ instead, the result of the aggregation is the same labeling $\calL_2$, which differs only on two arguments $\{A, B\}$. The example is valid for both Hamming set and Hamming distance based preferences.}
\end{observation}
\comment{
\begin{figure}[ht]
    \centering
  \includegraphics[scale=0.4]{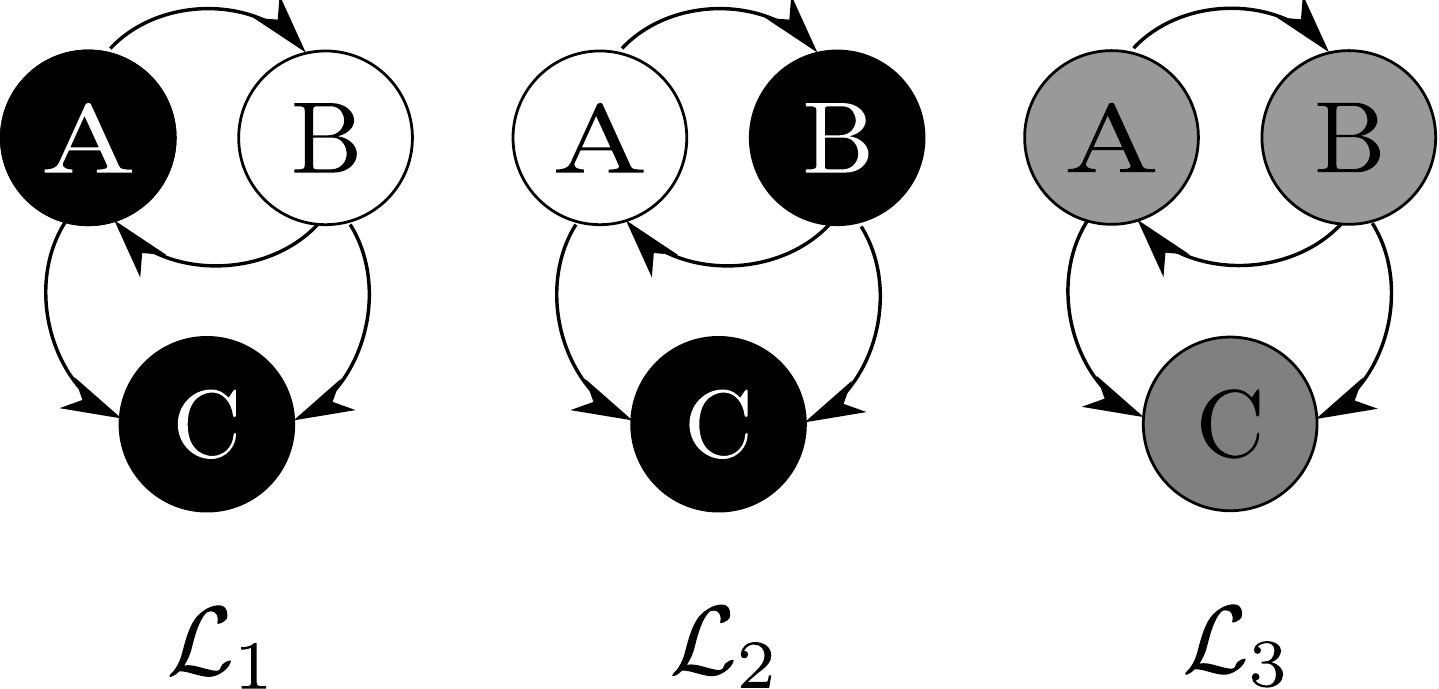}
  \caption{The skeptical operator is not strategy proof.}\label{fig:SPSo}
\end{figure}
}

\begin{observation}\label{obs.HaSpCoSco}
The credulous (resp. super credulous) aggregation operator is not strategy proof for neither Hamming set nor Hamming distance based preferences. \comment{See the example in Figure \ref{fig:SPCoSco}. Labeling $\calL_2$ of agent $2$ when aggregated with $\calL_1$ gives labeling $\calL_{CO}$, which disagrees with $\calL_2$ on the two arguments. But, when the agent strategically lies and reports $\calL'_2$ instead, the result of the aggregation is $\calL'_{CO}$, which matches the labeling $\calL_2$. This lie by agent $2$ makes the agent with labeling $\calL_1$ worse off. The example is valid for both Hamming set and Hamming distance based preferences.}
\end{observation}

\comment{
\begin{figure}[ht]
    \centering
  \includegraphics[scale=0.4]{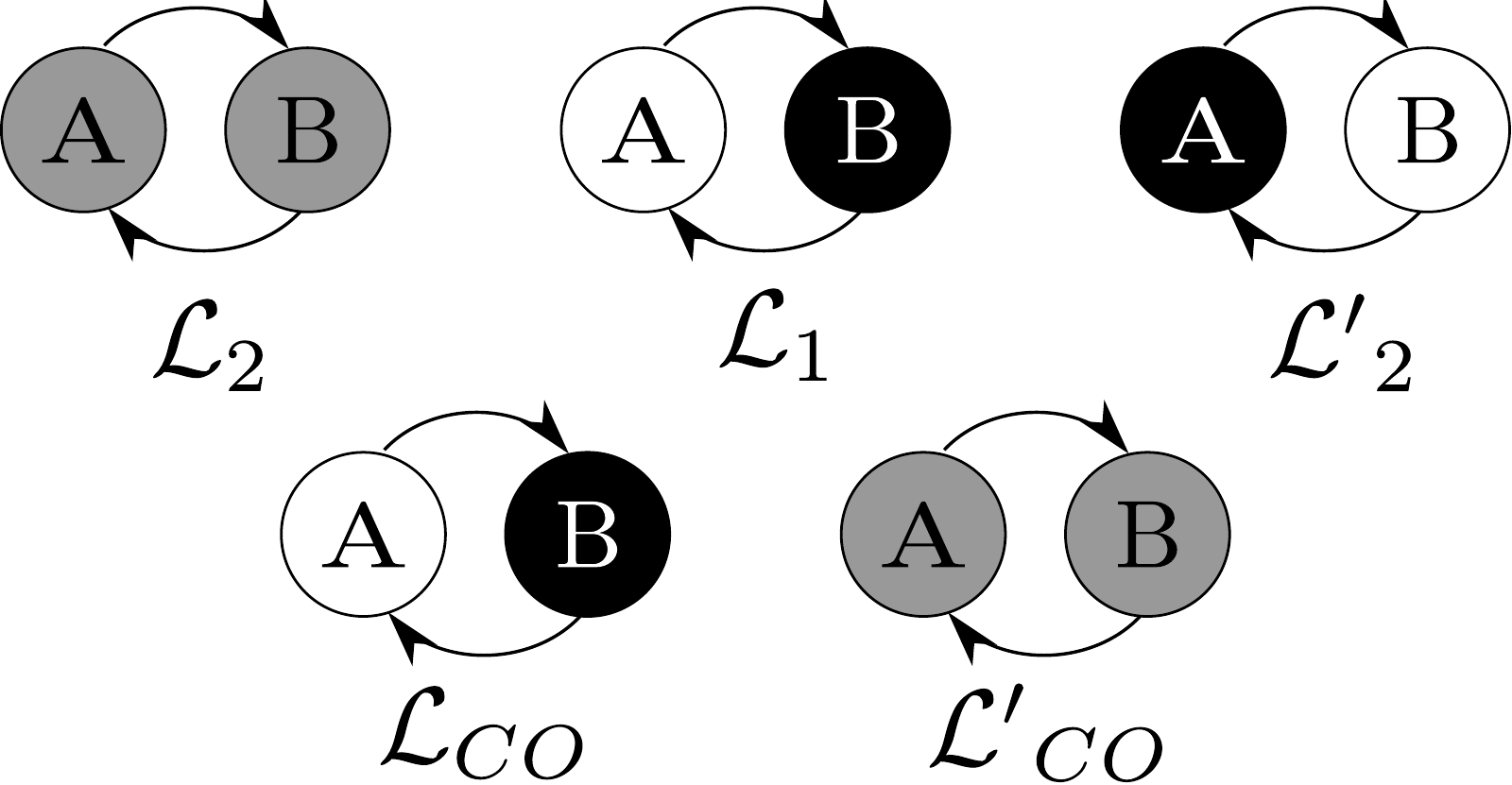}
  \caption{The (super) credulous operator is not strategy proof.}\label{fig:SPCoSco}
\end{figure}
}

For the skeptical aggregation operator, however, every strategic lie is benevolent, given Hamming set (resp. Hamming distance) based preferences. {Unfortunately, this is not the case for the credulous or the super credulous operators.}

\begin{theorem}\label{thm.HaSpSoSet}
Consider the skeptical aggregation operator and Hamming set based preferences. For any agent, her strategic lies are benevolent. 
\comment{
\begin{proof}
Let $P$ be a profile, and $\calL_k'$ a strategic lie of agent $k$. Denote $\calL_{SO} = so_{\AF}(P)$ and $\calL'_{SO} = so_{\AF}(P_{\calL_k/\calL_k'})$. Agent $k$'s preference is $\calL'_{SO} \succ_k \calL_{SO}$ (i). We will show that for any agent $i \neq k$, we have $\calL'_{SO} \succ_i \calL_{SO}$.  
Since the skeptical aggregation operator produces social outcomes that are less or equally committed to all the individual labelings, we have that $\calL'_{SO} \sqsubseteq \calL_i$ for all $i \neq k$ (ii). 
Similarly, we have $\calL_{SO} \sqsubseteq \calL_k$ (iii). From (i) and (iii), by Lemma \ref{lem.prefSetMoreComm}, we have that $\calL_{SO} \sqsubseteq \calL'_{SO}$ (iv). From (iv) and (ii) we have $\calL_{SO} \sqsubseteq \calL'_{SO} \sqsubseteq \calL_i$ for all $i \neq k$. Finally, we can apply Lemma \ref{lem.moreCommPref} to obtain $\calL'_{SO} \succeq_i \calL_{SO}$ for all $i \neq k$ (v). We showed that a lie cannot be malicious, now we show that it is benevolent.

(iii) implies $\myundec(\calL_{k}) \subseteq \myundec(\calL_{SO})$ (vi). (i) and (vi) imply $\exists A \in \mydec(\calL_{k}) : A \in \myundec(\calL_{SO}) \wedge A \in \mydec(\calL_{SO}')$ (vii). From (vii), (ii) and (v) $\calL_{SO}' \succ_i \calL_{SO}$ for $i \neq k$.
\end{proof}
}
\end{theorem}

\begin{theorem}\label{thm.HaSpSoDis}
Consider the skeptical aggregation operator and Hamming distance based preferences. For any agent, her strategic lies are benevolent.
\end{theorem}

{
Note that the previous two theorems raise an interesting point. Given the Pareto optimality of the skeptical operator for Hamming set/distance based preferences, one would expect that benevolent lies are not possible. Otherwise, it means there exists another labeling that is more preferred by every agent and strictly preferred by at lease one agent. This contradicts the Pareto optimality result found earlier.

However, it is important to remember that the Pareto optimality results found earlier are all with respect to the sets of labelings that are smaller or equal (or compatible in the case of the other operators) to each individuals' labelings. On the other hand, an outcome given a benevolent lie is not compatible with every individual's labeling i.e. while the skeptical operator does produce labelings that are compatible with each individual's true labeling, it does so for the submitted labelings only. Hence, when an agent $k$ lies and submits $\calL'_k$ instead of $\calL_k$, the outcome $\calL'_{SO}$ (which is the outcome when $k$ submits $\calL'_k$) is compatible with $\calL'_k$ but not necessarily to $\calL_k$. As a result, the labeling $\calL'_{SO}$ does not belong to the set of labelings that $\calL_{SO}$ is compared to when studying Pareto optimality.

This highlights another interesting point that can be implied by the benevolence and Pareto optimality of the skeptical operator. When using the skeptical operator, whenever an agent $k$ considers lying in order to get a closer labeling to $\calL_k$, she is faced with an inevitable trade-off between getting a less or equally committed outcome and getting a closer (i.e. more preferred) outcome.
}

\subsubsection{Issue-wise Set and Issue-wise Distance}~\\
{
Similar to the Hamming based preferences, none of the three operators is strategy proof given Issue-wise set (resp. Issue-wise distance) based preferences.
}
\begin{observation}\label{obs.SpSo}
The skeptical aggregation operator is not strategy proof for neither Issue-wise set nor Issue-wise distance based preferences. \comment{Consider the three labelings in Figure \ref{fig:SPSoI}.\footnote{This figure is the same as Fig \ref{fig:SPSo} with issues being evidenced.} Labeling $\calL_1$ of agent $1$ when aggregated with $\calL_2$ gives labeling $\calL_3$, which disagrees with $\calL_1$ on both of the two issues. But, when the agent strategically lies and reports labeling $\calL_2$ instead, the result of the aggregation is the same labeling $\calL_2$, which differs only on one issue $\{\{A,B\}\}$. The example is valid for both Issue-wise set and Issue-wise distance based preferences.}
\end{observation}
\comment{
\begin{figure}[ht]
    \centering
  \includegraphics[scale=0.4]{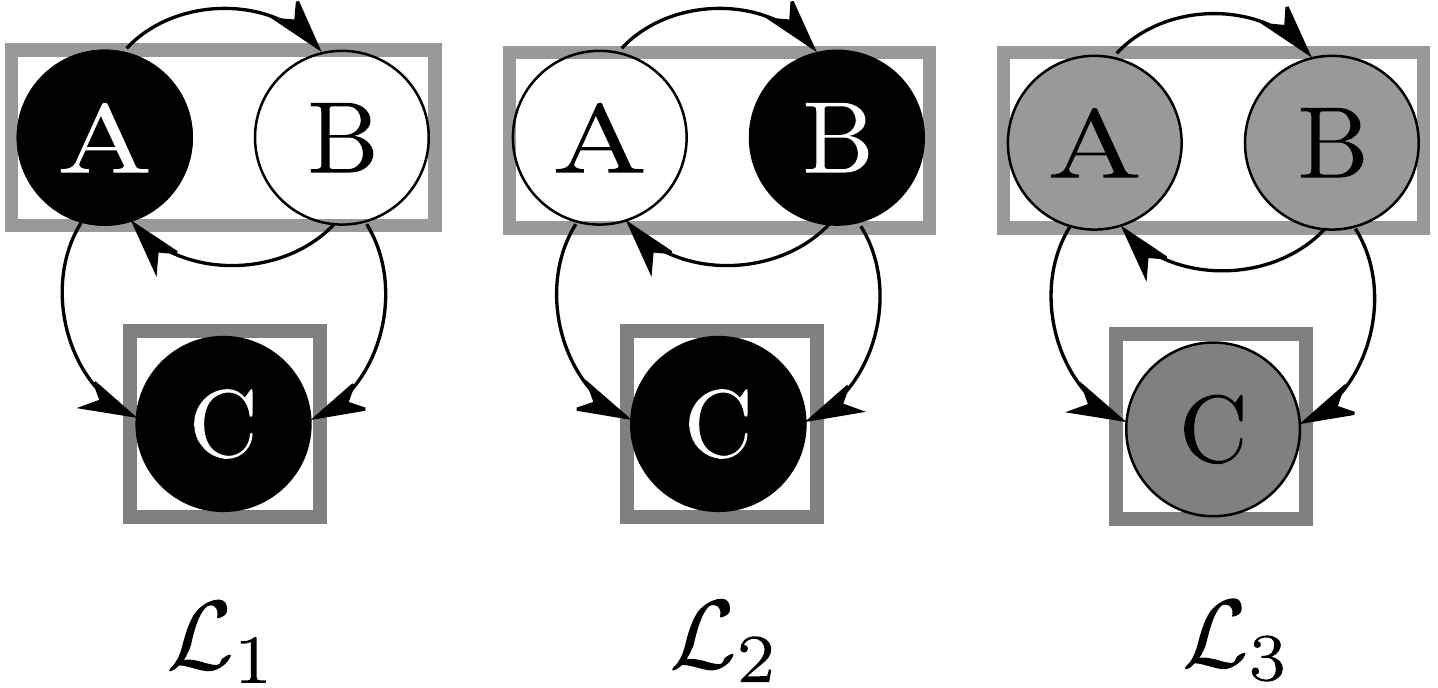}
  \caption{The skeptical operator is not strategy proof. The set of arguments located in one box form an issue.}\label{fig:SPSoI}
\end{figure}
}

\begin{observation}\label{obs.SpCoSco}
The credulous (resp. super credulous) aggregation operator is not strategy proof for neither Issue-wise set nor Issue-wise distance based preferences. \comment{In Figure \ref{fig:SPCoScoI},\footnote{This figure is the same as Fig \ref{fig:SPCoSco} with issues being evidenced.} labeling $\calL_2$ of agent $2$ when aggregated with $\calL_1$ gives labeling $\calL_{CO}$, which disagrees with $\calL_2$ on the one and only issue. But, when the agent strategically lies and reports $\calL'_2$ instead, the result of the aggregation is $\calL'_{CO}$, which matches the labeling $\calL_2$. This lie by agent $2$ makes the agent with labeling $\calL_1$ worse off. The example is valid for both Issue-wise set and Issue-wise distance based preferences.}
\end{observation}
\comment{
\begin{figure}[ht]
    \centering
  \includegraphics[scale=0.4]{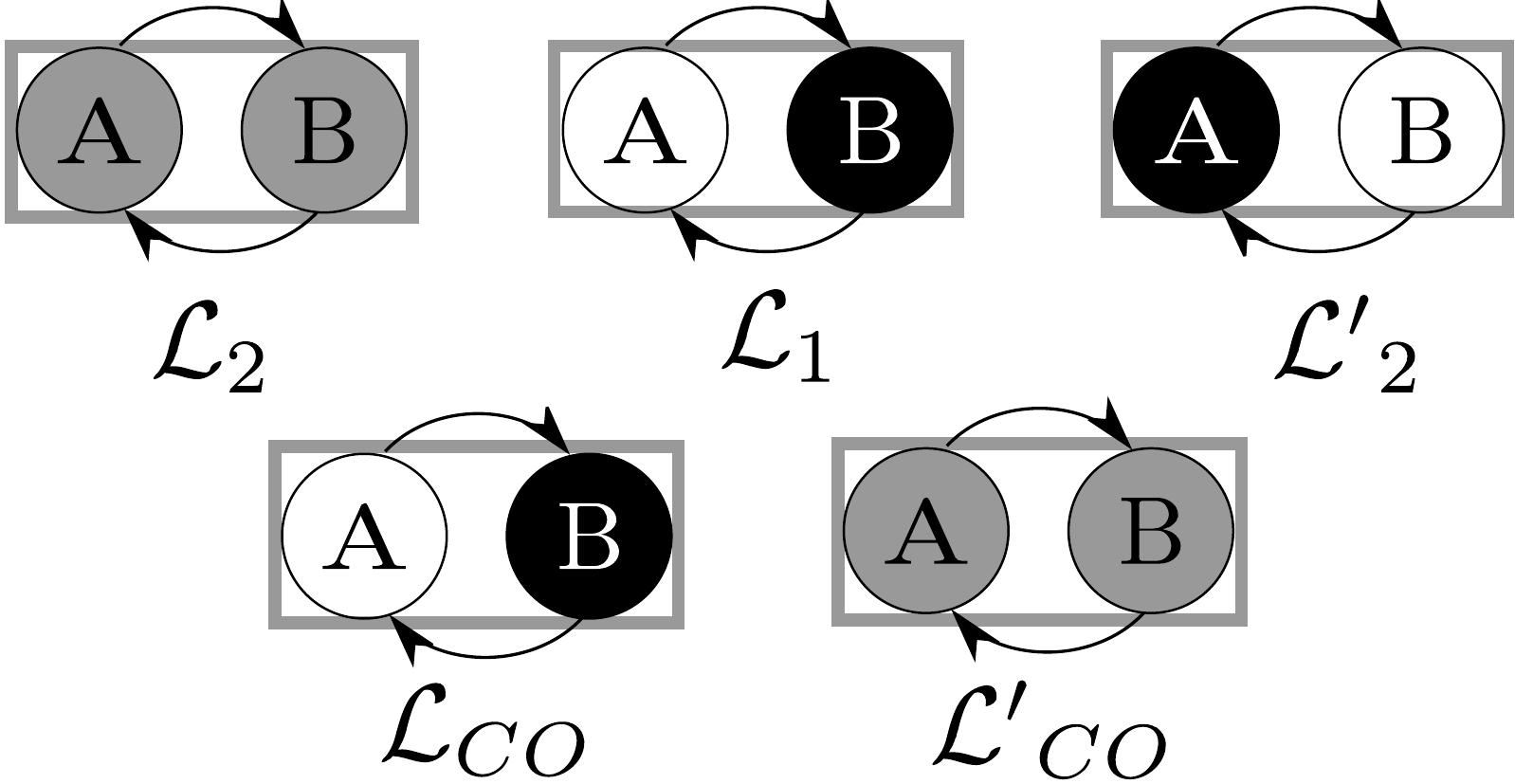}
  \caption{The (super) credulous operator is not strategy proof. The set of arguments located in one box form an issue.}\label{fig:SPCoScoI}
\end{figure}
}

{
Again, similar to the Hamming based preference, only the skeptical aggregation operator has the benevolent property (every strategic lie is benevolent), given Issue-wise set (resp. Issue-wise distance) based preferences. As for the Issue-wise set based preferences, we can use the result of Lemma \ref{lem.heterosets} (that is, Issue-wise set based preferences coincide with Hamming set based preferences) to substitute ``Hamming set'' with ``Issue-wise set'' in Theorem \ref{thm.HaSpSoSet}. As for the Issue-wise distance based preferences, it is shown in the following theorem.}

\begin{theorem}\label{thm.SpSoDis}
Consider the skeptical aggregation operator and Issue-wise distance based preferences. For any agent, her strategic lies are benevolent.
\comment{
\begin{proof}
Let $P$ be a profile of labelings, and $\calL'_k$ be a strategic lie of agent $k$ whose most preferred labeling is $\calL_k$. Denote $\calL_{SO}=so_{\AF}(P)$ and $\calL'_{SO}=so_{\AF}(P_{\calL_k/\calL'_k})$. We show that if $\calL'_{SO}$ is strictly preferred by an agent $k$ then it is also strictly preferred by any other agent. Without loss of generality, we can take agent $j$, $j \neq k$, whose most preferred labeling is $\calL_j$.

Let $\calI_{de}(\calL)$ (resp. $\calI_{un}(\calL)$) be the set of issues, each of which has arguments that are only decided (resp. undecided) according to $\calL$. We call $\calI_{de}(\calL)$ (resp. $\calI_{un}(\calL)$) a decided (resp. undecided) issue (w.r.t $\calL$). Let us partition the issues into the following disjoint groups: 

\begin{itemize}
\item $\calX = \calI_{de}(\calL_{SO})\setminus\calI_{de}(\calL'_{SO})$ (decided issues that became undecided).
\item $\calY = \calI_{de}(\calL'_{SO})\setminus\calI_{de}(\calL_{SO})$ (undecided issues that became decided).
\item $\calZ = \calI_{de}(\calL_{SO}) \cap \calI_{de}(\calL'_{SO})$ (issues decided in both labelings).
\item $\calV = \calI_{un}(\calL_{SO}) \cap \calI_{un}(\calL'_{SO})$ (issues undecided in both labelings).
\end{itemize}

The rest is similar to Theorem \ref{thm.HaSpSoDis}, but using issues instead of arguments.
\end{proof}
}
\end{theorem}

\subsection{Case 2: $\myundec$ is in the Middle between $\myin$ and $\myout$}
In this part, we analyze the strategy proofness for the three operators given the classes of preferences that assume $\myundec$ is in the middle between $\myin$ and $\myout$ ($dist({\mydec},\myundec)$ $< dist (\myin,\myout)$). 

\subsubsection{IUO Hamming Set{s} and IUO Hamming Distance}
{Following, we show the strongest result for this section. The skeptical operator is strategy proof given the IUO Hamming set{s} based preferences. This result also holds for IUO Issue-wise set{s} given the discussion for Lemma \ref{lem.heterosets}.
}
\begin{theorem}\label{thm.MHaSpSoSet}
The skeptical aggregation operator is strategy proof when individuals have IUO Hamming set{s} based preferences.
\end{theorem}

{
The previous result does not hold for the credulous or the super credulous operators. Further, none of the three operators is strategy-proof when individuals have IUO Hamming distance based preferences. However, as was the case with other classes of preferences, lies with the skeptical operators are always benevolent, unlike those with the credulous or the super credulou operators.
}

\begin{observation}\label{obs.MHaSpSo}
The skeptical aggregation operator is not strategy proof when individuals have IUO Hamming distance based preferences. \comment{Consider the three labelings in Figure \ref{fig:SPSoIUOHamDis}. Labeling $\calL_1$ of agent 1 when aggregated (using skeptical operator) with $\calL_2$ gives labeling $\calL_3$, which differs from $\calL_1$ on all five arguments with respect to $\mydec-\myundec$ Hamming set. Then, $\calL_1 \left |\varominus^\calM \right | \calL_3 = 2 \times 0 + 1 \times 5= 5$. But, when the agent strategically lies and reports labeling $\calL_2$ instead, the result of the aggregation is the same labeling $\calL_2$, which differs only on two arguments $\{A, B\}$ with respect to $\myin-\myout$ Hamming set. Then, $\calL_1 \left |\varominus^\calM \right | \calL_2 = 2 \times 2 + 1 \times 0= 4$.} 
\end{observation}
\comment{
\begin{figure}[ht]
    \centering
  \includegraphics[scale=0.6]{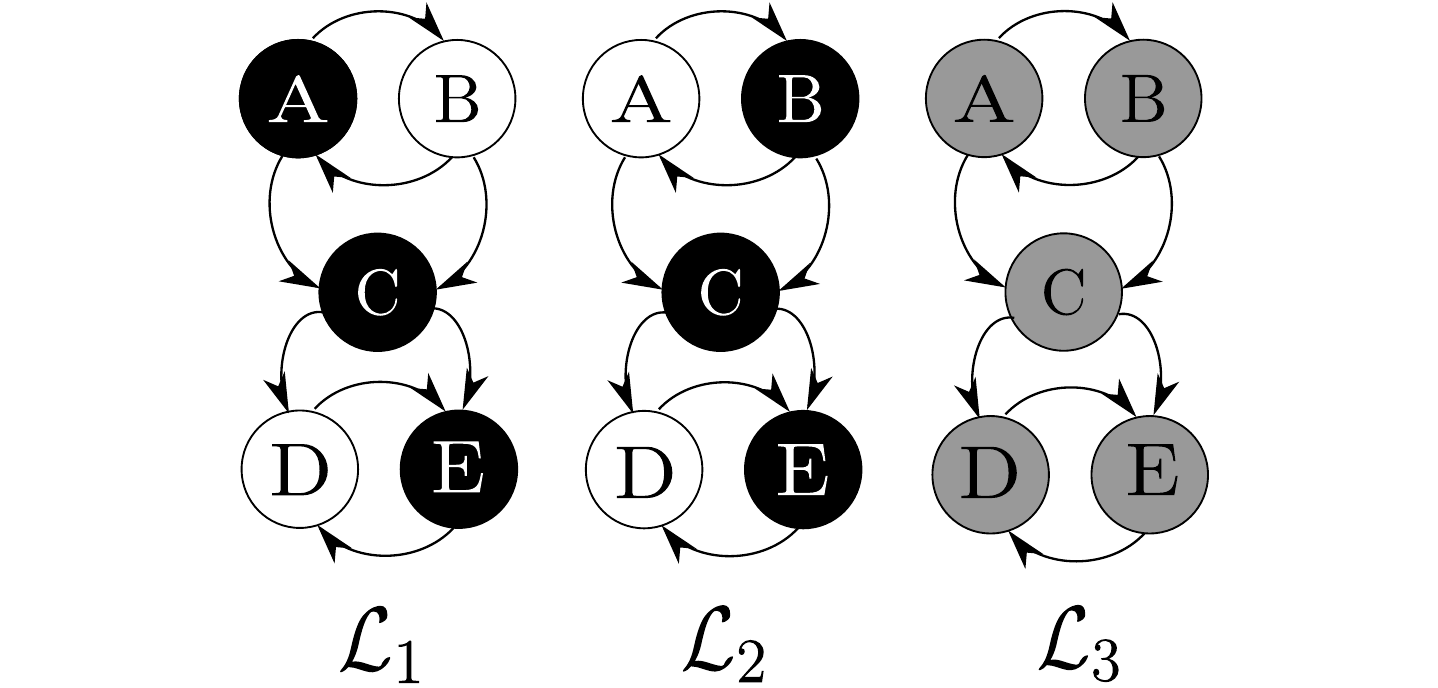}
  \caption{The skeptical operator is not strategy proof when agents have IUO Hamming distance preferences.}\label{fig:SPSoIUOHamDis}
\end{figure}
}


\begin{observation}\label{obs.MHaSpCoSco}
The credulous (resp. super credulous) aggregation operator is not strategy proof for neither IUO Hamming set{s} nor IUO Hamming distance based preferences. \comment{The example in Figure \ref{fig:SPCoSco} can serve as a counterexample for the case where individuals have IUO Hamming set{s} (or IUO Hamming distance) based preferences. The agent with labeling $\calL_2$ can insincerely report $\calL'_2$ to obtain her preferred labeling. This makes an agent with labeling $\calL_1$ worse off.} 
\end{observation}


\begin{proposition}\label{prop.MHaSpSo}
Consider the skeptical aggregation operator and IUO Hamming distance based preferences. For any agent, her strategic lies are benevolent.
\comment{
\begin{proof}
From Theorem \ref{thm.HaSpSoDis} and Theorem \ref{thm.MHaSpSo}, the strategic lies are benevolent when individuals have IUO Hamming distance based.
\end{proof}
}
\end{proposition}

\subsubsection{IUO Issue-wise Set{s} and IUO Issue-wise Distance}~\\
{

{
Similar to the results in the previous part, the skeptical operator is strategy-proof given IUO Issue-wise sets based preferences (by substituting ``IUO Hamming set{s}'' with ``IUO Issue-wise set{s}'' in Theorem \ref{thm.MHaSpSoSet}, given the discussion for Lemma \ref{lem.heterosets}), unlike the credulous or the super credulous operators for which lies are possible and might not be benevolent. Further, none of the three operators is strategy-proof when individuals have IUO Hamming distance based preferences, but lies with the skeptical operators are always benevolent, unlike those with the credulous or the super credulou operators.
}

}


\begin{observation}\label{obs.MSpSo}
The skeptical aggregation operator is not strategy proof when individuals have IUO Issue-wise distance based preferences. \comment{Consider the three labelings in Figure \ref{fig:SPSoIUOIssueDis}.\footnote{This figure is the same as Fig \ref{fig:SPSoIUOHamDis} with issues being evidenced.} Labeling $\calL_1$ of agent 1 when aggregated (using skeptical operator) with $\calL_2$ gives labeling $\calL_3$, which differs from $\calL_1$ on all three issues with respect to $\mydec-\myundec$ Issue-wise set. Then, $\calL_1 \left |\varominus_\calW^\calM \right | \calL_3 = 2 \times 0 + 1 \times 3= 3$. But, when the agent strategically lies and reports labeling $\calL_2$ instead, the result of the aggregation is the same labeling $\calL_2$, which differs only on one is $\{\{A, B\}\}$ with respect to $\myin-\myout$ Issue-wise set. Then, $\calL_1 \left |\varominus_\calW^\calM \right | \calL_2 = 2 \times 1 + 1 \times 0= 2$.} 
\end{observation}
\comment{
\begin{figure}[ht]
    \centering
  \includegraphics[scale=0.6]{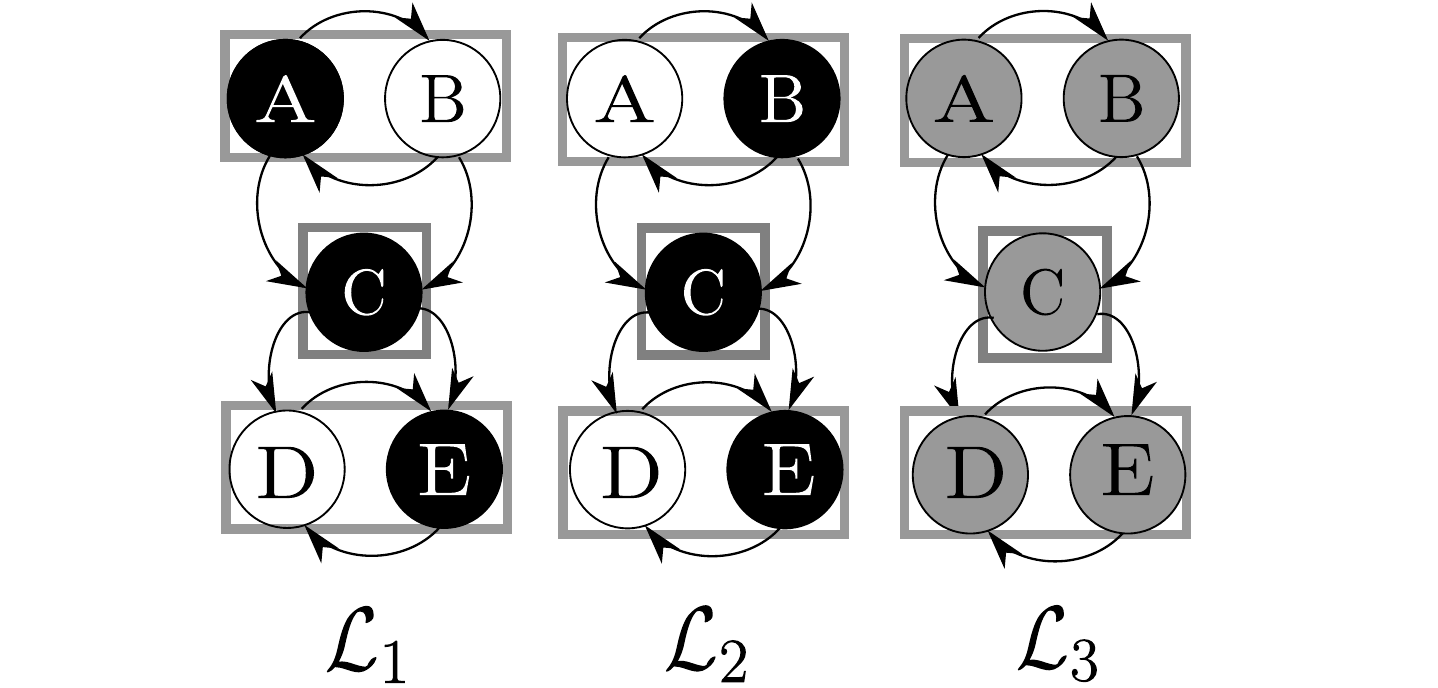}
  \caption{The skeptical operator is not strategy proof when agents have IUO Issue-wise distance preferences. The set of arguments located in one box form an issue.}\label{fig:SPSoIUOIssueDis}
\end{figure}
}


\begin{observation}\label{obs.MSpCoSco}
The credulous and super credulous aggregation operators are not strategy proof when individuals have IUO Issue-wise set{s} (resp. distance) based preferences. \comment{The example in Figure \ref{fig:SPCoScoI} can serve as a counter example for the case where individuals have IUO Issue-wise set{s} (resp. distance) based preferences. The agent with labeling $\calL_2$ can insincerely report $\calL'_2$ to obtain her preferred labeling. This makes an agent with labeling $\calL_1$ worse off. }
\end{observation}


\begin{proposition}\label{prop.MIwSpSo}
Consider the skeptical aggregation operator and IUO Issue-wise distance based preferences. For any agent, her strategic lies are benevolent.
\comment{
\begin{proof}
From Theorem \ref{thm.SpSoDis} and Theorem \ref{thm.MIwSpSo}, the strategic lies are benevolent when individuals have IUO Issue-wise distance based.
\end{proof}
}
\end{proposition}

Table \ref{tab:SP} summarizes the strategy proofness results for the three operators given all the eight classes of preferences.

\begin{table}[htbp]
  \centering
  
   \begin{tabular}{|c|c|c|c|}
    \hline
          & \textbf{Skeptical} & \textbf{Credulous} & \textbf{Super Credulous} \\
          & \textbf{Operator} & \textbf{Operator} & \textbf{Operator} \\
    \hline
    \textbf{Hamming set}   & No (Obs. \ref{obs.HaSpSo} and \ref{obs.SpSo})  & No, and                            & No, and\\
    \textbf{Issue-wise set} & but benev. (Thm. \ref{thm.HaSpSoSet})         & not benev. (Obs. \ref{obs.HaSpCoSco} and \ref{obs.SpCoSco}) & not benev. (Obs. \ref{obs.HaSpCoSco} and \ref{obs.SpCoSco})\\
    \hline
    \textbf{Hamming dist.} & No (Obs. \ref{obs.HaSpSo})               & No, and                            & No, and\\
    \textbf{}                 & but benev. (Thm. \ref{thm.HaSpSoDis}) & not benev. (Obs. \ref{obs.HaSpCoSco})& not benev. (Obs. \ref{obs.HaSpCoSco})\\
    \hline
    \textbf{Issue-wise dist.} & No (Obs. \ref{obs.SpSo})            & No, and                            & No, and\\
    \textbf{}                 & but benev. (Thm. \ref{thm.SpSoDis}) & not benev. (Obs. \ref{obs.SpCoSco})& not benev. (Obs. \ref{obs.SpCoSco})\\
    \hline

    \textbf{IUO Hamming set{s}}    & Yes (Thm. \ref{thm.MHaSpSoSet}) & No, and                             & No, and\\
    \textbf{IUO Issue-wise set{s}} &                                 & not benev. (Obs. \ref{obs.MHaSpCoSco} and \ref{obs.MSpCoSco}) & not benev. (Obs. \ref{obs.MHaSpCoSco} and \ref{obs.MSpCoSco})\\
    \hline
    \textbf{IUO Hamming dist.}& No (Obs. \ref{obs.MHaSpSo})            & No, and                             & No, and\\
    \textbf{}                & but benev. (Prop. \ref{prop.MHaSpSo}) & not benev. (Obs. \ref{obs.MHaSpCoSco})& not benev. (Obs. \ref{obs.MHaSpCoSco})\\
    \hline
    \textbf{IUO Issue-wise dist.} & No (Obs. \ref{obs.MSpSo})            & No, and                            & No, and\\
    \textbf{}                     & but benev. (Prop. \ref{prop.MIwSpSo}) & not benev. (Obs. \ref{obs.MSpCoSco})& not benev. (Obs. \ref{obs.MSpCoSco})\\
    \hline
    \end{tabular}%
  \caption{Strategy proofness of operators depending on the type of preferences.}\label{tab:SP}
\end{table}%

{
\subsection{Heterogeneous Preferences}
Following Subsection \ref{subs.hetero}, we do a similar analysis for the case where agents have heterogeneous preferences. Since strategy proofness is usually considered given other agents's preferences are fixed, it is easy to show the result for the heterogeneous preferences given the homogeneous preferences.

\begin{theorem}
Let $\calF$ be the set of all possible classes of preferences, $\calR$ be some set s.t. $\calR \subseteq \calF$, and $\Ag$ be the set of agents. If an operator is strategy proof given that $\Ag$ have homogeneous preferences from $\calR$, then it is strategy proof given that $\Ag$ have heterogeneous preferences from $\calR$.
\comment{
\begin{proof}
Let $Op$ be an operator that is strategy proof given that $\Ag$ have homogeneous preferences from $\calR$ (i.e. $\forall i,j \in \Ag$ $c(i)=c(j) \in \calR$). Then, there exists no single agent $j$ that has an incentive to lie about her preferences (given that all agents have the same class of preferences). Note that agent $j$ has no incentive to lie given that the submitted labelings by all agents other than $j$ are fixed. Hence, the classes of preferences that are assumed for any agent $k \neq j$ do not affect the incentive of agent $j$ to lie or otherwise. Thus, the same result would hold whether other agents' preferences are different from $c(j)$ or not.
\end{proof}
}
\end{theorem}

\begin{corollary}
Let $\calR = \{\varominus^\calM,\varominus^\calM_\calW\}$. The skeptical aggregation operator is strategy proof given that agents have heterogeneous preferences from $\calR$. 
\end{corollary}
}

\section{Discussion and Future Work}\label{sec.conc}
{
}

In order to apply argumentation to multi-agent conflict resolution, it is crucial to take into account not only postulates about logical consistency, but also measures of social optimality and strategic manipulation. Two key criteria are Pareto optimality and strategy proofness, which are fundamental in any social choice and multi-agent setting. In this study, we have analyzed and compared three aggregation operators, namely the skeptical, the credulous and the super credulous operators with respect to a wide range of classes of preferences. Our comparison is based on two fundamental criteria, namely Pareto optimality and strategy proofness. Eight different classes of preferences were considered by using eight different distance methods. {Additionally, we established relations between the different classes of preferences. Some of these relations hold for any aggregation operator and others for some special aggregation operators. Moreover, we also consider cases where agents do not share the same classes of preference.}

We showed that the skeptical operator guarantees Pareto optimal outcomes given all the different classes of preferences, while the credulous and super credulous operators only guarantee Pareto optimal outcomes given the set-based preferences. Since more committed outcomes might be more desirable in general, credulous and super credulous operators will be preferable if individuals' preferences are known to be set-based. However, if the individuals preferences are unknown or are known to be distance-based, then there is a trade-off between Pareto optimality and the more committed outcomes. As for the strategy proofness, the three operators are vulnerable to manipulation given most classes of preferences. However, the skeptical operator guarantees benevolent lies. Understandably, unlike malicious lies, benevolent ones are not harmful to the group. Hence, there is another trade-off in choosing an appropriate operator between avoiding the malicious lies and choosing the more committed outcomes.

{ 
}

Few studies have considered Pareto optimality and strategy proofness with argument based aggregation. Rahwan and Larson \cite{rahwan:larson:2008a} defined a set of simplistic agents preferences over argumentation outcomes, and studied the Pareto optimality of different argument evaluation rules defined using classical semantics (e.g. \emph{complete},...etc.) given agents with these simple types of preferences. Unlike Rahwan and Larson, we study the Pareto optimality of labeling aggregation operators that produce a collective evaluation given many different evaluations. Another difference is that we consider more realistic, distance-based preferences. As for strategy proofness, since the Gibbard-Satterthwaite theorem \cite{gibbard:1973,satterthwaite:1975}, much research has been done towards analyzing strategic manipulation of preference aggregation (PA) rules \cite{gibbard1977manipulation,gibbard1978straight,moulin1980strategy,chamberlin1982mathematical,nitzan1985vulnerability,favardin2002borda}. Strategy proofness of judgment aggregation (JA) operators have been first studied by Dietrich and List \cite{dietrich2006judgment,dietrich2007strategy}. In the former, Dietrich mentioned some \emph{independence} conditions that make the rule strategy proof. In the latter, Dietrich and List showed equivalence between satisfying strategy proofness and satisfying both the \emph{independence} and \emph{monotonicity} postulates.



The first study of strategy proofness of labeling aggregation operator has been done by Rahwan and Tohm\'e \cite{rahwan:tohme:2010} in the context of a specific labeling aggregation operator (argument-wise plurality rule). They showed the strategy proofness of this operator given agents with a particular class of preferences, dubbed focal set preferences. Our work considers different labeling aggregation operators, and we provide the first broad analysis for strategy proofness of labeling aggregation operators given a wide variety of preferences. Strategic manipulation in argumentation has also been studied by Rahwan, Larson and others \cite{rahwan:etal:2009,rahwan:larson:2008,pan:etal:2010}, when arguments are distributed among agents, and where these agents may choose to show or hide arguments. Thus, that work focuses on how agents contribute to the construction of the argument graph itself, which is then evaluated centrally by the mechanism (e.g. a judge). In contrast, our present paper is concerned with strategic manipulation of labeling aggregation operators which take as input different evaluations of a given fixed graph. In fact, the difference between our problem and theirs can be analogized to jury versus litigators. The former are provided with shared information, while the latter propose their information on a dialog basis.  

{ 
}


{
}


{
We believe that one of the strengths of the argumentation approach is that it is able to provide a dialectical specification of non-monotonic inference. That is, whether or not an argument is accepted (w.r.t. a specific argumentation semantics) can be assessed using dialectical proof procedures. For instance whether an argument is labelled $\myin$ by the grounded labelling can be assessed using either the \emph{Standard Grounded Game} \cite{modgil2009proof} or the more recently defined \emph{Grounded Persuasion Game} \cite{caminada2012grounded}. Another example is the \emph{Admissibility Game} \cite{caminada2014preferred} that assesses whether an argument $A$ is in an admissible set (equiv. in a preferred extension or a complete extension). The seminal work by Dung with model-based semantics such as the grounded and preferred semantics laid the building blocks for these games. Likewise, we believe our work and the work of Caminada and Pigozzi \cite{caminada:pigozzi:2010} lays the building blocks for dialectical preference aggregation by studying the aggregation of opinions using model-based semantics. For instance, by using modified versions of the above mentioned games, one can define a game for the down-admissible operator and a game for the up-complete operator. Then, using these two games, one can provide dialectical proof procedures for each of the JA-operators (skeptical, credulous and super-credulous) studied in this work. The interested reader can refer to the technical report \cite{tech:Martin2014daucgames} for more details about how the previously mentioned games can be modified to define games for the studied aggregation operators.

}


\begin{acknowledgements}
Part of this paper was written during a visit of Edmond Awad to the University of Luxembourg which was generously supported by SINTELNET. The research of Martin Caminada was supported by the Engineering and Physical Sciences Research Council (EPSRC, UK), grant ref. EP/J012084/1 (SAsSY project). The research of Gabriella Pigozzi benefited from the support of the project AMANDE ANR-13-BS02-0004 of the French National Research Agency (ANR). The research of Miko\l{}aj Podlaszewski was supported by the National Research Fund, Luxembourg (LAAMIcomp project).
\end{acknowledgements}

\bibliographystyle{plain}
\bibliography{references}

\begin{thebibliography}{10}

\bibitem{arrow:1951}
Kenneth~J. Arrow.
\newblock {\em Social choice and individual values}.
\newblock Wiley, New York NY, USA, 1951.

\bibitem{arrow:etal:2002}
Kenneth~J. Arrow, A.~K. Sen, and K.~Suzumura, editors.
\newblock {\em Handbook of Social Choice and Welfare}, volume~1.
\newblock Elsevier Science Publishers (North-Holland), 2002.

\bibitem{awad2015judgment}
Edmond Awad, Richard Booth, Fernando Tohme, and Iyad Rahwan.
\newblock Judgment aggregation in multi-agent argumentation.
\newblock {\em Journal of Logic and Computation}, 2015.

\bibitem{benchcapon:dunne:2007}
Trevor J.~M. Bench-Capon and Paul~E. Dunne.
\newblock Argumentation in artificial intelligence.
\newblock {\em Artificial Intelligence}, 171(10--15):619--641, 2007.

\bibitem{besnard2004checking}
Philippe Besnard and Sylvie Doutre.
\newblock Checking the acceptability of a set of arguments.
\newblock In {\em NMR}, volume~4, pages 59--64, 2004.

\bibitem{besnard2014encoding}
Philippe Besnard, Sylvie Doutre, and Andreas Herzig.
\newblock Encoding argument graphs in logic.
\newblock In {\em Information Processing and Management of Uncertainty in
  Knowledge-Based Systems}, pages 345--354. Springer, 2014.

\bibitem{besnard:hunter:2008}
Philippe Besnard and Anthony Hunter.
\newblock {\em Elements of Argumentation}.
\newblock MIT Press, Cambridge MA, USA, 2008.

\bibitem{bonnefon2010behavioral}
Jean-Fran{\c{c}}ois Bonnefon.
\newblock Behavioral evidence for framing effects in the resolution of the
  doctrinal paradox.
\newblock {\em Social choice and welfare}, 34(4):631--641, 2010.

\bibitem{booth2014interval}
Richard Booth, Edmond Awad, and Iyad Rahwan.
\newblock Interval methods for judgment aggregation in argumentation.
\newblock In {\em Proceedings of the 14th International Conference on
  Principles of Knowledge Representation and Reasoning (KR 2014)}, pages
  594--597, 2014.

\bibitem{booth2012quantifying}
Richard Booth, Martin Caminada, Miko{\l}aj Podlaszewski, and Iyad Rahwan.
\newblock Quantifying disagreement in argument-based reasoning.
\newblock In {\em Proceedings of the 11th International Conference on
  Autonomous Agents and Multiagent Systems-Volume 1}, pages 493--500.
  International Foundation for Autonomous Agents and Multiagent Systems, 2012.

\bibitem{booth2014distances}
Richard Booth and Miko\l{}aj Podlaszewski.
\newblock Using distances for aggregation in abstract argumentation.
\newblock In {\em Proceedings of the 26th Benelux Conference on Artificial
  Intelligence (BNAIC2014)}, pages 17--24, 2014.

\bibitem{tech:Martin2014daucgames}
Martin Caminada and Richard Booth.
\newblock Towards dialogue games for the down-admissible and up-complete
  procedures.
\newblock Technical Report http://hdl.handle.net/10993/16504, University of
  Luxembourg, April 2014.

\bibitem{caminada2014preferred}
Martin Caminada, Wolfgang Dvor{\'a}k, and Srdjan Vesic.
\newblock Preferred semantics as socratic discussion.
\newblock {\em Journal of Logic and Computation}, 2014.

\bibitem{caminada:gabbay:2009}
Martin Caminada and Dov~M. Gabbay.
\newblock A logical account of formal argumentation.
\newblock {\em Studia Logica}, 93(2-3):109--145, 2009.

\bibitem{caminada:pigozzi:2010}
Martin Caminada and Gabriella Pigozzi.
\newblock On judgment aggregation in abstract argumentation.
\newblock {\em Autonomous Agents and Multi-Agent Systems}, 22(1):64--102, 2011.

\bibitem{caminada2011manipulation}
Martin Caminada, Gabriella Pigozzi, and Miko{\l}aj Podlaszewski.
\newblock Manipulation in group argument evaluation.
\newblock In {\em Proceedings of the Twenty-Second international joint
  conference on Artificial Intelligence-Volume One}, pages 121--126. AAAI
  Press, 2011.

\bibitem{caminada2012grounded}
Martin Caminada and Mikolaj Podlaszewski.
\newblock Grounded semantics as persuasion dialogue.
\newblock {\em COMMA}, 245:478--485, 2012.

\bibitem{caminada:2006}
Martin W.~A. Caminada.
\newblock On the issue of reinstatement in argumentation.
\newblock In Michael Fisher, Wiebe van~der Hoek, Boris Konev, and Alexei
  Lisitsa, editors, {\em {Proceedings of the 10th European Conference on Logics
  in Artificial Intelligence (JELIA)}}, volume 4160 of {\em Lecture Notes in
  Computer Science}, pages 111--123. Springer, 2006.

\bibitem{chamberlin1982mathematical}
John~R Chamberlin.
\newblock A mathematical programming approach to assessing the manipulability
  of social choice functions.
\newblock {\em Political Methodology}, 8(4):25--38, 1982.

\bibitem{dietrich2006judgment}
Franz Dietrich.
\newblock Judgment aggregation:(im) possibility theorems.
\newblock {\em Journal of Economic Theory}, 126(1):286--298, 2006.

\bibitem{dietrich2007strategy}
Franz Dietrich and Christian List.
\newblock Strategy-proof judgment aggregation.
\newblock {\em Economics and Philosophy}, 23:269--300, 2007.

\bibitem{dokow2010aggregation}
Elad Dokow and Ron Holzman.
\newblock Aggregation of binary evaluations with abstentions.
\newblock {\em Journal of Economic Theory}, 145(2):544--561, 2010.

\bibitem{dokow2010}
Elad Dokow and Ron Holzman.
\newblock Aggregation of non-binary evaluations.
\newblock {\em Advances in Applied Mathematics}, 45(4):487--504, 2010.

\bibitem{dung:1995}
Phan~Minh Dung.
\newblock On the acceptability of arguments and its fundamental role in
  nonmonotonic reasoning, logic programming and n-person games.
\newblock {\em Artificial Intelligence}, 77(2):321--358, 1995.

\bibitem{favardin2002borda}
Pierre Favardin, Dominique Lepelley, and J{\'e}r{\^o}me Serais.
\newblock Borda rule, {C}opeland method and strategic manipulation.
\newblock {\em Review of Economic Design}, 7(2):213--228, 2002.

\bibitem{gartner:2006}
Wulf G\"{a}rtner.
\newblock {\em A Primer on Social Choice Theory}.
\newblock Oxford University Press, 2006.

\bibitem{gibbard:1973}
A.~Gibbard.
\newblock Manipulation of voting schemes.
\newblock {\em Econometrica}, 41:587--601, 1973.

\bibitem{gibbard1977manipulation}
Allan Gibbard.
\newblock Manipulation of schemes that mix voting with chance.
\newblock {\em Econometrica}, 45:665--681, 1977.

\bibitem{gibbard1978straight}
Allan Gibbard.
\newblock Straightforwardness of game forms with lotteries as outcomes.
\newblock {\em Econometrica}, 46:595--614, 1978.

\bibitem{grossi2014judgment}
Davide Grossi and Gabriella Pigozzi.
\newblock {\em Judgment Aggregation: A Primer}.
\newblock Morgan \& Claypool, 2014.

\bibitem{list2010theory}
C.~List.
\newblock The theory of judgment aggregation: An introductory review.
\newblock {\em Synthese}, 187(1):179--207, 2012.

\bibitem{list2010introduction}
Christian List and Ben Polak.
\newblock Introduction to judgment aggregation.
\newblock {\em Journal of economic theory}, 145(2):441--466, 2010.

\bibitem{list:puppe:2009}
Christian List and Clemens Puppe.
\newblock Judgment aggregation: a survey.
\newblock In Paul Anand, Clemens Puppe, and Prasanta Pattanaik, editors, {\em
  The handbook of rational and social choice}. Oxford University Press, Oxford,
  UK, 2009.

\bibitem{modgil2009proof}
Sanjay Modgil and Martin Caminada.
\newblock Proof theories and algorithms for abstract argumentation frameworks.
\newblock In {\em Argumentation in artificial intelligence}, pages 105--129.
  Springer, 2009.

\bibitem{moulin1980strategy}
Herv{\'e} Moulin.
\newblock On strategy-proofness and single peakedness.
\newblock {\em Public Choice}, 35(4):437--455, 1980.

\bibitem{nitzan1985vulnerability}
Shmuel Nitzan.
\newblock The vulnerability of point-voting schemes to preference variation and
  strategic manipulation.
\newblock {\em Public Choice}, 47(2):349--370, 1985.

\bibitem{pan:etal:2010}
Shengying Pan, Kate Larson, and Iyad Rahwan.
\newblock Argumentation mechanism design for preferred semantics.
\newblock {\em Proceedings of the 3rd International Conference on Computational
  Models of Argument (COMMA)}, pages 403--414, 2010.

\bibitem{rahwan:larson:2008}
Iyad Rahwan and Kate Larson.
\newblock Mechanism design for abstract argumentation.
\newblock In L.~Padgham, D.~Parkes, J.~Mueller, and S.~Parsons, editors, {\em
  7th International Joint Conference on Autonomous Agents \& Multi Agent
  Systems, AAMAS'2008, Estoril, Portugal}, pages 1031--1038, 2008.

\bibitem{rahwan:larson:2008a}
Iyad Rahwan and Kate Larson.
\newblock Pareto optimality in abstract argumentation.
\newblock In Dieter Fox and Carla Gomes, editors, {\em Proceedings of the 23rd
  AAAI Conference on Artificial Intelligence (AAAI-2008)}, pages 150--155,
  Menlo Park CA, USA, 2008.

\bibitem{rahwan:etal:2009}
Iyad Rahwan, Kate Larson, and Fernando Tohm\'{e}.
\newblock A characterisation of strategy-proofness for grounded argumentation
  semantics.
\newblock In {\em Proceedings of the 21st International Joint Conference on
  Artificial Intelligence ({IJCAI})}, pages 251--256, 2009.

\bibitem{rahwan:simari:2009}
Iyad Rahwan and Guillermo~R. Simari, editors.
\newblock {\em Argumentation in Artificial Intelligence}.
\newblock Springer, 2009.

\bibitem{rahwan:tohme:2010}
Iyad Rahwan and Fernando Tohm\'{e}.
\newblock {Collective Argument Evaluation as Judgement Aggregation}.
\newblock In {\em {9th International Joint Conference on Autonomous Agents \&
  Multi Agent Systems, {AAMAS}'2010, Toronto, Canada}}, 2010.

\bibitem{ronnegard2015fallacy}
David R{\"o}nnegard.
\newblock {\em The Fallacy of Corporate Moral Agency}.
\newblock Springer, 2015.

\bibitem{satterthwaite:1975}
M.~A. Satterthwaite.
\newblock Strategy-proofness and {A}rrow's conditions: Existence and
  correspondence theorems for voting procedures and social welfare functions.
\newblock {\em Journal of Economic Theory}, 10:187--217, 1975.

\bibitem{vincke1982aggregation}
Philippe Vincke.
\newblock Aggregation of preferences: a review.
\newblock {\em European Journal of Operational Research}, 9(1):17--22, 1982.

\end{thebibliography}
%
%
\clearpage
\section*{Appendix}
This part contains proofs and counterexamples for the results presented in the paper ``Pareto Optimality and Strategy Proofness in Group Argument Evaluation''
\setcounter{section}{0}
\setcounter{theorem}{1}
\setcounter{lemma}{0}
\setcounter{proposition}{0}
\setcounter{example}{0}
\section{General Lemmas} 
{

The following two lemmas are crucial for establishing the relations between the different classes of preferences. The first lemma implies a very interesting result. While Hamming distance and Issue-wise distance based preferences are different, as we showed earlier, Hamming set and Issue-wise set based preferences are equivalent. The same can also be said about IUO Hamming set{s} and IUO Issue-wise set{s}. Hence, all the results in this paper that hold (resp. do not hold) for Hamming set based preferences would also hold (resp. do not hold) for Issue-wise set based preferences. The same can be also said about IUO Hamming set{s} and IUO issue-wise set. 

\begin{lemma}\label{Alem.heterosets}
{Issue-wise set based preferences coincide with Hamming set based preferences, and IUO Issue-wise set{s} based preferences coincide with IUO Hamming set{s} based preferences. Formally, let $\AF=\langle \calA, \rightharpoonup \rangle$ be an argumentation framework, and }let $\calL_1$, $\calL_2$, and $\calL_3$ be three labelings. Then:
\begin{enumerate}
\item $\calL_1 \varominus \calL_2 \subseteq \calL_1 \varominus \calL_3 \Leftrightarrow \calL_1 \varominus_\calW \calL_2 \subseteq \calL_1 \varominus_\calW \calL_3$.

{ (or equivalently $\calL_2 \succeq_{1,\ominus} \calL_3 \Leftrightarrow \calL_2 \succeq_{1,\ominus_\calW} \calL_3$)}
\item $\calL_1 \varominus^\calM \calL_2 \subseteq \calL_1 \varominus^\calM \calL_3 \Leftrightarrow \calL_1 \varominus_\calW^\calM \calL_2 \subseteq \calL_1 \varominus_\calW^\calM \calL_3$.

{ (or equivalently $\calL_2 \succeq_{1,\ominus^\calM} \calL_3 \Leftrightarrow \calL_2 \succeq_{1,\ominus_\calW^\calM} \calL_3$}
\end{enumerate}

\begin{proof}
Let $\calI$ be the set of all issues in $\AF$. 
\begin{enumerate}
\item ($\Rightarrow$): From the definition of issues, we have that $\forall \calB \in \calL_1 \varominus_\calW \calL_2$ (where $\calB \in \calI$):
\[\forall A \in \calA: A \in \calB \Rightarrow A \in \calL_1 \varominus \calL_2 \]
Then, by assumption, we have $A \in \calL_1 \varominus \calL_3$. Hence, we have $\forall A \in \calB: A \in \calL_1 \varominus \calL_3$. Then, $\calB \in \calL_1 \varominus_\calW \calL_3$.

($\Leftarrow$): Consider an arbitrary argument $A$ s.t. $A \in \calL_1 \varominus \calL_2$. Let $\calB \in \calI$ be s.t. $A \in \calB$. Then:
\[\forall A' \neq A: A' \in \calB \Rightarrow A' \in \calL_1 \varominus \calL_2\]
from the definition of issues. This means that $\calB \in \calL_1 \varominus_\calW \calL_2$. By assumption, $\calB \in \calL_1 \varominus_\calW \calL_3$. Then, $A \in \calL_1 \varominus \calL_3$.

\item Similar to (1), but instead of showing from $\varominus$ to $\varominus_\calW$ and vice versa, it is enough to show from $\varominus^{io}$ and $\varominus^{du}$ to $\varominus_\calW^{io}$ and $\varominus_\calW^{du}$ and vice versa, respectively.
\end{enumerate}

{ Using the definitions of set-based preferences yield $\calL_2 \succeq_{1,\ominus} \calL_3 \Leftrightarrow \calL_2 \succeq_{1,\ominus_\calW} \calL_3$ and $\calL_2 \succeq_{1,\ominus^\calM} \calL_3 \Leftrightarrow \calL_2 \succeq_{1,\ominus_\calW^\calM} \calL_3$.}

\end{proof}
\end{lemma}

The following lemma is important in the context of compatible operators. For each agent $i \in \Ag$, let $\calL_i = \calL_1$. Then, provided the conditions below, the lemma says an individual's preference over $\calL_2$ and $\calL_3$ would coincide whether she has a Hamming set (resp. distance) or IUO Hamming set{s} (resp. distance). The same can be said about Issue-wise set (resp. distance) and IUO Issue-wise set{s} (resp. distance).
} 
\begin{lemma}\label{Alem.heterosetscomp}
{Let $\AF=\langle \calA, \rightharpoonup \rangle$ be an argumentation framework. }Let $\calL_1$, $\calL_2$, and $\calL_3$ be three labelings and let $\calL_1 \approx \calL_2$ and $\calL_1 \approx \calL_3$:
\begin{enumerate}
\item $\calL_1 \varominus \calL_2 \subseteq \calL_1 \varominus \calL_3 \Leftrightarrow \calL_1 \varominus^\calM \calL_2 \subseteq \calL_1 \varominus^\calM \calL_3$, and\\
$\calL_1 \left|\varominus\right | \calL_2 \leq \calL_1 \left|\varominus\right | \calL_3 \Leftrightarrow \calL_1 \left|\varominus^\calM\right | \calL_2 \leq \calL_1 \left|\varominus^\calM\right | \calL_3$.

{ (or equivalently $\calL_2 \succeq_{1,\ominus} \calL_3 \Leftrightarrow \calL_2 \succeq_{1,\ominus^\calM} \calL_3$ and $\calL_2 \succeq_{1,|\ominus|} \calL_3 \Leftrightarrow \calL_2 \succeq_{1,|\ominus^\calM|} \calL_3$)}

\item $\calL_1 \varominus_\calW \calL_2 \subseteq \calL_1 \varominus_\calW \calL_3 \Leftrightarrow \calL_1 \varominus_\calW^\calM \calL_2 \subseteq \calL_1 \varominus_\calW^\calM \calL_3$, and\\
$\calL_1 \left|\varominus_\calW\right | \calL_2 \leq \calL_1 \left|\varominus_\calW\right | \calL_3 \Leftrightarrow \calL_1 \left|\varominus_\calW^\calM\right | \calL_2 \leq \calL_1 \left|\varominus_\calW^\calM\right | \calL_3$.

{ (or equivalently $\calL_2 \succeq_{1,\ominus_\calW} \calL_3 \Leftrightarrow \calL_2 \succeq_{1,\ominus_\calW^\calM} \calL_3$ and $\calL_2 \succeq_{1,|\ominus_\calW|} \calL_3 \Leftrightarrow \calL_2 \succeq_{1,|\ominus_\calW^\calM|} \calL_3$)}

\end{enumerate}

\begin{proof}
\begin{enumerate}
\item Since $\calL_1 \approx \calL_2$ then:
\begin{equation}
\neg \exists A \in \calA \text{ s.t. } (\calL_1(A)=\myin \wedge \calL_2(A)=\myout) \vee (\calL_1(A)=\myout \wedge \calL_2(A)=\myin) 
\end{equation}
Therefore $\calL_1 \varominus^{io} \calL_2 = \emptyset$ which implies:

\begin{equation}\label{eq.S12}
\calL_1 \varominus^\calM  \calL_2  = \calL_1 \varominus^{du} \calL_2 = \calL_1 \varominus \calL_2
\end{equation}
and
\begin{equation}\label{eq.D12}
\calL_1 \left |\varominus^\calM \right | \calL_2 = \left  |\calL_1 \varominus^{du}  \calL_2 \right | = \calL_1 \left |\varominus\right | \calL_2
\end{equation}

{Similarly, we can show that:
\begin{equation}\label{eq.S13}
\calL_1 \varominus^\calM \calL_3 = \calL_1 \varominus^{du} \calL_3 = \calL_1 \varominus \calL_3
\end{equation}
and
\begin{equation}\label{eq.D13}
\calL_1 \left |\varominus^\calM \right | \calL_3 = \left  |\calL_1 \varominus^{du}  \calL_3 \right | = \calL_1 \left |\varominus\right | \calL_3
\end{equation}

Now, using Eq.\ref{eq.S12} and Eq.\ref{eq.S13} together, and using Eq.\ref{eq.D12} and Eq.\ref{eq.D13} together, we can prove both directions for the results in 1. Then, the equilvance follows from the definitions of set-based and distance-based preferences.

}

\item Let $\calI$ be the set of all issues in $\AF$. Since $\calL_1 \approx \calL_2$ then:
\begin{equation}
\begin{aligned}[b]
\neg \exists \calB \in \calI \text{ s.t. } (\calL_1(A)=\myin \wedge \calL_2(A)=\myout) \vee (\calL_1(A)=\myout \wedge \calL_2(A)=\myin) \\ \text{ for some (equiv. all) } A \in \calB
\end{aligned}
\end{equation}
The rest is similar.
\end{enumerate}
\end{proof}
\end{lemma}

The following lemmas are also crucial for the proofs of theorems in this paper. Since the labelings have only three values, we can use the following lemma.

\begin{lemma} \label{Alem.hamSetFormulas}
Let $\AF=\langle \calA, \rightharpoonup \rangle$ be an argumentation framework. Let $\mydec(\calL) = \myin(\calL) \cup \myout(\calL)$ $\forall \calL \in \labs$. For any pair $\calL_1,\calL_2 \in \labs$:

\begin{itemize}
	\item[a)] {$\calL_1 \varominus \calL_2 = 
	(\myin(\calL_1) \cap \myout(\calL_2)) \cup
	(\myin(\calL_1) \cap \myundec(\calL_2)) \cup
	(\myout(\calL_1) \cap \myin(\calL_2)) \cup
	(\myout(\calL_1) \cap \myundec(\calL_2)) \cup
	(\myundec(\calL_1) \cap \myin(\calL_2)) \cup
	(\myundec(\calL_1) \cap \myout(\calL_2))$}
	\item[b)] if $\calL_1 \sqsubseteq \calL_2$ then $\calL_1 \varominus \calL_2 = \myundec(\calL_1) \cap \mydec(\calL_2)$
	\item[c)] if $\calL_1 \approx \calL_2$ then {$\calL_1 \varominus \calL_2 = (\mydec(\calL_1) \cap \myundec(\calL_2)) \cup (\myundec(\calL_1) \cap \mydec(\calL_2))$}
\end{itemize}

\begin{proof}$\;$

\begin{itemize}
\item[a)] {This} follows from the fact that $\myin(\calL)$, $ \myout(\calL)$ and $\myundec(\calL)$ partition the domain of any labeling $\calL$.
{
\item[b)] From $\calL_1 \sqsubseteq \calL_2$, the sets $(\myin(\calL_1) \cap \myout(\calL_2))$, $(\myin(\calL_1) \cap \myundec(\calL_2))$, $(\myout(\calL_1) \cap \myin(\calL_2))$, and $(\myout(\calL_1) \cap \myundec(\calL_2))$ are all empty sets. Then, we are left with the following:
\[(\myundec(\calL_1) \cap \myin(\calL_2)) \cup	(\myundec(\calL_1) \cap \myout(\calL_2))\]
which can be written as:
\[\myundec(\calL_1) \cap (\myin(\calL_2) \cup \myout(\calL_2))\]
and replacing $\myin(\calL) \cup \myout(\calL)$ by $\mydec(\calL)$ would give the result.

\item[c)] From $\calL_1 \approx \calL_2$, the sets $(\myin(\calL_1) \cap \myout(\calL_2))$, and $(\myout(\calL_1) \cap \myin(\calL_2))$ are empty. The rest can be rearranged similarly to b), and replacing $\myin(\calL) \cup \myout(\calL)$ by $\mydec(\calL)$ would give the result.
}
\end{itemize}
\end{proof}
\end{lemma}
We now prove two lemmas establishing the relations between less or equally committed labelings and Hamming based preferences over labelings.

\begin{lemma}
\label{Alem.moreCommPref}
{Let $\AF=\langle \calA, \rightharpoonup \rangle$ be an argumentation framework. }Let $\calL$, $\calL'$ and $\calL_i$ be three labelings such that $\calL \sqsubseteq \calL' \sqsubseteq \calL_i$. If $\calL_i$ is the most preferred labeling of agent $i$ and her preference is Hamming set or Hamming distance based, then $\calL' \succeq_{i, \varominus} \calL$ and $\calL' \succeq_{i, \left |\varominus\right |} \calL$ respectively.

\begin{proof}
From $\calL \sqsubseteq \calL'$, we have that $\mydec(\calL) \subseteq \mydec(\calL')$, which is equivalent to $\myundec(\calL')$ $\subseteq \myundec(\calL)$ because $\myundec$ is the complement of $\mydec$. From this, it follows that $\myundec(\calL') \cap \mydec(\calL_i) \subseteq \myundec(\calL) \cap \mydec(\calL_i)$. Since $\calL \sqsubseteq \calL_i$ and $\calL' \sqsubseteq \calL_i$ (by assumption and transitivity of $\sqsubseteq$), we can use Lemma \ref{Alem.hamSetFormulas}b to obtain $\calL' \varominus \calL_i \subseteq \calL \varominus \calL_i$. 
By definition we have that $\calL' \succeq_{i, \varominus} \calL$ and $\calL' \succeq_{i, \left |\varominus\right |} \calL$.
\end{proof}
\end{lemma}

\begin{lemma}
\label{Alem.prefSetMoreComm}
{Let $\AF=\langle \calA, \rightharpoonup \rangle$ be an argumentation framework. }Let $\calL$, $\calL'$ and $\calL_i$ be three labelings and let $\calL \sqsubseteq \calL_i$. If $\calL_i$ is the most preferred labeling of agent $i$, her preference is Hamming set based and $\calL' \succeq_{i, \varominus} \calL$, then $\calL \sqsubseteq \calL'$.

\begin{proof}
$\calL' \succeq_{i, \varominus} \calL$ implies $\calL' \varominus \calL_i \subseteq \calL \varominus \calL_i$ which implies $\calL(A) = \calL_i(A) \Rightarrow \calL'(A) = \calL_i(A)$ for any argument $A$ (i). Now, $\calL \sqsubseteq \calL_i$ implies $\calL(A) = \calL_i(A)$ for any $A \in \mydec(\calL)$ (ii). From (i) and (ii) it follows that $\calL(A) = \calL'(A)$ for any $A \in \mydec(\calL)$. Hence $\calL \sqsubseteq \calL'$.
\end{proof}
\end{lemma}

\section{Pareto Optimality}\label{Asec.Pa}

{
\begin{theorem}\label{Athm.HdHs}
Let $\otimes \in \{\ominus,\ominus^\calM,\ominus_\calW,\ominus^\calM_\calW\}$ be a set measure and $|\otimes|$ be its corresponding distance measure (i.e. if $\otimes = \ominus^\calM$ then $|\otimes|=|\ominus^\calM|$). If a labeling is Pareto optimal in a set $\calS$ given agents with $|\otimes|$-based preferences, then it is Pareto optimal in $\calS$ given agents with $\otimes$-based preferences. 
\begin{proof}
Let $\calS$ be a set of labelings, and $\calL$ be a labeling that is Pareto optimal in $\calS$ given agents with $|\otimes|$-based preferences. Suppose, towards a contradiction, that $\calL$ is not Pareto optimal in $\calS$ given agents with $\otimes$-based preferences. Then, $\exists \calL_X \in \calS$ such that:
\begin{equation}
(\forall i \in \Ag \text{: } \calL_X \succeq_{i,\otimes} \calL) \wedge (\exists j \in \Ag \text{ s.t. } \calL_X \succ_{j,\otimes} \calL) 
\end{equation}

From the definition of strict preferences $\succ$:

\begin{equation}
(\forall i \in \Ag \text{: } \calL_X \succeq_{i,\otimes} \calL) \wedge (\exists j \in \Ag \text{ s.t. } \calL_X \succeq_{j,\otimes} \calL \wedge \neg \calL \succeq_{j,\otimes} \calL_X)
\end{equation}

From the definition of set-based preference:
\begin{equation}
(\forall i \in \Ag \text{: } \calL_X \otimes \calL_i \subseteq \calL \otimes \calL_i) \wedge (\exists j \in \Ag \text{ s.t. } \calL_X \otimes \calL_j \subset \calL \otimes \calL_j) 
\end{equation}

This implies:
\begin{equation}
(\forall i \in \Ag \text{: } \calL_X \left |\otimes \right | \calL_i \leq \calL \left |\otimes \right | \calL_i) \wedge (\exists j \in \Ag \text{ s.t. } \calL_X \left |\otimes \right | \calL_j < \calL \left |\otimes \right | \calL_j) 
\end{equation}
Which means that $\calL$ is not Pareto optimal in $\calS$ given agents with $|\otimes|$-based preferences. Contradiction.
\end{proof}
\end{theorem}
}


\begin{theorem}\label{Athm.MHaPaSo}
Let $\calX$ be the set of all admissible labelings that are compatible ($\approx$) with each of the participants' individual labelings. Let $\calS$ be any arbitrary set such that $\calS \subseteq \calX$. A labeling from $\calS$ is Pareto optimal in $\calS$ when individual preferences are Hamming set (resp. distance) based iff it is Pareto optimal in $\calS$ when individual preferences are IUO Hamming set{s} (resp. distance) based.
\begin{proof}
($\Rightarrow$): Let $\calL$ and $\calL'$ be two labelings in $\calS$. Suppose $\calL$ is Pareto optimal in $\calS$ when agents have Hamming set based preferences. Then{, there is no labeling $\calL_X$ that Pareto dominates $\calL$ w.r.t Hamming set based preferences}:

\begin{equation}
\begin{aligned}[b]
\neg \exists \calL_X \in \calS \text{ s.t. }(\forall i \in \Ag \text{ : } \calL_X \varominus \calL_i \subseteq \calL \varominus \calL_i) \wedge (\exists j \in \Ag \text{ s.t. } \calL \varominus \calL_j \not\subseteq \calL_X \varominus \calL_j)
\end{aligned}
\end{equation}

Note that since $\calL,\calL_X \in \calS \subseteq \calX$, then $\calL \approx \calL_i$, and $\calL_X \approx \calL_i$, $\forall i \in \Ag$, then by using Lemma \ref{Alem.heterosetscomp} (1) for each label $\calL_i$, this is equivalent to:
\begin{equation}
\begin{aligned}[b]
\neg \exists \calL_X \in \calS \text{ s.t. }(\forall i \in \Ag \text{ : } \calL_X \varominus^\calM \calL_i \subseteq \calL \varominus^\calM \calL_i) \wedge (\exists j \in \Ag \text{ s.t. } \calL \varominus^\calM \calL_j \not\subseteq \calL_X \varominus^\calM \calL_j)
\end{aligned}
\end{equation}

Similarly, suppose $\calL'$ is Pareto optimal in $\calS$ when agents have Hamming distance based preferences. Then{, there is no labeling $\calL_X$ that Pareto dominates $\calL$ w.r.t Hamming distance based preferences}:
\begin{equation}
\begin{aligned}[b]
\neg \exists \calL_X \in \calS \text{ s.t. }(\forall i \in \Ag \text{ : } \calL_X \left|\varominus\right| \calL_i \leq \calL' \left|\varominus \right| \calL_i) \wedge (\exists j \in \Ag \text{ s.t. } \calL' \left|\varominus\right| \calL_j \not\leq \calL_X \left|\varominus\right| \calL_j)
\end{aligned}
\end{equation}

Also by using Lemma \ref{Alem.heterosetscomp} (1) for each label $\calL_i$, this is equivalent to:
\begin{equation}
\begin{aligned}[b]
\neg \exists \calL_X \in \calS \text{ s.t. }(\forall i \in \Ag \text{ : } \calL_X \left|\varominus^\calM\right| \calL_i \leq \calL' \left|\varominus^\calM \right| \calL_i) \wedge (\exists j \in \Ag \text{ s.t. } \calL' \left|\varominus^\calM\right| \calL_j \not\leq \calL_X \left|\varominus^\calM\right| \calL_j)
\end{aligned}
\end{equation}

($\Leftarrow$): Similar to ($\Rightarrow$)
\end{proof}
\end{theorem}

\begin{theorem}\label{Athm.MIwPaSo}
Let $\calX$ be the set of all admissible labelings that are compatible ($\approx$) with each of the participants' individual labelings. Let $\calS$ be any arbitrary set such that $\calS \subseteq \calX$. A labeling from $\calS$ is Pareto optimal in $\calS$ when individual preferences are Issue-wise set (resp. distance) based iff it is Pareto optimal in $\calS$ when individual preferences are IUO Issue-wise set{s} (resp. distance) based.
\begin{proof}
($\Rightarrow$): Let $\calI$ be the set of all issues in $\AF$, and let $\calL$ and $\calL'$ be two labelings in $\calS$. Suppose $\calL$ is Pareto optimal in $\calS$ when agents have Issue-wise set based preferences. Then{, there is no labeling $\calL_X$ that Pareto dominates $\calL$ w.r.t Issue-wise set based preferences}:

\begin{equation}
\begin{aligned}[b]
\neg \exists \calL_X \in \calS \text{ s.t. }(\forall i \in \Ag \text{ : } \calL_X \varominus_\calW \calL_i \subseteq \calL \varominus_\calW \calL_i) \wedge (\exists j \in \Ag \text{ s.t. } \calL \varominus_\calW \calL_j \not\subseteq \calL_X \varominus_\calW \calL_j)
\end{aligned}
\end{equation}

The rest is similar to the proof of Theorem \ref{Athm.MHaPaSo} (using ``issues'' instead of ``arguments'' with the help of Lemma \ref{Alem.heterosetscomp} (2)).

\end{proof}
\end{theorem}

The following two examples are used to show the results discussed in the subsection \emph{3.1.3 Failed Connections}.

\begin{example}\label{Aexm.notHaDist}
Consider Figure \ref{Afig:PoGen}, let $\Ag = \{1,2\}$ be the set of agents $1$ and $2$, whose preferred labelings are respectively $\calL_1$ and $\calL_2$, and let $\calS = \{\calL_{CO},\calL_X\}$. Note the following:

\begin{itemize}
\item $\calL_{CO}$ is Pareto optimal in $\calS$ given agents with Hamming (Issue-wise) set based preferences. 
\item $\calL_{CO}$ is Pareto optimal in $\calS$ given agents with Issue-wise distance based preferences.
\item $\calL_{CO}$ is not Pareto optimal in $\calS$ given agents with Hamming distance based preferences.
\end{itemize} 
Since both $\calL_{CO}$ and $\calL_X$ are admissible labelings that are compatible ($\approx$) with both of $\calL_1$ and $\calL_2$, then using Theorems \ref{Athm.MHaPaSo} and \ref{Athm.MIwPaSo}:
\begin{itemize}
\item $\calL_{CO}$ is Pareto optimal in $\calS$ given agents with IUO Hamming (Issue-wise) set based preferences. 
\item $\calL_{CO}$ is Pareto optimal in $\calS$ given agents with IUO Issue-wise distance based preferences.
\item $\calL_{CO}$ is not Pareto optimal in $\calS$ given agents with IUO Hamming distance based preferences.
\end{itemize} 
\end{example}

\begin{figure}[ht]
    \centering
  \includegraphics[scale=0.4]{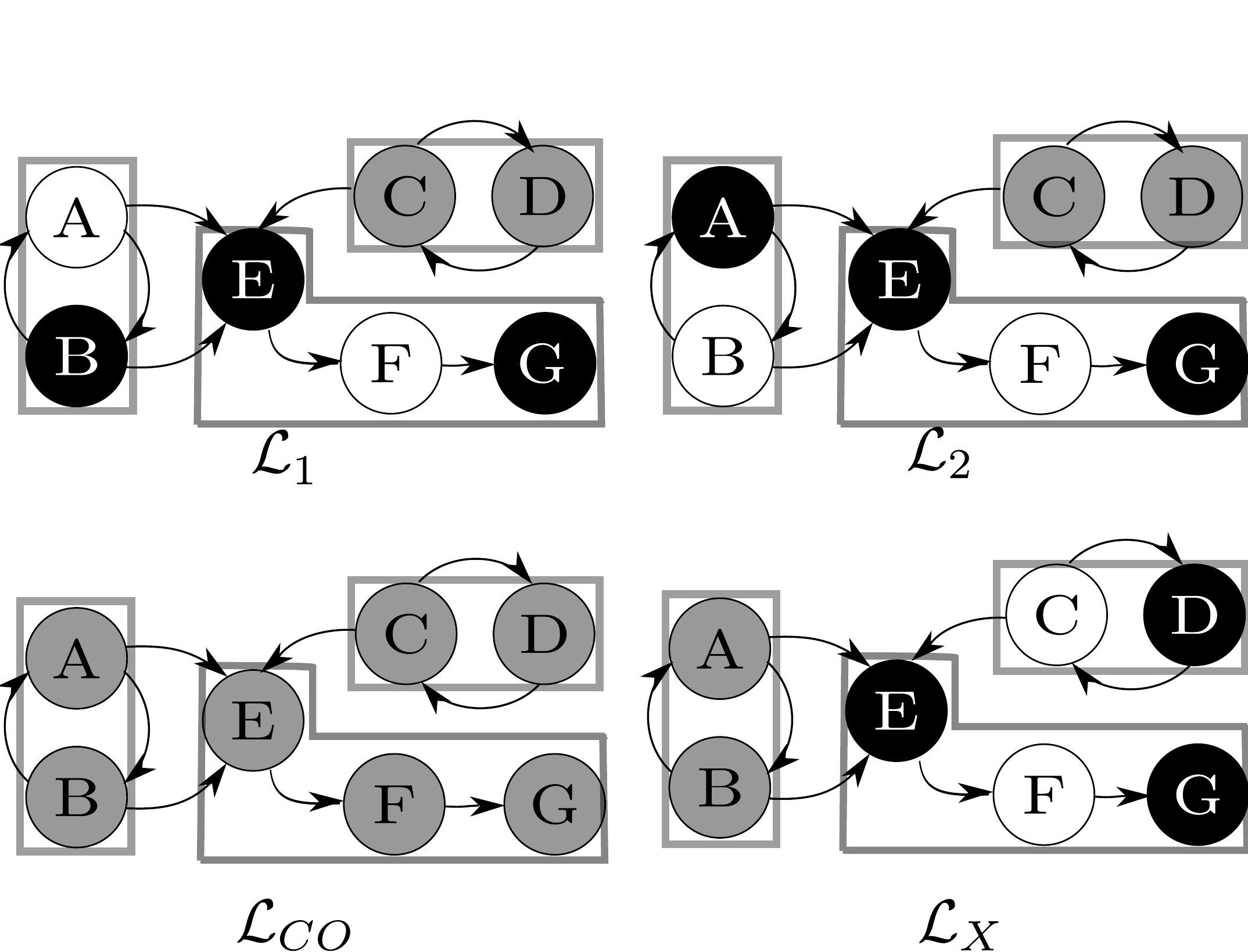}
  \caption{An example showing how Pareto optimality given agents with (IUO) Hamming distance based preferences cannot be inferred from other classes of preferences. The set of arguments located in one box form an issue.}\label{Afig:PoGen}
\end{figure}

Also consider the following example.

\begin{example}\label{Aexm.notIwDist}
Consider Figure \ref{Afig:PoGen2}, let $\Ag = \{1,2\}$ be the set of agents $1$ and $2$, whose preferred labelings are respectively $\calL_1$ and $\calL_2$, and let $\calS = \{\calL_{CO},\calL_X\}$. Note the following:

\begin{itemize}
\item $\calL_{CO}$ is Pareto optimal in $\calS$ given agents with Hamming (Issue-wise) set based preferences. 
\item $\calL_{CO}$ is Pareto optimal in $\calS$ given agents with Hamming distance based preferences.
\item $\calL_{CO}$ is not Pareto optimal in $\calS$ given agents with Issue-wise distance based preferences.
\end{itemize} 
Also, since both $\calL_{CO}$ and $\calL_X$ are admissible labelings that are compatible ($\approx$) with both of $\calL_1$ and $\calL_2$, then using Theorems \ref{Athm.MHaPaSo} and \ref{Athm.MIwPaSo}:
\begin{itemize}
\item $\calL_{CO}$ is Pareto optimal in $\calS$ given agents with IUO Hamming (Issue-wise) set based preferences. 
\item $\calL_{CO}$ is Pareto optimal in $\calS$ given agents with IUO Hamming distance based preferences.
\item $\calL_{CO}$ is not Pareto optimal in $\calS$ given agents with IUO Issue-wise distance based preferences.
\end{itemize} 

\end{example}

\begin{figure}[ht]
    \centering
  \includegraphics[scale=0.4]{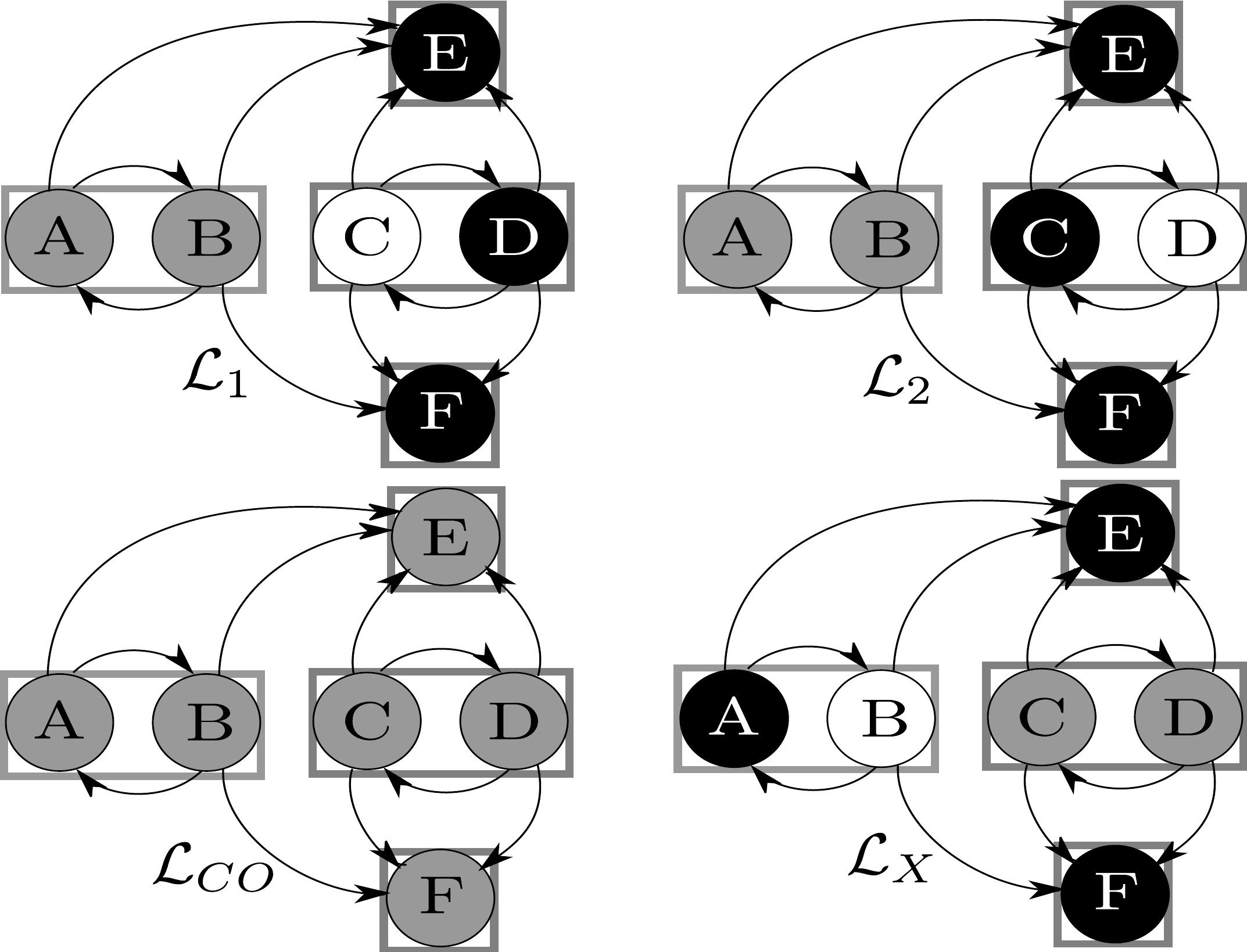}
  \caption{An example showing how Pareto optimality given agents with (IUO) Issue-wise distance based preferences cannot be inferred from other classes of preferences. The set of arguments located in one box form an issue.}\label{Afig:PoGen2}
\end{figure}

Following, we show the results about Pareto optimality regarding the three operators: the skeptical, the credulous and the super credulous.

\begin{theorem}\label{Athm.HaPaSoSet}
If individual preferences are Hamming set based, then the skeptical aggregation operator is Pareto optimal {in the set of} admissible labelings that are smaller or equal ($\sqsubseteq$) to each of the participants' labelings.
\begin{proof}
Let $P$ be a profile of labelings, $\calL_{SO} = so_{\AF}(P)$ and $\calL_X$ some admissible labeling with the property $\forall i \in \Ag, \calL_X \sqsubseteq \calL_i$. From Theorem 1 we know that $\calL_{SO}$ is the biggest admissible labeling with such property, so $\calL_X \sqsubseteq \calL_{SO}$. So we have $\forall i \in \Ag, \calL_X \sqsubseteq \calL_{SO} \sqsubseteq \calL_i$. From Lemma \ref{Alem.moreCommPref} we have $\calL_{SO} \succeq_i \calL_X$ for any $i$. So no agent strictly prefers $\calL_X$ and hence there is no labeling that Pareto dominates $\calL_{SO}$.  
\end{proof}

\end{theorem}

\begin{theorem}\label{Athm.HaPaSoDis}
If individual preferences are Hamming distance based, then the skeptical aggregation operator is Pareto optimal {in the set of} admissible labelings that are smaller or equal ($\sqsubseteq$) to each of the participants' labelings.
\begin{proof}
In the proof of Theorem \ref{Athm.HaPaSoSet}, Hamming set may be replaced by Hamming distance because it is only used in Lemma \ref{Alem.moreCommPref}, which works for Hamming distance as well.
\end{proof}
\end{theorem}

\begin{theorem}\label{Athm.HaPaCoSet}
If individual preferences are Hamming set based, then the credulous aggregation operator is Pareto optimal {in the set of} admissible labelings that are compatible ($\approx$) with each of the participants' labelings.
\begin{proof}
Let $P$ be a profile of labelings, $\calL_{CO} = co_{\AF}(P)$, $\calL_{CIO} = cio_{\AF}(P)$. Assume by contradiction that there exists some admissible labeling $\calL_X$ with the property $\forall i \in \Ag, \calL_X \approx \calL_i$ that Pareto dominates $\calL_{CO}$.

First notice that compatibility ensures that there are no $\myin / \myout$ conflicts between $\calL_X$ and $\calL_{CO}$. If there is an $\myin / \myout$ conflict between agents' labelings on some argument, then both $\calL_X$ and $\calL_{CO}$ need to label it $\myundec$. If there exists an agent whose labeling decides on some argument and other agents' labelings agree or {refrain} from decision, $\calL_{CO}$ and $\calL_X$ also agree or {refrain} from decision. If all agents {refrain} from decision on some argument, $\calL_{CO}$ by definition also {refrains}, and $\calL_X$ may label freely. 

Let us take $A \in \mydec(\calL_X)$. Then, there needs to be an agent with a labeling that agrees on $A$. Otherwise all agents' labelings would be undecided on such argument and, according to definition, $\calL_{CO}$ would not decide either. But then all agents' labelings will agree on such argument with $\calL_{CO}$ and disagree with $\calL_X$, so no agent will strongly prefer $\calL_X$, which contradicts with domination. So there exists at least one agent whose labeling agrees with  $\calL_X$ on $A$. Other agents' labelings also need to agree on $A$ or label it $\myundec$ because of the compatibility of $\calL_X$. Then by definition $\calL_{CIO}(A) = \calL_{X}(A)$. This holds for any argument $A \in {\mydec}(\calL_X)$, so we have $\calL_X \sqsubseteq \calL_{CIO}$. But $\calL_X$ is admissible and, by definition, $\calL_{CO}$ is the biggest admissible labeling less or equally committed as $\calL_{CIO}$. So we have $\calL_X \sqsubseteq \calL_{CO} \sqsubseteq \calL_{CIO}$.  

$\calL_X$ must be different from $\calL_{CO}$ to dominate it. Let $A$ be an argument on which these labelings differ. From the previous {considerations}, it follows that $A \in \myundec(\calL_X)$ and $A \in \mydec(\calL_{CO})$. $\calL_{CO}$ decides on an argument only if there exists an agent that decides on such argument. But then this agent will agree on $A$ with $\calL_{CO}$ and disagree with $\calL_X$, so it will not prefer $\calL_X$. This is in contradiction with dominance. Hence, such dominating labeling cannot exist.
\end{proof}
\end{theorem}

\begin{observation}\label{Aobs.HaPaCoDis}
If individual preferences are Hamming distance based, then the credulous (resp. the super credulous) aggregation operator is not Pareto optimal {in the set of} admissible (resp. complete) labelings that are compatible ($\approx$) with each of the participants' labelings. An example is given in Figure \ref{Afig:HSco} where $\calL_{CO}$ represents the outcome of the credulous (or the super credulous) aggregation operator. Both labelings $\calL_{CO}$ and $\calL_X$ are compatible with both $\calL_1$ and $\calL_2$, but $\calL_X$ is closer when applying Hamming distance. $\calL_1 \varominus \calL_{CO} = \calL_2 \varominus \calL_{CO} = \{A, B, E, F, G\}$, so the Hamming distance is 5, whereas $\calL_1 \varominus \calL_X = \calL_2 \varominus \calL_X = \{A, B, C, D\}$, so the Hamming distance is 4.

\end{observation}

\begin{figure}[ht]
    \centering
  \includegraphics[scale=0.4]{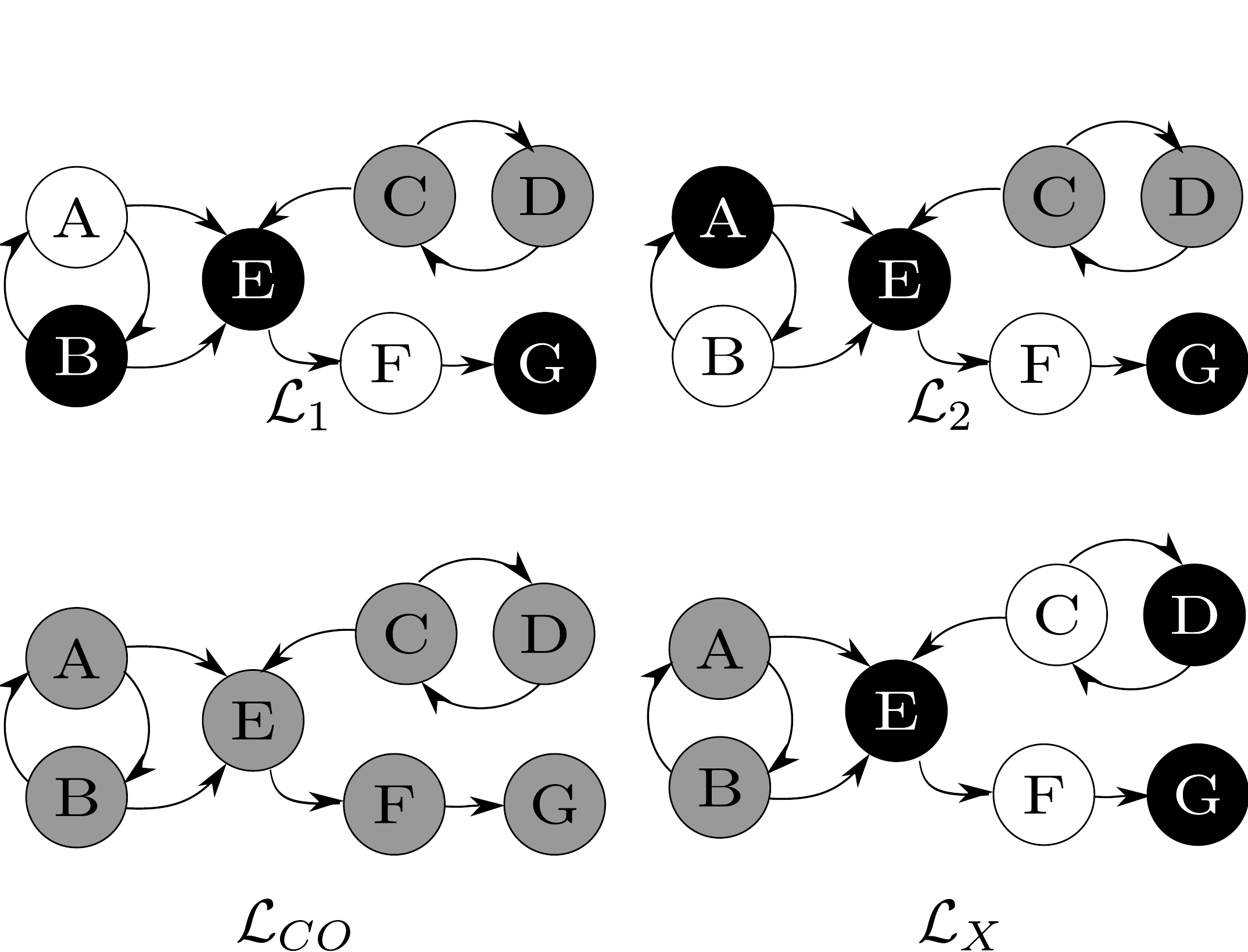}
  \caption{If individuals' preferences are Hamming distance based, the (super) credulous aggregation operator is not Pareto optimal {in the set of admissible (resp. complete) labelings that are compatible ($\approx$) with each of the participants' labelings.}}\label{Afig:HSco}
\end{figure}

\begin{theorem}\label{Athm.HaPaScoSet}
{If individual preferences are Hamming set based, then the super credulous aggregation operator is Pareto optimal {in the set of} complete labelings that are compatible ($\approx$) with each of the participants' labelings.}

\begin{proof}
Let $P$ be a profile of labelings, $\calL_{CIO}=cio_{\AF}(P)$, $\calL_{CO}=co_{\AF}(P)$, and $\calL_{SCO}=sco_{\AF}(P)$. Suppose, towards a contradiction, that there exists a complete labeling $\calL_X$ s.t. $\calL_X \approx \calL_i$ $\forall i \in \Ag$, and $\calL_X$ dominates $\calL_{SCO}$ (w.r.t $\succeq_{i,\varominus}$).

Let $A \in \mydec(\calL_{CO})$, then $\calL_{SCO}$ agrees on $A$ with $\calL_{CO}$. However, $\calL_{CO}$ only decides on an argument if at least one agent decides on this argument and agrees with $\calL_{CO}$ on it. Then, this agent also agrees on $A$ with $\calL_{SCO}$. Since $\calL_X$, by assumption, Pareto dominates $\calL_{SCO}$, $\calL_X$ also needs to agree with this agent on $A$. This is the case for every argument $A \in \mydec(\calL_{CO})$. Hence, $\forall A \in \mydec(\calL_{CO}): \calL_{CO}(A) = \calL_X(A)$. Then, $\calL_{CO} \sqsubseteq \calL_X$. By definition, $\calL_{SCO}$ is the smallest element (w.r.t $\sqsubseteq$) of the set of all complete labelings that are bigger or equally committed than $\calL_{CO}$. Then, $\calL_{CO} \sqsubseteq \calL_{SCO} \sqsubseteq \calL_X$.  

$\calL_X$ should be different from $\calL_{SCO}$ to dominate it. Then, $\exists A \in \myundec(\calL_{SCO}) \cap \mydec(\calL_X)$. We will show that $\forall A \in \myundec(\calL_{SCO}) \cap \mydec(\calL_X)$ then $\forall i \in \Ag: \calL_i(A) = \myundec$. This is enough to reach a contradiction because it shows that all agents agree on at least one argument with $\calL_{SCO}$ while disagree with $\calL_X$ on that argument.

Suppose, for contradiction, that there exists an agent $j$ such that $\calL_j(A) = \calL_X(A) \in \{\myin,\myout\}$. Since $A \in \myundec(\calL_{SCO})$, then $A \in \myundec(\calL_{CO})$. However, $\calL_X$ is a complete labeling which means that it is also an admissible labeling, {and from Theorem \ref{Athm.HaPaCoSet}, $\calL_{CO}$ is Pareto optimal in the set of all admissible labelings that are compatible ($\approx$) with each of the participants' labelings}. Then:
\begin{equation}
\forall B \in \calA, \neg \exists i \in \Ag \text{ s.t. } \calL_{CO}(B) \neq \calL_i(B) \wedge \calL_X(B) = \calL_i(B)
\end{equation}
Contradiction. Then, all agents need to agree with $\calL_{CO}$ and $\calL_{SCO}$ on every $A$ s.t. $A \in \myundec(\calL_{SCO}) \cap \mydec(\calL_X)$ (and disagree with $\calL_X$ on $A$).
\end{proof}
\end{theorem}


\begin{theorem}\label{Athm.IwPaSoDis}
{If individual preferences are Issue-wise distance based, then the skeptical aggregation operator is Pareto optimal in the set of admissible labelings that are smaller or equal (w.r.t $\sqsubseteq$) to each of the participants' labelings.}
\begin{proof}
Let $\calI$ be the set of all issues in $\AF$, $P$ be a profile of labelings, and $\calL_{SO}=so_{\AF}(P)$. Suppose, towards a contradiction, that there exists an admissible labeling $\calL_X$ s.t. $\calL_X \sqsubseteq \calL_i$ $\forall i \in \Ag$, and $\calL_X$ dominates $\calL_{SO}$ (w.r.t $\succeq_{i,\left |\varominus_\calW\right |}$). Then, {there needs to be at least one issue on which $\calL_X$ agrees with some labeling $\calL_j$ (by an agent $j$), while $\calL_{SO}$ disagrees with $\calL_j$ on that issue:}
\begin{equation}
\begin{aligned}[b]
\exists j \in \Ag, \exists \calB \in \calI \text{ s.t. } (\calL_X(A) = \calL_j(A)) \wedge (\calL_{SO}(A) \neq \calL_j(A)) \\\text{ for some (equiv. all) } A \in \calB
\end{aligned}
\end{equation}

However, from Theorem 1, $\calL_{SO}$ is the biggest labeling (w.r.t $\sqsubseteq$) in $\calX$. Then $\calL_X \sqsubseteq \calL_{SO} \sqsubseteq \calL_i$ $\forall i \in \Ag$. $\calL_X$ should be different from $\calL_{SO}$ to dominate it. Then, $\exists A \in \myundec(\calL_X) \cap \mydec(\calL_{SO})$ {(where $A$ belongs to some issue $\calB \in \calI$)}. However, $\calL_{SO}$ only decides on an argument if all agents decide on this argument and agree on it with $\calL_{SO}$. Accordingly, all agents disagree with $\calL_X$ on $A$. Note that this holds for all $A \in \myundec(\calL_X) \cap \mydec(\calL_{SO})$. Additionally, $\forall B \notin \myundec(\calL_X) \cap \mydec(\calL_{SO}): \calL_{SO}(B) = \calL_X(B)$. Hence:
\begin{equation}
\neg \exists \calL_X \in \calX \text{ s.t. } \exists j \in \Ag:  (\calL_X(A) = \calL_j(A)) \wedge (\calL_{SO}(A) \neq \calL_j(A)) \text{ for any } A \in \calA
\end{equation}

Contradiction.
\end{proof}
\end{theorem}

\begin{observation}\label{Aobs.IwPaCoSco}
If individual preferences are Issue-wise distance based, then the credulous (resp. the super credulous) aggregation operator is not Pareto optimal {in the set of} admissible (resp. complete) labelings that are compatible ($\approx$) with each of the participants' labelings. In Figure \ref{Afig:CoSco}, $\calL_{CO}$ represents the outcome of the credulous (or the super credulous) aggregation operator. Note that, both labelings of $\calL_{CO}$ and $\calL_X$ are compatible with both $\calL_1$ and $\calL_2$, but $\calL_X$ is closer when applying Issue-wise distance. $\calL_1 \varominus_\calW \calL_{CO}=$ $\calL_2 \varominus_\calW \calL_{CO}=$ $\{\{C,D\},\{E\},\{F\}\}$, so Issue-wise distance is $3$, whereas $\calL_1 \varominus_\calW \calL_X=$ $\calL_2 \varominus_\calW \calL_{X}=$ $\{\{A,B\},\{C,D\}\}$, so Issue-wise distance is $2$.
\end{observation}

\begin{figure}[ht]
    \centering
  \includegraphics[scale=0.4]{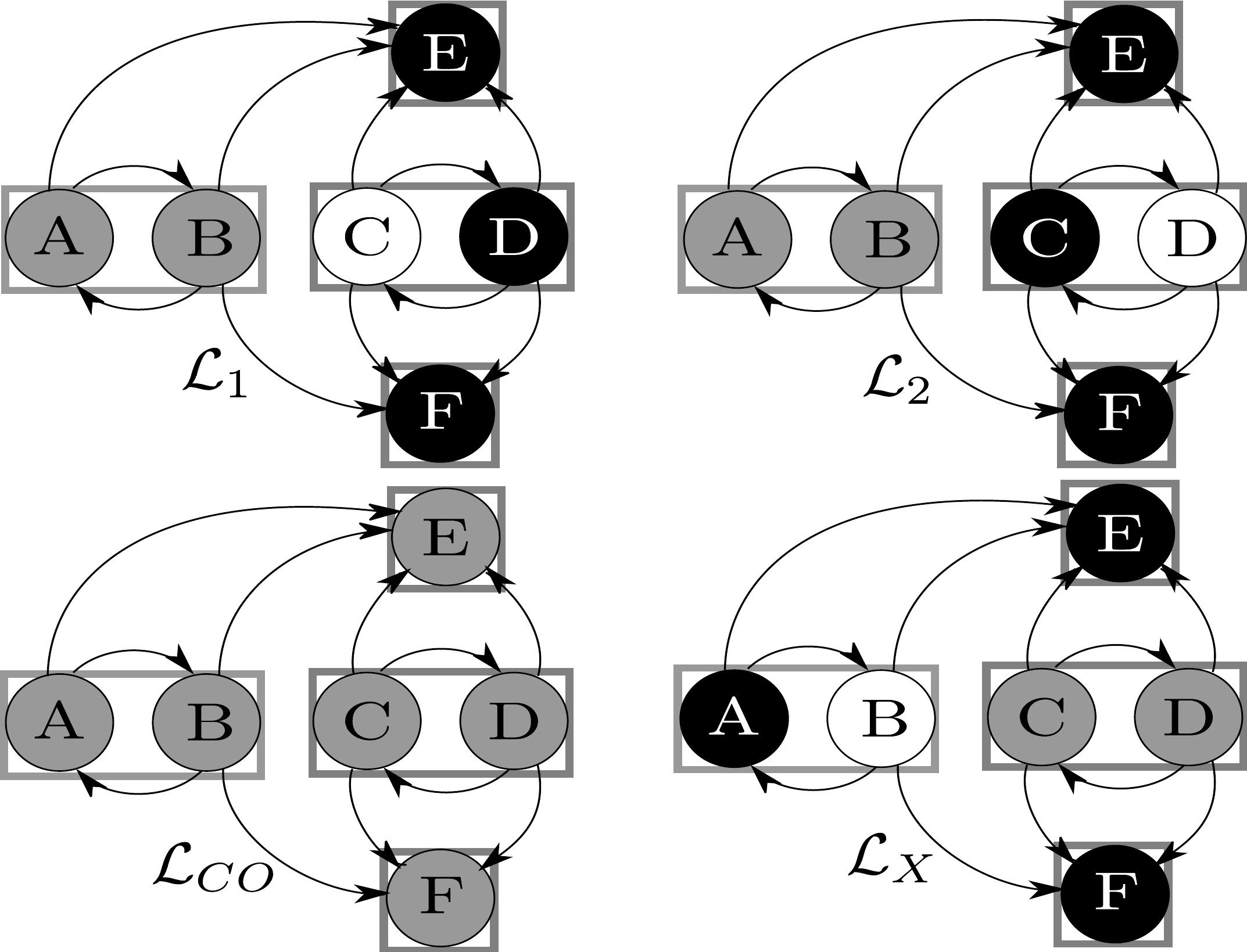}
  \caption{If individuals' preferences are Issue-wise distance based, the (super) credulous aggregation operator is not Pareto optimal {in the set of admissible (resp. complete) labelings that are compatible ($\approx$) with each of the participants' labelings.} The set of arguments located in one box form an issue.}\label{Afig:CoSco}
\end{figure}



{
\begin{proposition}\label{Aprop.MHaPaSo}
If individual preferences are IUO Hamming set{s} (resp. distance) based, then the skeptical aggregation operator is Pareto optimal {in the set of} admissible labelings that are smaller or equal ($\sqsubseteq$) to each of the participants' labelings.
\begin{proof}
Let $\calS$ be the set of admissible labelings that are smaller or equal ($\sqsubseteq$) to each of the participants' labelings. Then, $\calS \subseteq \calX$, where $\calX$ is defined in Theorem \ref{Athm.MHaPaSo}. From Theorem \ref{Athm.HaPaSoSet} and Theorem \ref{Athm.MHaPaSo} the skeptical aggregation operator is Pareto optimal in $\calS$ when individual preferences are IUO Hamming set{s} based. From Theorem \ref{Athm.HaPaSoDis} and Theorem \ref{Athm.MHaPaSo} the skeptical aggregation operator is Pareto optimal in $\calS$ when individual preferences are IUO Hamming distance based.
\end{proof}
\end{proposition}

\begin{proposition}\label{Aprop.MHaPaSoSet}
If individual preferences are IUO Hamming set{s} based, then the credulous aggregation operator is Pareto optimal {in the set of} admissible labelings that are compatible ($\approx$) with each of the participants' labelings.
\begin{proof}
Let $\calS$ be the set of admissible labelings that are compatible ($\approx$) with each of the participants' labelings. Then, $\calS \subseteq \calX$, where $\calX$ is defined in Theorem \ref{Athm.MHaPaSo} (actually $\calS = \calX$ here). From Theorem \ref{Athm.HaPaCoSet} and Theorem \ref{Athm.MHaPaSo} the credulous aggregation operator is Pareto optimal in $\calS$ when individual preferences are IUO Hamming set{s} based.
\end{proof}
\end{proposition}

\begin{proposition}\label{Aprop.MHaPaSoDis}
If individual preferences are IUO Hamming distance based, then the credulous aggregation operator is not Pareto optimal {in the set of} admissible labelings that are compatible ($\approx$) with each of the participants' labelings.
\begin{proof}
Similar to the previous proposition, from Observation \ref{Aobs.HaPaCoDis} and Theorem \ref{Athm.MHaPaSo} the credulous aggregation operator is not Pareto optimal in $\calS$ ($\calS$ is defined in the previous proposition) when individual preferences are IUO Hamming distance based.
\end{proof}
\end{proposition}

\begin{proposition}\label{Aprop.MHaPaScoSet}
If individual preferences are IUO Hamming set{s} based, then the super credulous aggregation operator is Pareto optimal {in the set of} complete labelings that are compatible ($\approx$) with each of the participants' labelings.
\begin{proof}
Let $\calS$ be the set of complete labelings that are compatible ($\approx$) with each of the participants' labelings. Then, $\calS \subseteq \calX$, where $\calX$ is defined in Theorem \ref{Athm.MHaPaSo}. From Theorem \ref{Athm.HaPaScoSet} and Theorem \ref{Athm.MHaPaSo} the super credulous aggregation operator is Pareto optimal in $\calS$ when individual preferences are IUO Hamming set{s} based.
\end{proof}
\end{proposition}

\begin{proposition}\label{Aprop.MHaPaScoDis}
If individual preferences are IUO Hamming distance based, then the super credulous aggregation operator is not Pareto optimal {in the set of} complete labelings that are compatible ($\approx$) with each of the participants' labelings.
\begin{proof}
Similar to the previous proposition, from Observation \ref{Aobs.HaPaCoDis} and Theorem \ref{Athm.MHaPaSo} the super credulous aggregation operator is not Pareto optimal in $\calS$ ($\calS$ is defined in the previous proposition) when individual preferences are IUO Hamming distance based.
\end{proof}
\end{proposition}
}
{

\begin{proposition}\label{Aprop.MIwPaSo}
If individual preferences are IUO Issue-wise distance based, then the skeptical aggregation operator is Pareto optimal {in the set of} admissible labelings that are smaller or equal ($\sqsubseteq$) to each of the participants' labelings.
\begin{proof}
Let $\calS$ be the set of admissible labelings that are smaller or equal ($\sqsubseteq$) to each of the participants' labelings. Then, $\calS \subseteq \calX$, where $\calX$ is defined in Theorem \ref{Athm.MIwPaSo}. From Theorem \ref{Athm.IwPaSoDis} and Theorem \ref{Athm.MIwPaSo} the skeptical aggregation operator is Pareto optimal in $\calS$ when individual preferences are IUO Issue-wise distance based.
\end{proof}
\end{proposition}

\begin{proposition}\label{Aprop.MIwPaSoDis}
If individual preferences are IUO Issue-wise distance based, then the credulous aggregation operator is not Pareto optimal {in the set of} admissible labelings that are compatible ($\approx$) with each of the participants' labelings.
\begin{proof}
Similar to the previous proposition, from Observation \ref{Aobs.IwPaCoSco} and Theorem \ref{Athm.MIwPaSo} the credulous aggregation operator is not Pareto optimal in $\calS$ ($\calS$ is defined in the previous proposition) when individual preferences are IUO Issue-wise distance based.
\end{proof}
\end{proposition}

\begin{proposition}\label{Aprop.MIwPaScoDis}
If individual preferences are IUO Issue-wise distance based, then the super credulous aggregation operator is not Pareto optimal {in the set of} complete labelings that are compatible ($\approx$) with each of the participants' labelings.
\begin{proof}
Similar to the previous proposition, from Observation \ref{Aobs.IwPaCoSco} and Theorem \ref{Athm.MIwPaSo} the super credulous aggregation operator is not Pareto optimal in $\calS$ ($\calS$ is defined in the previous proposition) when individual preferences are IUO Issue-wise distance based.
\end{proof}
\end{proposition}

}

{
The following example shows how Pareto optimality does not generally carry over from homogeneous preferences to heterogeneous preferences.

\begin{example}\label{Aex:heteroCounter}
Consider the framework and labelings in Figure \ref{Afig:heteroCounter}. Let $\Ag=\{1,2\}$, $\calS=\{\calL,\calL_X\}$ and $\calR=\{\left|\varominus\right|,\left|\varominus_\calW\right|\}$. When $\Ag$ have homogeneous preferences from $\calR$ (i.e. both agents have Hamming distance based preferences or both have Issue-wise distance based preferences), then $\calL$ is Pareto optimal in $\calS$:
\begin{equation}
\begin{aligned}[b]
\calL_1 \left|\varominus\right| \calL=4\;,\; \calL_1 \left|\varominus\right| \calL_X=3 \text{ (i.e. } \calL_X \succ_{1,\left|\varominus\right|} \calL \text{)}\\
\calL_2 \left|\varominus\right| \calL=3\;,\; \calL_2 \left|\varominus\right| \calL_X=4 \text{ (i.e. } \calL \succ_{2,\left|\varominus\right|} \calL_X \text{)}\\
\calL_1 \left|\varominus_\calW\right| \calL=1\;,\; \calL_1 \left|\varominus_\calW\right| \calL_X=2 \text{ (i.e. } \calL \succ_{1,\left|\varominus_\calW\right|} \calL_X \text{)}\\
\calL_2 \left|\varominus_\calW\right| \calL=2\;,\; \calL_2 \left|\varominus_\calW\right| \calL_X=1 \text{ (i.e. } \calL_X \succ_{2,\left|\varominus_\calW\right|} \calL \text{)}\nonumber
\end{aligned}
\end{equation}

However, if agent $1$ has Hamming distance based preferences and agent $2$ has Issue-wise distance based preferences then both agents would strictly prefer $\calL_X$ over $\calL$ and this means that $\calL$ would not be Pareto optimal in $\calS$ when $\Ag$ have heterogeneous preferences from $\calR$.

\end{example}
\begin{figure}[ht]
    \centering
  \includegraphics[scale=0.4]{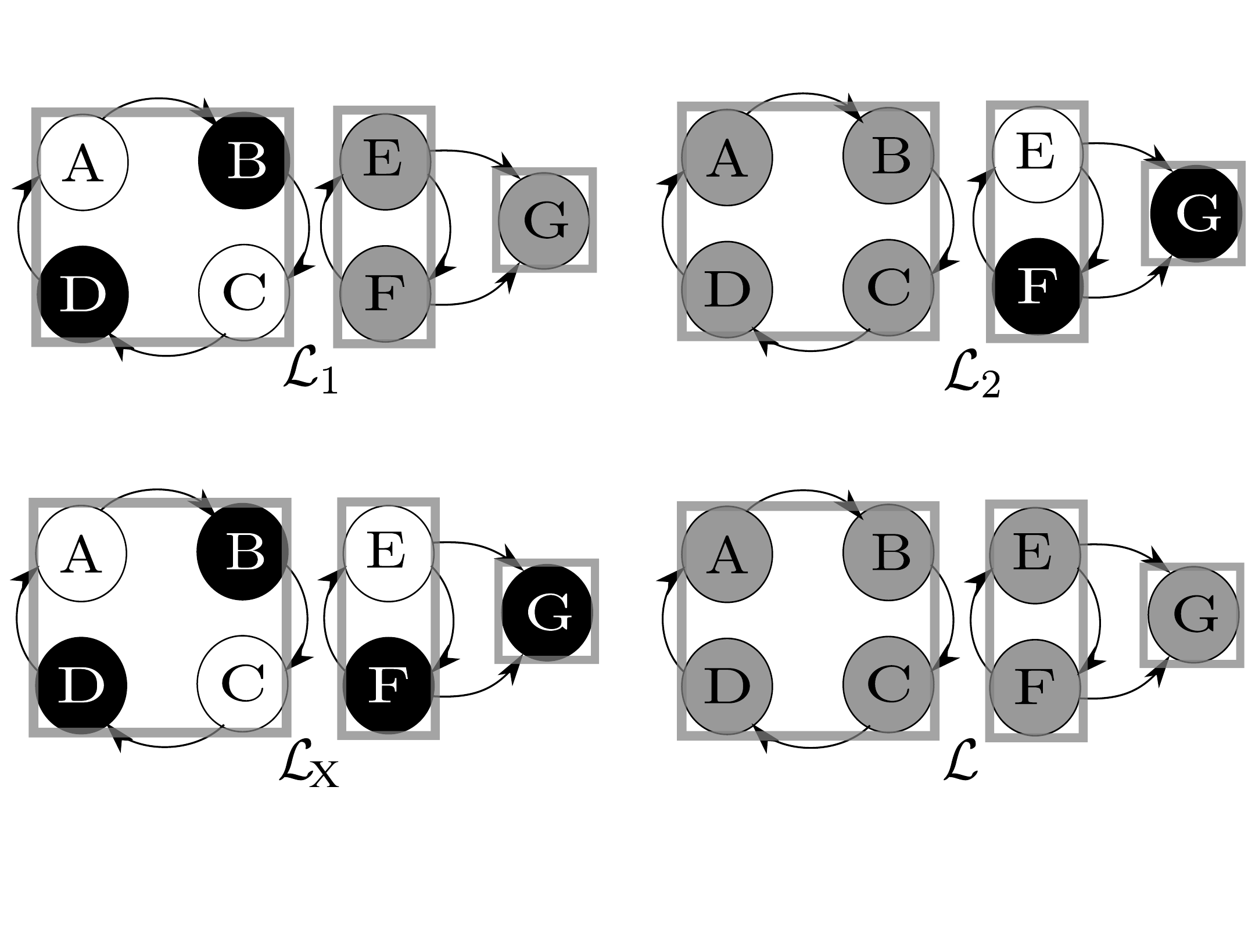}
  \caption{An example showing how a labeling that is Pareto optimal in a set $\calS$ given that $\Ag$ have \emph{homogeneous} preferences from a set $\calR$, might not be Pareto optimal if $\Ag$ have \emph{heterogeneous} preferences from $\calR$. The set of arguments located in one box form an issue.}\label{Afig:heteroCounter}
\end{figure}

\begin{theorem}
Let $\calR=\{\varominus,\varominus^\calM\}$ be a set of preference classes, $\Ag$ be a set of agents, $\calS$ be the set of all admissible labelings that are compatible with each individual's labeling, and $\calL$ be a labeling from $\calS$. If $\calL$ is Pareto optimal in $\calS$ given that $\Ag$ have \emph{homogeneous} preferences from $\calR$, then $\calL$ is Pareto optimal in $\calS$ given that $\Ag$ have \emph{heterogeneous} preferences from $\calR$.
\begin{proof}
Let $\Ag$ be s.t. agents have heterogeneous preferences from $\calR$. Suppose towards a contradiction, that $\calL$ is not Pareto optimal in $\calS$ when each agent $i$ has $c(i)\in\calR$ based preferences. Then, {there is a labeling $\calL_X$ that Pareto dominates $\calL$ w.r.t $c(i)$ based preferences, i.e.} $\exists \calL_X \in \calS$ s.t:
\begin{equation}
(\forall i \in \Ag: \calL_X \succeq_{i,c(i)} \calL) \wedge (\exists j \in \Ag \text{ s.t. } \calL_X \succ_{j,c(j)} \calL)
\end{equation}
where $c(i) \in \calR, \forall i \in \Ag$ (i.e. $c(i)=\varominus$ or $c(i)=\varominus^\calM$). Then{, from the definition of set based preference}:
\begin{equation}
(\forall i \in \Ag: \calL_X \;c(i)\; \calL_i \subseteq \calL \;c(i)\; \calL_i) \wedge (\exists j \in \Ag \text{ s.t. } \calL \;c(j)\; \calL_j \not\subseteq \calL_X \;c(j)\; \calL_j)
\end{equation}
However, given the compatibility of $\calL$ and $\calL_X$ with every individuals' labeling (since $\calL,\calL_X\in \calS$) and from Lemma \ref{Alem.heterosetscomp} (1), if agents who have Hamming set based preferences switched their classes of preferences to IUO Hamming set{s} based preferences or vice versa, then their preferences would not change. As a result, the previous equation would hold even when $c(k)=c(l), \forall k,l \in \Ag$. This means $\calL$ is not Pareto optimal in $\calS$ when $\Ag$ have homogeneous preferences from $\calR$. Contradiction.
\end{proof}
\end{theorem}

\begin{theorem}
Let $\calR=\{\varominus,\varominus_\calW,\varominus^\calM,\varominus_\calW^\calM, \left|\varominus\right|,\left|\varominus_\calW\right|,\left|\varominus^\calM\right|,\left|\varominus_\calW^\calM\right|\}$. The skeptical operator is Pareto optimal in the set of all admissible labelings that are smaller or equal ($\sqsubseteq$) to each of the participants' labelings given that individuals have heterogeneous preferences from $\calR$.  
\begin{proof}
Let $\calS$ be the set of all admissible labelings that are smaller or equal ($\sqsubseteq$) to individuals' labelings. Suppose, towards a contradiction that the skeptical operator is not Pareto optimal in $\calS$ given that the set of individuals $\Ag$ have heterogeneous preferences from $\calR$. Then, $\exists \calL_X \in \calS$ s.t. $\calL_X$ Pareto dominates $\calL$ (given heterogeneous preferences). Then:
\begin{equation}
\exists j \in \Ag \text{ s.t. } \calL_X \succ_{j,c(j)} \calL
\end{equation} 

However, from Theorem 1, $\calL$ is the biggest admissible labeling that is smaller or equal ($\sqsubseteq$) to each individual's labeling. Then, $\forall \calL' \in \calS: \calL' \sqsubseteq \calL \sqsubseteq \calL_i, \forall i \in \Ag$. Hence, for any agent $i$'s labeling $\calL_i$, and for any argument (and consequently, any issue) on which $\calL$ disagrees with $\calL_i$, then $\calL_X$ would disagree in exactly the same way. Contradiction.  
\end{proof}
\end{theorem}
}

\section{Strategy Proofness}\label{Asec.SP}

{
Consider an operator $Op$ that only produces labelings that are compatible ($\approx$) with each individual's labeling. The following lemma shows that every {strategic lie} with the operator $Op$ given IUO Hamming distance based preferences is also a strategic lie given Hamming distance based preferences. This lemma is crucial to show that the benevolence property of lies with the skeptical operator carries over from Hamming distance based preferences to IUO Hamming distance based preferences.

\begin{lemma}\label{Alem.carryOverHam}
Let $Op$ be {a \emph{compatible operator}}. Let $\calL_k$ denote the top preference labeling of agent $k$. Let $P$ be a profile where each agent submits her most preferred labeling, and {let $P'=P_{\calL_k/\calL'_k}$ be a profile that results from $P$ by changing $\calL_k$ to $\calL'_k$}. Let $\calL_{Op}=Op_\AF(P)$ be the outcome when agent $k$ does not lie. Let $X^k_{\left |\varominus^\calM \right |}$ (resp. $X^k_{\left |\varominus\right |}$) be the set of all labelings $\calL'_{Op}$ that satisfy the following two properties:
\begin{enumerate}
\item There exists some labeling $\calL'_k$ s.t. $\calL'_{Op}=Op_\AF(P_{\calL_k/\calL'_k})$ (i.e. $\calL'_{Op}$ is a possible outcome given some lie by agent $k$), and
\item $\calL'_{Op} \succ_{k,\left |\varominus^\calM \right |} \calL_{Op}$ (resp. $\calL'_{Op} \succ_{k,\left |\varominus\right |} \calL_{Op}$).
\end{enumerate}
Then $X^k_{\left |\varominus^\calM \right |} \subseteq X^k_{\left |\varominus\right |}$.

\begin{proof}
$\forall \calL'_{Op} \in X^k_{\left |\varominus^\calM \right |}$, we have:
\begin{enumerate}
\item There exists some labeling $\calL'_k$ s.t. $\calL'_{Op}=so_\AF(P_{\calL_k/\calL'_k})$, and
\item $\calL'_{Op} \succ_{k,\left |\varominus^\calM \right |}\calL_{Op}$. 
\end{enumerate}
We just need to show that $\calL'_{Op} \succ_{k,\left |\varominus\right |}\calL_{Op}$.

Since $\calL'_{Op} \succ_{k,\left |\varominus^\calM \right |}\calL_{Op}$, then $\calL'_{Op} \left |\varominus^\calM \right | \calL_k < \calL_{Op} \left |\varominus^\calM \right | \calL_k$. Then:
\begin{equation}
2 \times |\calL'_{Op} \varominus^{io} \calL_k| + |\calL'_{Op} \varominus^{du} \calL_k| < 2 \times |\calL_{Op} \varominus^{io} \calL_k| + |\calL_{Op} \varominus^{du} \calL_k|
\end{equation}
Since $|\calL_{Op} \varominus^{io} \calL_k|=0$:

\begin{equation}
2 \times |\calL'_{Op} \varominus^{io} \calL_k| + |\calL'_{Op} \varominus^{du} \calL_k| < |\calL_{Op} \varominus^{du} \calL_k| 
\end{equation}
Which implies: 

\begin{equation}
|\calL'_{Op} \varominus^{io} \calL_k| + |\calL'_{Op} \varominus^{du} \calL_k| < |\calL_{Op} \varominus^{du} \calL_k| 
\end{equation}
But $|\calL'_{Op} \varominus \calL_k| = |\calL'_{Op} \varominus^{io} \calL_k| + |\calL'_{Op} \varominus^{du} \calL_k|$ and $|\calL_{Op} \varominus \calL_k| = |\calL_{Op} \varominus^{du} \calL_k|$. Then:

\begin{equation}
|\calL'_{Op} \varominus \calL_k| < |\calL_{Op} \varominus \calL_k| 
\end{equation}
Which means $\calL'_{Op} \succ_{\left |\varominus\right |} \calL_{Op}$. Hence, $\calL'_{Op} \in X^k_{\left |\varominus\right |}$.
\end{proof}
\end{lemma}

We now show that the benevolence property of lies with an operator carries over from Hamming distance based preferences to IUO Hamming distance based preferences.

\begin{theorem}\label{Athm.MHaSpSo}
Consider an operator $Op$ that only produces labelings that are compatible ($\approx$) with each individual's labeling. If all strategic lies are benevolent when agents have Hamming distance based preferences then all strategic lies are benevolent when agents have IUO Hamming distance based preferences.
\begin{proof}
Let $Op$ be {a \emph{compatible operator}}. Let $P$ be a profile, and $\calL'_k$ a strategic lie of agent $k$. Denote $\calL_{Op}=Op_{\AF}(P)$ and $\calL'_{Op}=Op_{\AF}(P_{\calL_k/\calL'_k})$. {From Lemma \ref{Alem.heterosetscomp} (1), since the operator $Op$ only produces labelings that are compatible with all individuals' labelings, then} for every agent $j$ s.t. $j \neq k$: $(\calL_{Op} \succeq_{j,\left |\varominus\right |} \calL'_{Op}$ iff $\calL_{Op} \succeq_{j,\left |\varominus^\calM \right |} \calL'_{Op})$ i.e. Hamming distance based preferences and IUO Hamming distance based preferences are equivalent for all agents other than agent $k$.

Now given Lemma \ref{Alem.carryOverHam}, every {strategic lie} with the operator $Op$ given IUO Hamming distance based preferences is also a strategic lie given Hamming distance based preferences. However, all those lies are benevolent for every agent $j \neq k$ whether she has Hamming distance based preferences or IUO Hamming distance based preferences. Hence, every lie given IUO Hamming distance based preferences is benevolent. 
\end{proof}
\end{theorem}

Consider an operator $Op$ that only produces labelings that are compatible ($\approx$) with each individual's labeling. The following lemma shows that every {strategic lie} with the operator $Op$ given IUO Issue-wise distance based preferences is also a strategic lie given Issue-wise distance based preferences. This lemma is crucial to show that the benevolence property of lies with the operator $Op$ carries over from Issue-wise distance based preferences to IUO Issue-wise distance based preferences.

\begin{lemma}\label{Alem.carryOverIss}
Let $Op$ be {a \emph{compatible operator}}. Let $\calL_k$ denote the top preference labeling of agent $k$. Let $P$ be a profile where each agent submits her most preferred labeling, and {let $P'=P_{\calL_k/\calL'_k}$ be a profile that results from $P$ by changing $\calL_k$ to $\calL'_k$}. Let $\calL_{Op}=Op_\AF(P)$ be the outcome when agent $k$ does not lie. Let $X^k_{\left |\varominus_\calW^\calM \right |}$ (resp. $X^k_{\left |\varominus_\calW\right |}$) be the set of all labelings $\calL'_{Op}$ that satisfy the following two properties:
\begin{enumerate}
\item There exists some labeling $\calL'_k$, $\calL'_{Op}=Op_\AF(P_{\calL_k/\calL'_k})$, and
\item $\calL'_{Op} \succ_{k,\left |\varominus_\calW^\calM \right |} \calL_{Op}$ (resp. $\calL'_{Op} \succ_{k,\left |\varominus_\calW\right |} \calL_{Op}$).
\end{enumerate}
Then $X^k_{\left |\varominus_\calW^\calM \right |} \subseteq X^k_{\left |\varominus_\calW\right |}$.

\begin{proof}
This proof is similar to the one in Lemma \ref{Alem.carryOverHam}.
\end{proof}
\end{lemma}


\begin{theorem}\label{Athm.MIwSpSo}
Consider an operator $Op$ that only produces labelings that are compatible ($\approx$) with each individual's labeling. If all strategic lies are benevolent when agents have Issue-wise distance based preferences then all strategic lies are benevolent when agents have IUO Issue-wise distance based preferences.
\begin{proof}
{This proof is similar to the one for Theorem \ref{Athm.MHaSpSo}, with the use of Lemma \ref{Alem.heterosetscomp} (2) and Lemma \ref{Alem.carryOverIss}.}
\end{proof}
\end{theorem}

}


\begin{observation}\label{Aobs.HaSpSo}
The skeptical aggregation operator is not strategy proof for neither Hamming set nor Hamming distance based preferences. Consider the three labelings in Figure \ref{Afig:SPSo}. Labeling $\calL_1$ of agent 1 when aggregated with $\calL_2$ gives labeling $\calL_3$, which disagrees with $\calL_1$ on all three arguments. But, when the agent strategically lies and reports labeling $\calL_2$ instead, the result of the aggregation is the same labeling $\calL_2$, which differs only on two arguments $\{A, B\}$. The example is valid for both Hamming set and Hamming distance based preferences.
\end{observation}

\begin{figure}[ht]
    \centering
  \includegraphics[scale=0.4]{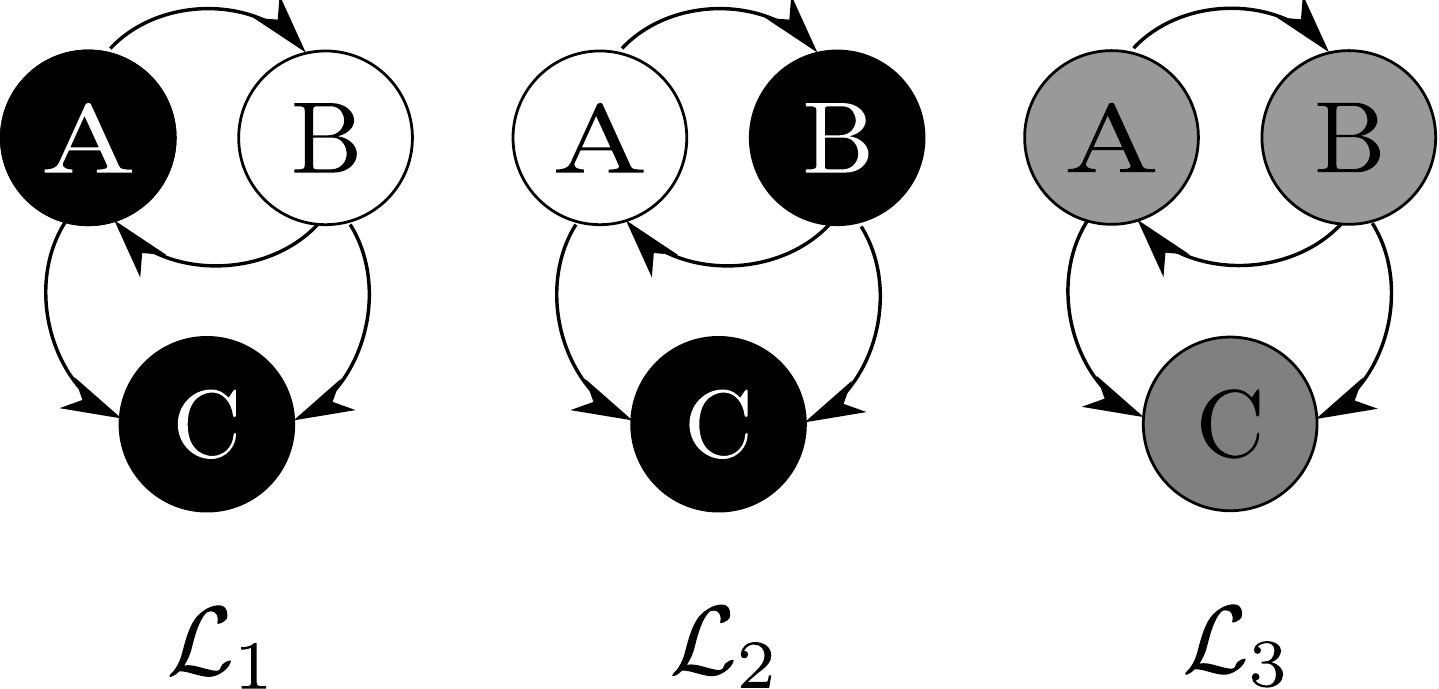}
  \caption{The skeptical operator is not strategy proof.}\label{Afig:SPSo}
\end{figure}

\begin{observation}\label{Aobs.HaSpCoSco}
The credulous (resp. super credulous) aggregation operator is not strategy proof for neither Hamming set nor Hamming distance based preferences. See the example in Figure \ref{Afig:SPCoSco}. Labeling $\calL_2$ of agent $2$ when aggregated with $\calL_1$ gives labeling $\calL_{CO}$, which disagrees with $\calL_2$ on the two arguments. But, when the agent strategically lies and reports $\calL'_2$ instead, the result of the aggregation is $\calL'_{CO}$, which matches the labeling $\calL_2$. This lie by agent $2$ makes the agent with labeling $\calL_1$ worse off. The example is valid for both Hamming set and Hamming distance based preferences.
\end{observation}


\begin{figure}[ht]
    \centering
  \includegraphics[scale=0.4]{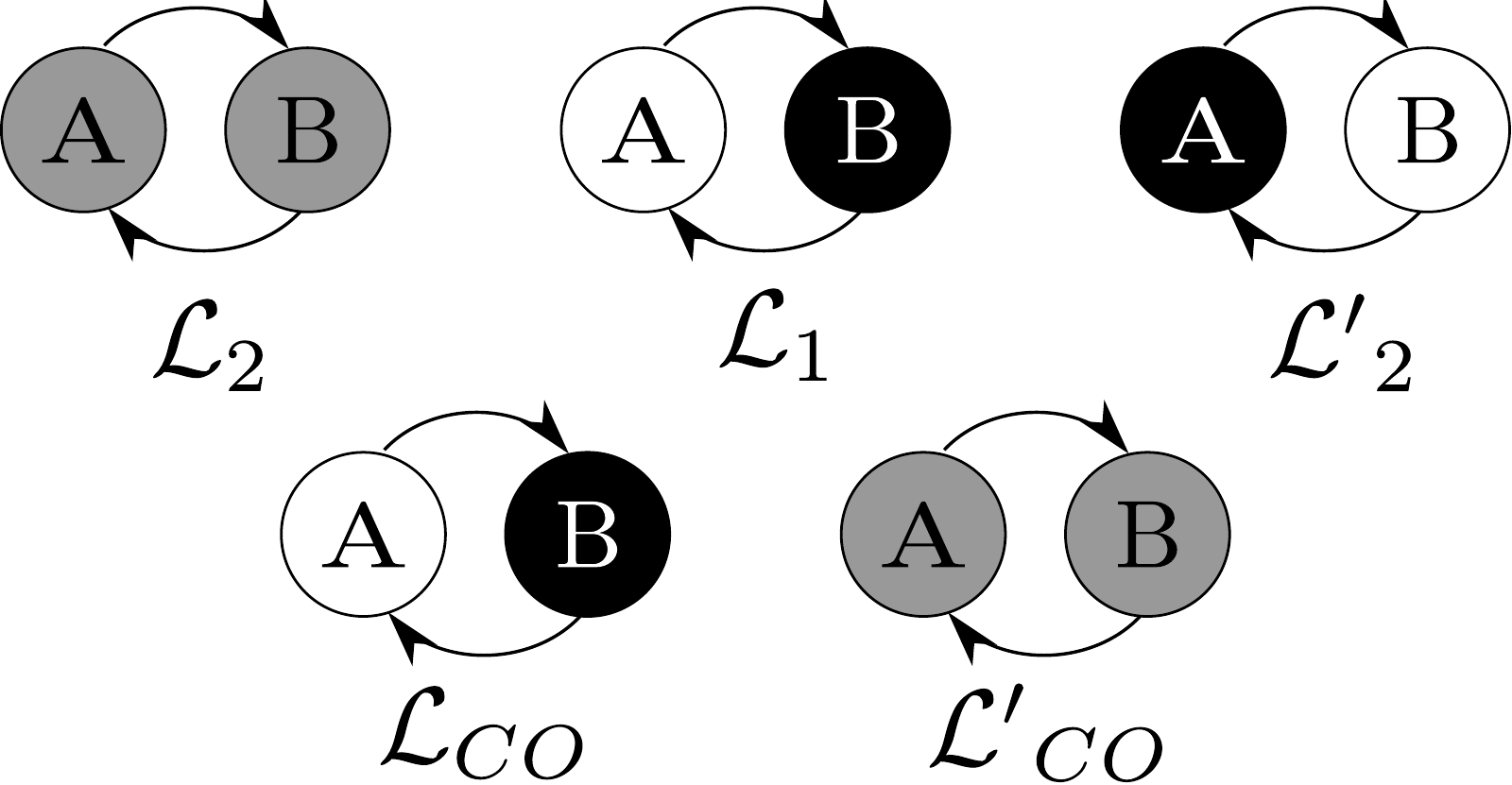}
  \caption{The (super) credulous operator is not strategy proof.}\label{Afig:SPCoSco}
\end{figure}

\begin{theorem}\label{Athm.HaSpSoSet}
Consider the skeptical aggregation operator and Hamming set based preferences. For any agent, her strategic lies are benevolent. 
\begin{proof}
Let $P$ be a profile, and $\calL_k'$ a strategic lie of agent $k$. Denote $\calL_{SO} = so_{\AF}(P)$ and $\calL'_{SO} = so_{\AF}(P_{\calL_k/\calL_k'})$. Agent $k$'s preference is $\calL'_{SO} \succ_k \calL_{SO}$ (i). We will show that for any agent $i \neq k$, we have $\calL'_{SO} \succ_i \calL_{SO}$.  
Since the skeptical aggregation operator produces social outcomes that are less or equally committed to all the individual labelings, we have that $\calL'_{SO} \sqsubseteq \calL_i$ for all $i \neq k$ (ii). 
Similarly, we have $\calL_{SO} \sqsubseteq \calL_k$ (iii). From (i) and (iii), by Lemma \ref{Alem.prefSetMoreComm}, we have that $\calL_{SO} \sqsubseteq \calL'_{SO}$ (iv). From (iv) and (ii) we have $\calL_{SO} \sqsubseteq \calL'_{SO} \sqsubseteq \calL_i$ for all $i \neq k$. Finally, we can apply Lemma \ref{Alem.moreCommPref} to obtain $\calL'_{SO} \succeq_i \calL_{SO}$ for all $i \neq k$ (v). We showed that a lie cannot be malicious, now we show that it is benevolent.

(iii) implies $\myundec(\calL_{k}) \subseteq \myundec(\calL_{SO})$ (vi). (i) and (vi) imply $\exists A \in \mydec(\calL_{k}) : A \in \myundec(\calL_{SO}) \wedge A \in \mydec(\calL_{SO}')$ (vii). From (vii), (ii) and (v) $\calL_{SO}' \succ_i \calL_{SO}$ for $i \neq k$.
\end{proof}
\end{theorem}

\begin{theorem}\label{Athm.HaSpSoDis}
Consider the skeptical aggregation operator and Hamming distance based preferences. For any agent, her strategic lies are benevolent.
\begin{proof}
Let $P$ be a profile, and $\calL_k'$ a strategic lie of agent $k$ whose most preferred labeling is $\calL_k$. Denote $\calL_{SO} = so_{\AF}(P)$ and $\calL'_{SO} = so_{\AF}(P_{\calL_k/\calL_k'})$. We will show that, if $\calL'_{SO}$ is strictly preferred to $\calL_{SO}$ by agent $k$, then it is also strictly preferred by any other agent. Without loss of generality we can take agent $j, j \neq k$,whose most preferred labeling is $\calL_j$. 

Let us partition the arguments into the following disjoint groups:
\begin{itemize}
\item $\calX = \mydec(\calL_{SO}) \setminus \mydec(\calL_{SO}')$ (decided arguments that became undecided).
\item $\calY = \mydec(\calL_{SO}') \setminus \mydec(\calL_{SO})$ (undecided arguments that became decided).
\item $\calZ = \mydec(\calL_{SO}') \cap \mydec(\calL_{SO})$ (arguments decided in both labelings).
\item $\calV = \myundec(\calL_{SO}') \cap \myundec(\calL_{SO})$ (arguments undecided in both labelings).
\end{itemize}
Labelings $\calL_{SO}$ and $\calL_{SO}'$ agree on the arguments in $\calV$ (which are labeled $\myundec$) and $\calZ$ (whose arguments are labeled $\myin$ or $\myout$). For the arguments in $\calZ$ there are no $\myin-\myout$ conflicts between $\calL_{SO}$ and $\calL_{SO}'$ as the skeptical aggregation operator guarantees social outcomes less or equally committed than $\calL_j$. Therefore, only arguments from $\calX$ and $\calY$ have an impact on the Hamming distance. 

Both labelings $\calL_k$ and $\calL_j$ agree with $\calL_{SO}$ on the arguments in $\calX$ because $\calL_{SO}$ decides on those arguments and is less or equally committed than both labelings. On the other side, $\calL_{SO}'$ remains undecided on the arguments in $\calX$ so both labelings $\calL_k$ and $\calL_j$ disagree with $\calL_{SO}'$ on $\calX$. 
 
$\calL_{SO}'$ is less or equally committed than $\calL_j$ so, as above, we obtain that on the arguments in $\calY$, $\calL_j$ agrees with $\calL_{SO}'$ and disagrees with $\calL_{SO}$. On the contrary, $\calL_{SO}'$ does not have to be less or equally committed than $\calL_k$ and so, for agent $k$, some of the arguments from $\calY$ increase the distance and some of them decrease. If agent $k$ prefers $\calL_{SO}'$ to $\calL_{SO}$, then the number of the arguments decreasing the distance must be greater than the number of those increasing by more than $|\calX|$. But for agent $j$ all the arguments from $\calY$ are decreasing the distance, as $\calL_j$ agrees with $\calL_{SO}'$ on the whole $\calY$. So, if agent $k$ gains by switching to labeling $\calL_{SO}'$, agent $j$ needs to gain at least the same.
\end{proof}
\end{theorem}

{


\begin{observation}\label{Aobs.SpSo}
The skeptical aggregation operator is not strategy proof for neither Issue-wise set nor Issue-wise distance based preferences. Consider the three labelings in Figure \ref{Afig:SPSoI}.\footnote{This figure is the same as Fig \ref{Afig:SPSo} with issues being evidenced.} Labeling $\calL_1$ of agent $1$ when aggregated with $\calL_2$ gives labeling $\calL_3$, which disagrees with $\calL_1$ on both of the two issues. But, when the agent strategically lies and reports labeling $\calL_2$ instead, the result of the aggregation is the same labeling $\calL_2$, which differs only on one issue $\{\{A,B\}\}$. The example is valid for both Issue-wise set and Issue-wise distance based preferences.
\end{observation}

\begin{figure}[ht]
    \centering
  \includegraphics[scale=0.4]{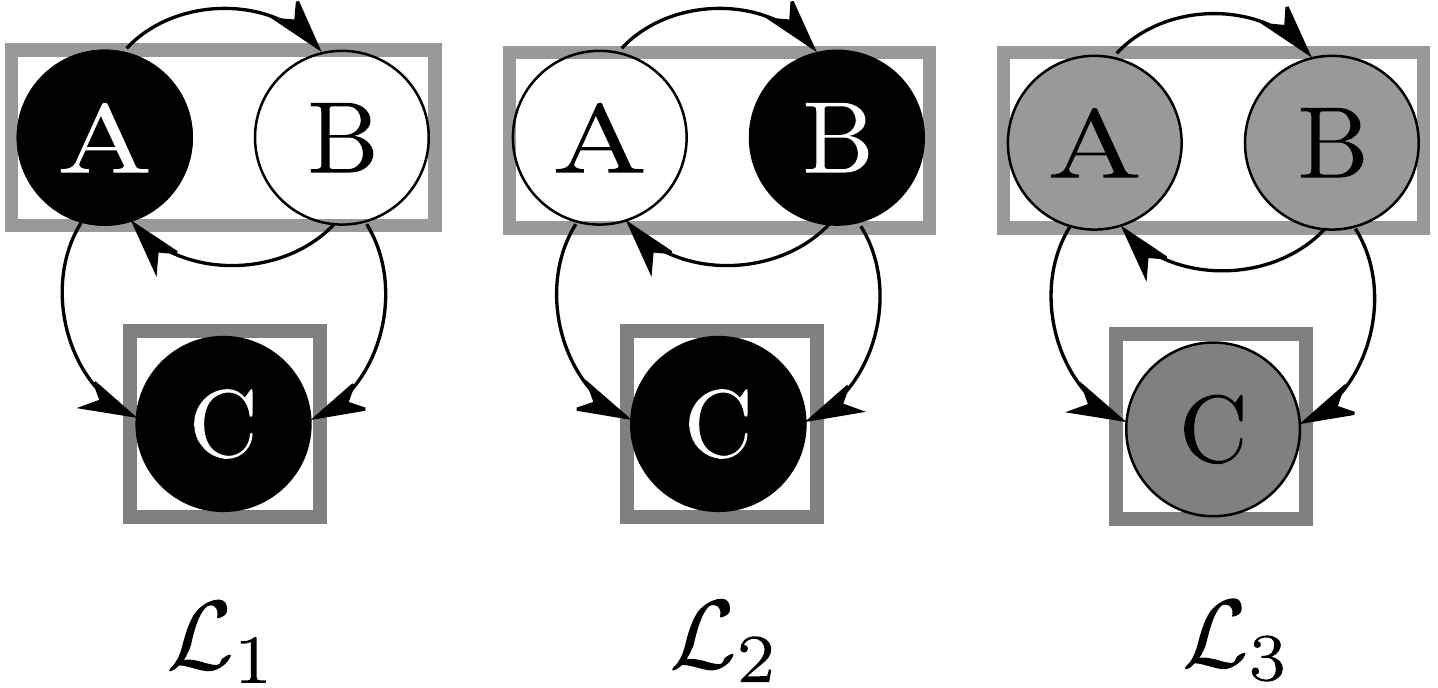}
  \caption{The skeptical operator is not strategy proof. The set of arguments located in one box form an issue.}\label{Afig:SPSoI}
\end{figure}

\begin{observation}\label{Aobs.SpCoSco}
The credulous (resp. super credulous) aggregation operator is not strategy proof for neither Issue-wise set nor Issue-wise distance based preferences. In Figure \ref{Afig:SPCoScoI},\footnote{This figure is the same as Fig \ref{Afig:SPCoSco} with issues being evidenced.} labeling $\calL_2$ of agent $2$ when aggregated with $\calL_1$ gives labeling $\calL_{CO}$, which disagrees with $\calL_2$ on the one and only issue. But, when the agent strategically lies and reports $\calL'_2$ instead, the result of the aggregation is $\calL'_{CO}$, which matches the labeling $\calL_2$. This lie by agent $2$ makes the agent with labeling $\calL_1$ worse off. The example is valid for both Issue-wise set and Issue-wise distance based preferences.
\end{observation}

\begin{figure}[ht]
    \centering
  \includegraphics[scale=0.4]{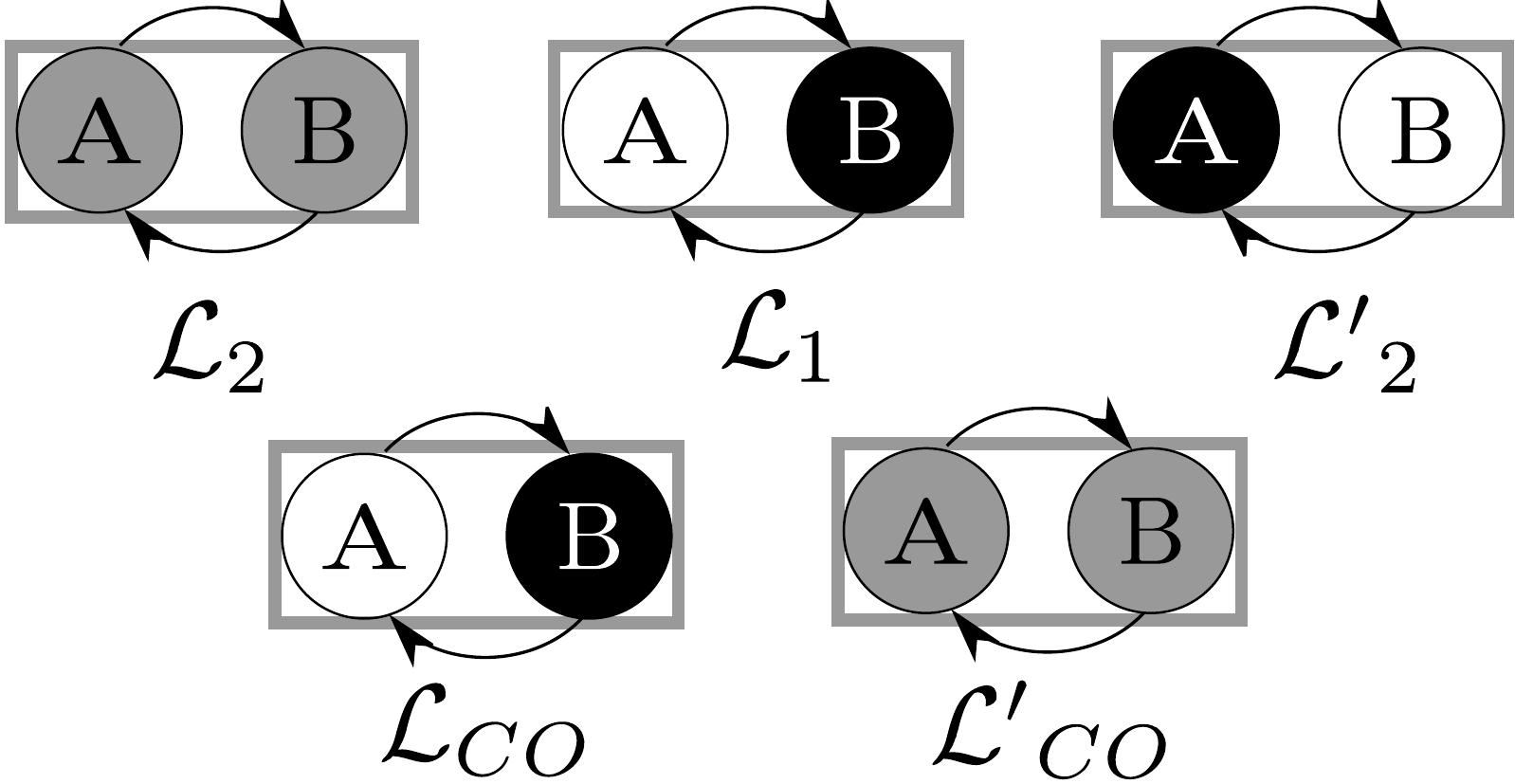}
  \caption{The (super) credulous operator is not strategy proof. The set of arguments located in one box form an issue.}\label{Afig:SPCoScoI}
\end{figure}

\begin{theorem}\label{Athm.SpSoDis}
Consider the skeptical aggregation operator and Issue-wise distance based preferences. For any agent, her strategic lies are benevolent.
\begin{proof}
Let $P$ be a profile of labelings, and $\calL'_k$ be a strategic lie of agent $k$ whose most preferred labeling is $\calL_k$. Denote $\calL_{SO}=so_{\AF}(P)$ and $\calL'_{SO}=so_{\AF}(P_{\calL_k/\calL'_k})$. We show that if $\calL'_{SO}$ is strictly preferred by an agent $k$ then it is also strictly preferred by any other agent. Without loss of generality, we can take agent $j$, $j \neq k$, whose most preferred labeling is $\calL_j$.

Let $\calI_{de}(\calL)$ (resp. $\calI_{un}(\calL)$) be the set of issues, each of which has arguments that are only decided (resp. undecided) according to $\calL$. We call $\calI_{de}(\calL)$ (resp. $\calI_{un}(\calL)$) a decided (resp. undecided) issue (w.r.t $\calL$). Let us partition the issues into the following disjoint groups: 

\begin{itemize}
\item $\calX = \calI_{de}(\calL_{SO})\setminus\calI_{de}(\calL'_{SO})$ (decided issues that became undecided).
\item $\calY = \calI_{de}(\calL'_{SO})\setminus\calI_{de}(\calL_{SO})$ (undecided issues that became decided).
\item $\calZ = \calI_{de}(\calL_{SO}) \cap \calI_{de}(\calL'_{SO})$ (issues decided in both labelings).
\item $\calV = \calI_{un}(\calL_{SO}) \cap \calI_{un}(\calL'_{SO})$ (issues undecided in both labelings).
\end{itemize}

The rest is similar to Theorem \ref{Athm.HaSpSoDis}, but using issues instead of arguments.
\end{proof}
\end{theorem}



\begin{theorem}\label{Athm.MHaSpSoSet}
The skeptical aggregation operator is strategy proof when individuals have IUO Hamming set{s} based preferences.
\begin{proof}
Let $P$ be a profile, $\calL_k$ be the top preference of agent $k$, and $\calL'_k \neq \calL_k$ be any potential lie that agent $k$ might consider. Denote $\calL_{SO}=so_{\AF}(P)$ and $\calL'_{SO}=so_{\AF}(P_{\calL_k/\calL'_k})$. We will show that $\neg (\calL'_{SO} \succ_{k,\varominus^\calM} \calL_{SO})$. Which means, we need to show:
\begin{equation}
\neg ((\calL'_{SO} \succeq_{k,\varominus^\calM} \calL_{SO}) \wedge \neg (\calL_{SO} \succeq_{k,\varominus^\calM} \calL'_{SO}))
\end{equation}
\begin{equation}
\neg(\calL'_{SO} \succeq_{k,\varominus^\calM} \calL_{SO}) \vee (\calL_{SO} \succeq_{k,\varominus^\calM} \calL'_{SO})
\end{equation}
In other words:
\begin{equation}
\begin{aligned}[b]
\neg ((\calL'_{SO} \varominus^{io} \calL_k \subseteq \calL_{SO}\varominus^{io} \calL_k) \wedge (\calL'_{SO} \varominus^{du} \calL_k \subseteq \calL_{SO}\varominus^{du} \calL_k))\\ \vee (\calL_{SO} \succeq_{k,\varominus^\calM} \calL'_{SO})
\end{aligned}
\end{equation}
To reformulate, we only need to show that one of the following holds:

\begin{enumerate}
\item $\neg (\calL'_{SO} \varominus^{io} \calL_k \subseteq \calL_{SO} \varominus^{io} \calL_k)$, {or}
\item $\neg (\calL'_{SO} \varominus^{du} \calL_k \subseteq \calL_{SO} \varominus^{du} \calL_k)$, or
\item $\,$ 
\begin{itemize}
\item[(a)] $\calL_{SO} \varominus^{io} \calL_k \subseteq \calL'_{SO} \varominus^{io} \calL_k$, and
\item[(b)] $\calL_{SO} \varominus^{du} \calL_k \subseteq \calL'_{SO} \varominus^{du} \calL_k$.
\end{itemize}
\end{enumerate}

First, by definition, $\calL_{SO}$ is less or equally committed ($\sqsubseteq$) than $\calL_k$. So, $\calL_{SO} \varominus^{io} \calL_k=\emptyset$, However, this is not the case for $\calL'_{SO}$ and $\calL_k$. So, $\calL'_{SO} \varominus^{io} \calL_k$ might not be an empty set. Hence, $\calL_{SO} \varominus^{io} \calL_k \subseteq \calL'_{SO} \varominus^{io} \calL_k$ i.e. (3)(a) is satisfied. Now we show that either (1),(2) or (3)(b) is satisfied.  

Suppose (1) and (2) are violated and we will show that (3)(b) is then satisfied. This shows that (1), (2), and (3)(b) cannot be all violated together. 

Since (1) is violated and since $\calL_{SO} \varominus^{io} \calL_k = \emptyset$ then $\calL'_{SO} \varominus^{io} \calL_k = \emptyset$ (i). Since (2) is violated then $\forall a: (a \in \calL'_{SO} \varominus^{du} \calL_k \Rightarrow a \in \calL_{SO} \varominus^{du} \calL_k)$ (ii). Note that $\forall a: (a \in \calL'_{SO} \varominus^{du} \calL_k \Rightarrow (a \in \myundec(\calL'_{SO}) \wedge a \in \mydec(\calL_k)))$(iii). Otherwise, we would have $a \in \mydec(\calL'_{SO}) \wedge a \in \myundec(\calL_k)$ and from (ii) we would have $a \in \mydec(\calL_{SO}) \wedge a \in \myundec(\calL_k)$ which contradicts $\calL_{SO} \sqsubseteq \calL_k$.

From (i) and (iii), $\forall a \in \myin(\calL'_{SO}) \Rightarrow a \in \myin(\calL_k)$ (iv) (from (i), $\calL_k(a) \neq \myout$, and from (iii), $\calL_k(a) \neq \myundec$). Similarly, from (i) and (iii), $\forall a \in \myout(\calL'_{SO}) \Rightarrow a \in \myout(\calL_k)$ (v). From (iv) and (v), $\calL'_{SO} \sqsubseteq \calL_k$. Since $\forall i \neq k$: $\calL'_{SO} \sqsubseteq \calL_i$, then $\forall i \in \Ag$: $\calL'_{SO} \sqsubseteq \calL_i$. By Theorem 1, $\calL'_{SO} \sqsubseteq \calL_{SO}$. Then, $\myundec(\calL_{SO}) \subseteq \myundec(\calL'_{SO})$ (vi). 

Now, $\forall a \in \calL_{SO} \varominus^{du} \calL_k$ then $a \in \myundec(\calL_{SO}) \wedge a \in \mydec(\calL_k)$. From (vi), $a \in \myundec(\calL'_{SO})$. Thus, $a \in \calL'_{SO} \varominus^{du} \calL_k$. Then, (3)(b) is satisfied. 
\end{proof}
\end{theorem}

\begin{observation}\label{Aobs.MHaSpSo}
The skeptical aggregation operator is not strategy proof when individuals have IUO Hamming distance based preferences. Consider the three labelings in Figure \ref{Afig:SPSoIUOHamDis}. Labeling $\calL_1$ of agent 1 when aggregated (using skeptical operator) with $\calL_2$ gives labeling $\calL_3$, which differs from $\calL_1$ on all five arguments with respect to $\mydec-\myundec$ Hamming set. Then, $\calL_1 \left |\varominus^\calM \right | \calL_3 = 2 \times 0 + 1 \times 5= 5$. But, when the agent strategically lies and reports labeling $\calL_2$ instead, the result of the aggregation is the same labeling $\calL_2$, which differs only on two arguments $\{A, B\}$ with respect to $\myin-\myout$ Hamming set. Then, $\calL_1 \left |\varominus^\calM \right | \calL_2 = 2 \times 2 + 1 \times 0= 4$. 
\end{observation}

\begin{figure}[ht]
    \centering
  \includegraphics[scale=0.6]{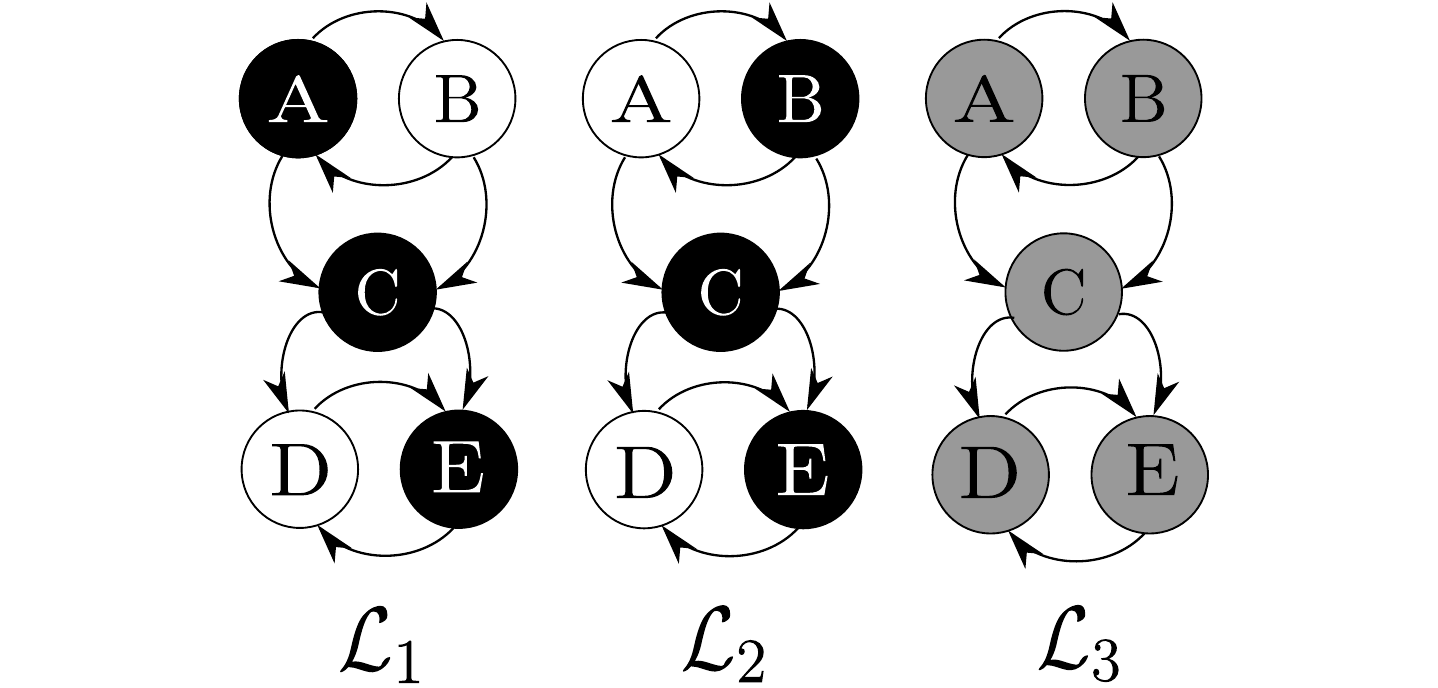}
  \caption{The skeptical operator is not strategy proof when agents have IUO Hamming distance preferences.}\label{Afig:SPSoIUOHamDis}
\end{figure}


\begin{observation}\label{Aobs.MHaSpCoSco}
The credulous (resp. super credulous) aggregation operator is not strategy proof for neither IUO Hamming set{s} nor IUO Hamming distance based preferences. The example in Figure \ref{Afig:SPCoSco} can serve as a counterexample for the case where individuals have IUO Hamming set{s} (or IUO Hamming distance) based preferences. The agent with labeling $\calL_2$ can insincerely report $\calL'_2$ to obtain her preferred labeling. This makes an agent with labeling $\calL_1$ worse off. 
\end{observation}


\begin{proposition}\label{Aprop.MHaSpSo}
Consider the skeptical aggregation operator and IUO Hamming distance based preferences. For any agent, her strategic lies are benevolent.
\begin{proof}
From Theorem \ref{Athm.HaSpSoDis} and Theorem \ref{Athm.MHaSpSo}, the strategic lies are benevolent when individuals have IUO Hamming distance based.
\end{proof}
\end{proposition}



\begin{observation}\label{Aobs.MSpSo}
The skeptical aggregation operator is not strategy proof when individuals have IUO Issue-wise distance based preferences. Consider the three labelings in Figure \ref{Afig:SPSoIUOIssueDis}.\footnote{This figure is the same as Fig \ref{Afig:SPSoIUOHamDis} with issues being evidenced.} Labeling $\calL_1$ of agent 1 when aggregated (using skeptical operator) with $\calL_2$ gives labeling $\calL_3$, which differs from $\calL_1$ on all three issues with respect to $\mydec-\myundec$ Issue-wise set. Then, $\calL_1 \left |\varominus_\calW^\calM \right | \calL_3 = 2 \times 0 + 1 \times 3= 3$. But, when the agent strategically lies and reports labeling $\calL_2$ instead, the result of the aggregation is the same labeling $\calL_2$, which differs only on one is $\{\{A, B\}\}$ with respect to $\myin-\myout$ Issue-wise set. Then, $\calL_1 \left |\varominus_\calW^\calM \right | \calL_2 = 2 \times 1 + 1 \times 0= 2$. 
\end{observation}

\begin{figure}[ht]
    \centering
  \includegraphics[scale=0.6]{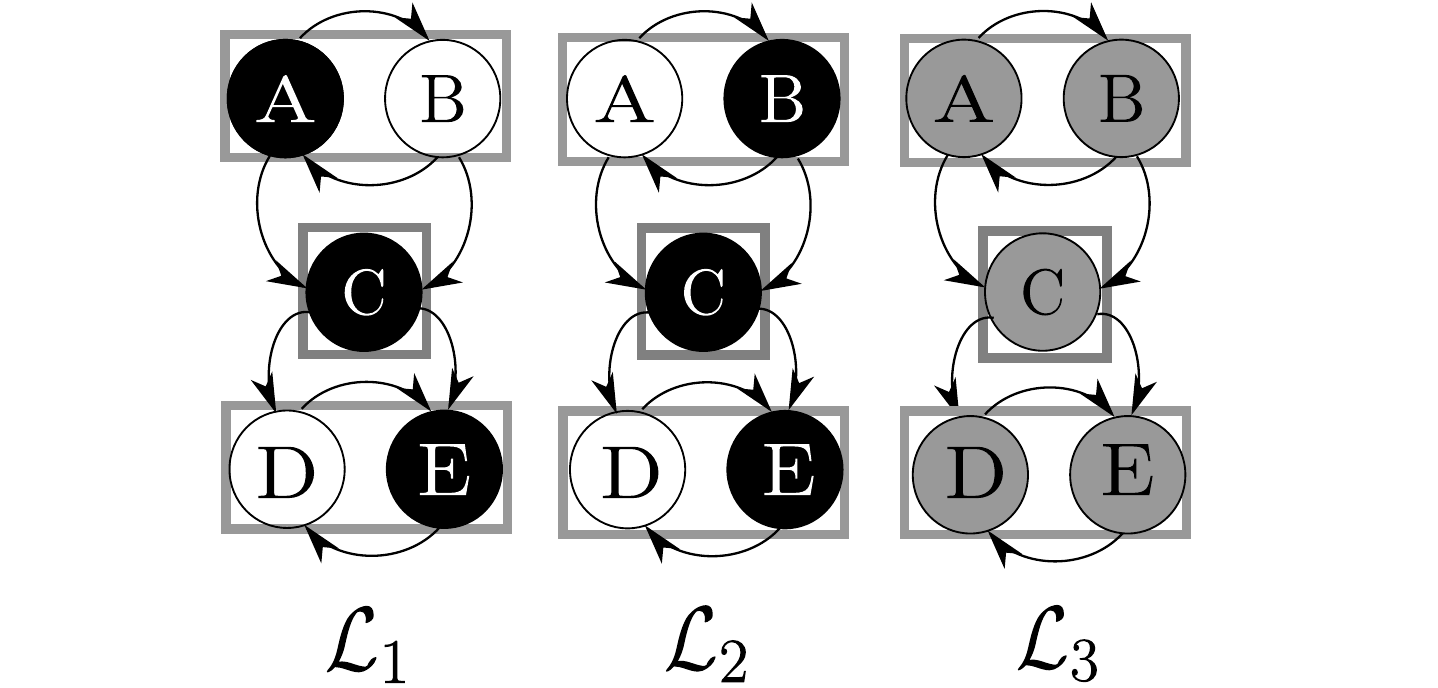}
  \caption{The skeptical operator is not strategy proof when agents have IUO Issue-wise distance preferences. The set of arguments located in one box form an issue.}\label{Afig:SPSoIUOIssueDis}
\end{figure}


\begin{observation}\label{Aobs.MSpCoSco}
The credulous and super credulous aggregation operators are not strategy proof when individuals have IUO Issue-wise set{s} (resp. distance) based preferences. The example in Figure \ref{Afig:SPCoScoI} can serve as a counter example for the case where individuals have IUO Issue-wise set{s} (resp. distance) based preferences. The agent with labeling $\calL_2$ can insincerely report $\calL'_2$ to obtain her preferred labeling. This makes an agent with labeling $\calL_1$ worse off. 
\end{observation}


\begin{proposition}\label{Aprop.MIwSpSo}
Consider the skeptical aggregation operator and IUO Issue-wise distance based preferences. For any agent, her strategic lies are benevolent.
\begin{proof}
From Theorem \ref{Athm.SpSoDis} and Theorem \label{Athm.MIwSpSo}, the strategic lies are benevolent when individuals have IUO Issue-wise distance based.
\end{proof}
\end{proposition}

{

\begin{theorem}
Let $\calF$ be the set of all possible classes of preferences, $\calR$ be some set s.t. $\calR \subseteq \calF$, and $\Ag$ be the set of agents. If an operator is strategy proof given that $\Ag$ have homogeneous preferences from $\calR$, then it is strategy proof given that $\Ag$ have heterogeneous preferences from $\calR$.
\begin{proof}
Let $Op$ be an operator that is strategy proof given that $\Ag$ have homogeneous preferences from $\calR$ (i.e. $\forall i,j \in \Ag$ $c(i)=c(j) \in \calR$). Then, there exists no single agent $j$ that has an incentive to lie about her preferences (given that all agents have the same class of preferences). Note that agent $j$ has no incentive to lie given that the submitted labelings by all agents other than $j$ are fixed. Hence, the classes of preferences that are assumed for any agent $k \neq j$ do not affect the incentive of agent $j$ to lie or otherwise. Thus, the same result would hold whether other agents' preferences are different from $c(j)$ or not.
\end{proof}
\end{theorem}

}

\end{document}